\documentclass{article}

\usepackage{microtype}
\usepackage{graphicx}
\usepackage{subcaption}
\usepackage{booktabs} 

\usepackage{hyperref}

\usepackage{xcolor} 
\usepackage{booktabs}
\usepackage{xcolor}
\newcommand{\tbd}[1]{\textcolor{red}{\textbf{??}}} %
\usepackage{lipsum}

\usepackage{standalone} 
\usepackage{tikz}
\usetikzlibrary{arrows.meta, positioning, calc, shapes.geometric, backgrounds, fit}

\usepackage{multirow}

\usepackage[preprint]{icml2026/icml2026}

\usepackage{amsmath}
\usepackage{amssymb}
\usepackage{mathtools}
\usepackage{amsthm}

\usepackage[capitalize,noabbrev]{cleveref}

\theoremstyle{plain}
\newtheorem{theorem}{Theorem}[section]
\newtheorem{proposition}[theorem]{Proposition}

\theoremstyle{definition}

\theoremstyle{remark}

\usepackage[textsize=tiny]{todonotes}

\begin{document}

\twocolumn[
  \icmltitle{Graph2TS: Structure-Controlled Time Series Generation via Quantile-Graph VAEs}



  \icmlsetsymbol{equal}{*}

  \begin{icmlauthorlist}
    \icmlauthor{Shaoshuai Du}{equal,uva}
    \icmlauthor{Joze M. Rozanec}{equal,utw,ijs}
    \icmlauthor{Andy Pimentel}{uva}
    \icmlauthor{Ana-Lucia Varbanescu}{uva,utw}
  \end{icmlauthorlist}

  \icmlaffiliation{uva}{Informatics Institute, University of Amsterdam, Amsterdam, The Netherlands}
  \icmlaffiliation{utw}{Computer Architecture for Embedded Systems, University of Twente, Enschede, The Netherlands}
  \icmlaffiliation{ijs}{Laboratory of Artificial Intelligence, Institute Jozef Stedan, Ljubljana, Slovenia}

  \icmlcorrespondingauthor{Shaoshuai Du}{s.du@uva.nl}

  \icmlkeywords{Machine Learning, Generative Models, Graphs,  Time Series}

  \vskip 0.3in
]



\printAffiliationsAndNotice{}  

\begin{abstract}

Although recent generative models can produce time series with close marginal distributions, they often face a fundamental tension between preserving global temporal structure and modeling stochastic local variations, particularly for highly volatile signals with weak or irregular periodicity. Direct distribution matching in such settings can amplify noise or suppress meaningful temporal patterns. In this work, we propose a structure–residual perspective on time-series generation, viewing temporal data as the combination of a structural backbone and stochastic residual dynamics, thereby motivating the separation of global organization from sample-level variability.

Based on this insight, we represent time-series structure using a quantile-based transition graph that compactly captures global distributional and temporal dependencies. Building on this representation, we propose \textbf{Graph2TS}, a quantile-graph conditioned variational autoencoder that performs cross-modal generation from structural graphs to time series. By conditioning generation on structure rather than labels or metadata, the model preserves global temporal organization while enabling controlled stochastic variation. Experiments on diverse datasets, including sunspot, electricity load, ECG, and EEG signals, demonstrate improved distributional fidelity, temporal alignment, and representativeness compared to diffusion- and GAN-based baselines, highlighting structure-controlled and cross-modal generation as a promising direction for time-series modeling.

\end{abstract}

\section{Introduction}
Time series generation is a fundamental problem in many domains, including healthcare, finance, and scientific modeling~\cite{DBLP:journals/jzusc/LinLLLG24diffusion_survey,DBLP:journals/corr/abs-2107-11098gan_ts_survey}. Unlike images or text, time series exhibit strong temporal dependencies, structured oscillatory patterns, and stochastic fluctuations, making realistic generation particularly challenging~\cite{DBLP:journals/corr/abs-2107-11098gan_ts_survey,DBLP:journals/corr/abs-2505-20446_ts_gen,DBLP:journals/ijmir/YangCLWSYXG25_multimodal}.

A key difficulty lies in the trade-off between preserving \emph{structural fidelity} - such as global wave patterns and frequency content - and modeling \emph{local variability}, including short-term fluctuations and noise~\cite{DBLP:journals/corr/abs-2107-11098gan_ts_survey,DBLP:journals/jzusc/LinLLLG24diffusion_survey,yang2022adaptability}. Existing generative models often fail to balance these objectives. GAN-based approaches tend to reproduce high-frequency fluctuations at the cost of structural distortion, while diffusion-based models frequently oversmooth the signal, suppressing meaningful variability and reducing diversity~\cite{DBLP:conf/nips/YoonJS19, DBLP:conf/iclr/YuanQ24diffusionts}.  

This trade-off is especially problematic for noisy time series, which are common in real-world settings such as biomedical recordings and financial signals~\cite{DBLP:journals/corr/abs-2107-11098gan_ts_survey,DBLP:journals/corr/abs-2401-03006casparsurveydiff}. In such cases, directly matching the full data distribution can either amplify noise or erase critical temporal patterns, limiting both their interpretability and downstream utility~\cite{DBLP:conf/nips/YoonJS19}.

We take a structure-oriented perspective on time-series generation, motivated by a structure–residual decomposition of temporal data. Rather than attempting to reproduce all sample-level variability, we observe that many real-world applications primarily require samples that preserve global temporal organization, while allowing stochastic variation at the residual level. This distinction is particularly important for volatile and irregular signals, where direct distribution matching often leads to noise amplification or over-smoothing of meaningful temporal patterns.

To operationalize this insight, we represent temporal structure using a quantile-based transition graph, which compactly captures global distributional and relational statistics while suppressing sample-specific fluctuations. This structural representation defines a complementary modality to raw time-series signals, enabling a graph-to-time-series generation paradigm. According to this formulation, we propose a graph-to-time series model (\textbf{Graph2TS}) based on a quantile-graph-conditioned variational autoencoder that learns a conditional generative model of time series given structural graphs. Unlike prior conditional VAEs that rely on labels or metadata, our model conditions generation directly on structural information, enabling cross-modal generation that preserves global temporal structure while supporting controlled stochastic diversity.

Our main contributions are:




\begin{itemize}
    \item We propose a \emph{structure--residual perspective} for time-series generation,
    modeling temporal data as a deterministic structural backbone with stochastic residual variation,
    which clarifies the trade-off between structural fidelity and generative diversity.

    \item We show that a \emph{quantile-based transition graph} provides a compact structural representation of time series.
    Under a first-order Markov assumption, this representation captures essential global temporal and distributional dependencies,
    enabling a principled \emph{cross-modal} formulation from graphs to sequences.

    \item Based on this formulation, we propose \textbf{Graph2TS}, a quantile-graph conditioned variational autoencoder that models $p(x \mid g, z)$. Extensive experiments and ablation studies demonstrate that both structural conditioning and stochastic residual modeling are essential for achieving faithful, diverse, and stable time-series generation, showing an advantage over diffusion- and GAN-based baselines across diverse datasets.
\end{itemize}

\section{Related Work}

\subsection{Time Series Generation}

Generative modeling of time series has attracted significant attention in recent years. 
GAN-based approaches extend adversarial learning to sequential data, often by combining recurrent, temporal convolutional architectures or attention mechanisms with discriminator-based training objectives~\cite{DBLP:journals/cacm/GoodfellowPMXWO20,ahmed2023sparse}. Representative examples include TimeGAN and its variants, which integrate supervised losses to encourage temporal coherence~\cite{DBLP:conf/nips/YoonJS19}. Despite their success, GAN-based models are known to suffer from training instability, mode collapse, and sensitivity to noise, particularly when applied to short or highly volatile time series~\cite{DBLP:journals/corr/ArjovskyCB17wgan}. In practice, such models often reproduce local fluctuations at the cost of distorting global temporal structure~\cite{DBLP:journals/corr/EstebanHR17realvalue}. 

More recently, diffusion-based models have been adapted to time series generation~\cite{yang2024survey_diffusion_spatio,rasul2021autoregressive}. By progressively denoising samples from a learned noise process, these methods achieve strong likelihood-based modeling and improved training stability~\cite{DBLP:journals/jzusc/LinLLLG24diffusion_survey,DBLP:conf/iclr/YuanQ24diffusionts}. However, diffusion models typically exhibit a strong smoothing bias, which can suppress meaningful variability and lead to over-regularized outputs~\cite{DBLP:conf/nips/LiuY0H24_retreval_diffusionts}. In addition, the iterative sampling procedure is computationally expensive, and the learned dynamics may drift away from the dominant structural patterns of the data~\cite{DBLP:journals/jzusc/LinLLLG24diffusion_survey}. 

These approaches aim to model the \emph{full data distribution}, and while they achieve a certain degree of success, their design makes them sensitive to noise and sample-specific artifacts, especially in short or noisy time series.

\subsection{Structure-Aware and Conditional Generation}

To improve controllability, several works explore conditional generation for time series~\cite{coletta2023constrained,li2023causal}. Conditional GANs and related architectures incorporate auxiliary information such as labels, context variables, or handcrafted features to guide generation~\cite{DBLP:journals/corr/abs-2006-16477cgan,DBLP:journals/tkde/LuRWNN24mtgan,DBLP:journals/corr/abs-2006-05421sigwgan,DBLP:conf/icml/NarasimhanAASC24timeweaver}. Other approaches focus on learning compact representations through autoencoders or self-supervised objectives, which are then used for downstream generation or reconstruction tasks~\cite{DBLP:journals/pami/ZhangWZCJLZLPSP24_survey,DBLP:conf/aaai/YueWDYHTX22ts2vec}. 

While conditioning improves flexibility, most existing methods rely on \emph{raw features} or latent embeddings that entangle structural statistics with stochastic variations~\cite{DBLP:conf/iclr/OublalLBLR24diosc}. As a result, these models lack an explicit mechanism to separate global temporal structure from local noise, limiting their ability to control fidelity and diversity independently~\cite{DBLP:conf/iclr/OublalLBLR24diosc}.




\subsection{Graph-Based Representations for Time Series}

Graph-based representations provide an alternative view of time series by encoding temporal relationships as structured graphs~\cite{DBLP:journals/corr/abs-2510-01169jozefin,PhysRevE.80.046103hvg}.
Prior work includes statistical or topology-inspired constructions as well as deep-learning graph time-series modeling. These representations capture certain structural properties of time series, such as periodicity, state recurrence, or distributional characteristics~\cite{silva2023multilayerquantilegraphmultivariate,Zou_2019complextononlinear} and have been used for a wide range of tasks, but, to the best of our knowledge, no machine learning model was developed to use them for generative purposes~\cite{doi:10.1073/pnas.0709247105visibilitygraph,Silva_2021tsvianetwork,chen2024graph}.

In this work, we adopt a \emph{quantile-based graph} as a structured representation of time series~\cite{Campanharo2011DualityBT,7483302/2015quantile_graph}.
Unlike prior uses of graph representations that primarily serve descriptive or analytical purposes, we leverage the quantile graph as a conditioning signal within a generative model. By encoding distributional transitions across time rather than raw sample trajectories, this representation preserves global structural statistics while filtering out sample-specific noise, enabling structure-controlled synthesis of time series.
\section{Problem Statement}

\subsection{Setting}

We consider a dataset of univariate time series
$\{x_i\}_{i=1}^N$, where $x_i \in \mathbb{R}^T$.
Our goal is to generate synthetic samples $\tilde{x} \in \mathbb{R}^T$
that preserve the global temporal structure of the data
while maintaining non-degenerate variability.

Formally, a desirable generator should satisfy
\begin{equation}
\tilde{x} \sim q(x)
\quad \text{s.t.} \quad
\mathcal{S}(\tilde{x}) \approx \mathcal{S}(x),
\qquad
\mathrm{Var}(\tilde{x}) > 0,
\end{equation}
where $\mathcal{S}(\cdot)$ denotes a structural descriptor
that captures invariant temporal patterns
(e.g., wave shape or transition dynamics),
rather than pointwise signal values.

\subsection{Why Full Distribution Modeling Is Ill-Conditioned}

Most existing time-series generators aim to learn the full data distribution
$p(x)$ directly.
However, for highly volatile sequences,
this objective is ill-conditioned.
A time series can be conceptually decomposed as
\begin{equation}
x = x^{\mathrm{struct}} + \varepsilon,
\end{equation}
where $x^{\mathrm{struct}}$ captures global temporal organization
and $\varepsilon$ represents stochastic fluctuations or noise.

Learning $p(x)$ entangles structural variation with sample-specific noise.
In practice, this leads to a fundamental trade-off:
models either overfit $\varepsilon$ and amplify noise,
or oversmooth the signal and distort $x^{\mathrm{struct}}$.
Such behavior has been widely observed in both GAN-based and diffusion-based
time-series generators, particularly in the volatile sequence regime.

\subsection{Structure-Conditioned Reformulation}

To decouple structural information from stochastic variation,
we reformulate time-series generation as modeling a conditional distribution
\begin{equation}
p(x \mid s),
\qquad s = \mathcal{S}(x),
\end{equation}
where $s$ is a deterministic structural summary of the time series.
Under this formulation, stochastic variability is restricted to directions
that are consistent with the structural constraint $s$,
allowing generated samples to preserve global temporal patterns
while retaining controlled, non-degenerate diversity.

 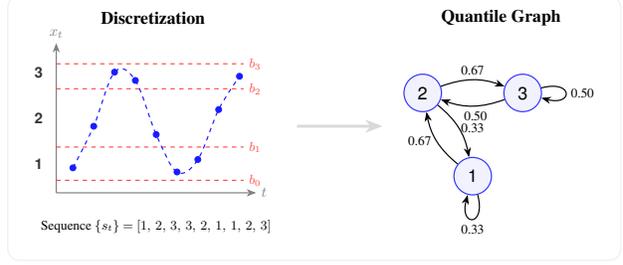
\begin{figure}[h!]
    \centering
    \resizebox{\linewidth}{!}{
        \begin{tikzpicture}[
    font=\rmfamily, 
    >=Stealth,
    state/.style={circle, draw=blue!80, fill=blue!5, thick, minimum size=0.9cm, inner sep=0pt, font=\sffamily\large},
    prob/.style={midway, font=\footnotesize, text=black} 
]

\begin{scope}[local bounding box=ts_scope]
    \draw[->, thick, gray] (0,0) -- (4.8,0) node[right] {$t$};
    \draw[->, thick, gray] (0,0) -- (0,3.6) node[above] {$x_t$};

    \draw[dashed, red!70, thick] (0,0.3) -- (4.5,0.3) node[right, text=red!70, font=\small] {$b_0$};
    \draw[dashed, red!70, thick] (0,1.1) -- (4.5,1.1) node[right, text=red!70, font=\small] {$b_1$};
    \draw[dashed, red!70, thick] (0,2.5) -- (4.5,2.5) node[right, text=red!70, font=\small] {$b_2$};
    \draw[dashed, red!70, thick] (0,3.1) -- (4.5,3.1) node[right, text=red!70, font=\small] {$b_3$};
    
    \node[black!80, font=\bfseries\sffamily, anchor=east] at (-0.2, 0.7) {1}; 
    \node[black!80, font=\bfseries\sffamily, anchor=east] at (-0.2, 1.8) {2}; 
    \node[black!80, font=\bfseries\sffamily, anchor=east] at (-0.2, 2.9) {3}; 

    \draw[blue!90, thick, dashed, smooth, tension=0.6] plot coordinates {
        (0.4, 0.6) (0.9, 1.6) (1.4, 2.9) (1.9, 2.7) 
        (2.4, 1.4) (2.9, 0.5) (3.4, 0.8) (3.9, 2.0) (4.4, 2.8)
    };

    \foreach \p in {
        (0.4, 0.6), (0.9, 1.6), (1.4, 2.9), (1.9, 2.7), 
        (2.4, 1.4), (2.9, 0.5), (3.4, 0.8), (3.9, 2.0), (4.4, 2.8)
    } {
        \filldraw[blue!90] \p circle (2pt);
    }
    
    \node[align=center, font=\small] (seq_label) at (2.4, -0.8) 
    {Sequence $\{s_t\}$ $=$ $[1,\, 2,\, 3,\, 3,\, 2,\, 1,\, 1,\, 2,\, 3]$};
\end{scope}

\begin{scope}[shift={(10.0, 1.6)}, local bounding box=graph_scope]
    \node[state] (s1) at (0, -1.2) {1};
    \node[state] (s2) at (-1.2, 0.8) {2};
    \node[state] (s3) at (1.2, 0.8) {3};

    \draw[->, thick] (s1) to [loop below, min distance=8mm] node[prob, below] {0.33} (s1);
    \draw[->, thick, bend left=20] (s1) to node[prob, left] {0.67} (s2);

    \draw[->, thick, bend left=20] (s2) to node[prob, right, xshift=-2pt] {0.33} (s1);
    \draw[->, thick, bend left=20] (s2) to node[prob, above] {0.67} (s3);
    
    \draw[->, thick] (s3) to [loop right, min distance=8mm] node[prob, right] {0.50} (s3);
    \draw[->, thick, bend left=20] (s3) to node[prob, below, xshift=2pt] {0.50} (s2); 

\end{scope}

\coordinate (top_line) at (0, 4.2); 
\node[font=\bfseries\large] at (ts_scope.center |- top_line) {Discretization};
\node[font=\bfseries\large] at (graph_scope.center |- top_line) {Quantile Graph};

\coordinate (arrow_y) at (0, 1.6);
\draw[->, line width=2pt, gray!30] 
    (ts_scope.east |- arrow_y) ++(0.5, 0) -- ($(graph_scope.west |- arrow_y) + (-0.5, 0)$);

\begin{scope}[on background layer]
    \node[draw=gray!15, rounded corners=8pt, fill=white, 
          fit=(ts_scope) (graph_scope) (seq_label) (top_line), 
          inner sep=15pt] {};
\end{scope}

\end{tikzpicture} 
    }
    \caption{Example of time-series to Quantile Graph mapping. Left: The time series is discretized into states based on quantile boundaries. Right: The resulting Markov transition dynamics represented as a graph.}
    \label{fig:quantile_graph}
\end{figure}

\subsection{Quantile Graph as Structural Representation}

To instantiate the structural summary operator $\mathcal{S}(\cdot)$,
we seek a representation that (i) is invariant to local amplitude noise,
(ii) preserves global temporal organization,
and (iii) admits a fixed-size, sample-aligned form for conditional modeling.
We therefore adopt a \emph{quantile graph}, which provides a compact
representation of temporal dynamics by encoding transition statistics
between globally shared quantile states.

Given a univariate time series $\{x_t\}_{t=1}^T$,
we first estimate a set of globally shared quantile boundaries
$\mathcal{B} = \{b_0,\dots,b_Q\}$ from the training data.
Each observation is mapped to a discrete state
\begin{equation}
s_t = q(x_t),
\qquad
q(x) = k \;\; \text{if } x \in [b_{k-1}, b_k),
\end{equation}
yielding a discrete stochastic process $\{s_t\}$ over $Q$ states.
This discretization marginalizes fine-grained amplitude variations
while preserving the relative ordering of signal values.

From the resulting state sequence, we estimate empirical transition probabilities
\begin{equation}
P_{ij}
=
\mathbb{P}(s_{t+1}=j \mid s_t=i)
\approx
\frac{\sum_{t=1}^{T-1} \mathbf{1}[s_t=i,\, s_{t+1}=j]}
     {\sum_{t=1}^{T-1} \mathbf{1}[s_t=i]},
\end{equation}
which define a first-order Markov chain over quantile states.
The transition matrix $P \in \mathbb{R}^{Q\times Q}$ is treated as a weighted
directed graph, referred to as the \emph{quantile graph}.

This construction induces the abstraction
\[
\{x_t\}
\;\longrightarrow\;
\{s_t\}
\;\longrightarrow\;
P(s_{t+1}\mid s_t),
\]
which marginalizes pointwise noise while preserving invariant transition
statistics of the underlying dynamics.
Globally shared quantile bins ensure semantic alignment across samples,
yielding a fixed-size and well-conditioned structural representation
for conditional generation.

Figure~\ref{fig:quantile_graph} illustrates the process of quantile graph construction.

\begin{figure*}[t!]
    \centering
    \resizebox{0.95\linewidth}{!}{%
        \begin{tikzpicture}[
    node distance=1.5cm and 1.5cm,
    font=\sffamily\small,
    process/.style={rectangle, draw=blue!50!black, fill=blue!5, thick, minimum width=2.5cm, minimum height=1cm, rounded corners=2pt, align=center},
    tensor/.style={rectangle, draw=green!50!black, fill=green!5, thick, minimum width=1.5cm, minimum height=0.8cm, align=center, dashed},
    input/.style={cylinder, shape border rotate=90, draw=black, fill=gray!10, aspect=0.25, minimum height=1.2cm, minimum width=2cm, align=center},
    g_input/.style={rectangle, shape border rotate=90, draw=black, fill=gray!10, aspect=0.25, minimum height=1.2cm, minimum width=2cm, align=center},
    loss/.style={circle, draw=red!70!black, fill=red!10, thick, minimum size=1.2cm, align=center, inner sep=0pt},
    connection/.style={-Latex, thick, draw=black!70},
    skip/.style={-Latex, thick, draw=blue!70, dashed}
]

    \node[input] (ts_in) {TS Input\\$[B, 32]$};
    \node[g_input, below=2.5cm of ts_in] (gr_in) {Graph Input\\$[B, 100]$};

    \node[process, right=of ts_in] (ts_enc) {TS Encoder\\(MLP)};
    \node[process, right=of gr_in] (gr_enc) {Graph Encoder\\(MLP)};

    \draw[connection] (ts_in) -- (ts_enc);
    \draw[connection] (gr_in) -- (gr_enc);

    \node[tensor, right=of ts_enc] (t_raw) {$t_{raw}$\\$[B, 128]$};
    \node[tensor, right=of gr_enc] (g_raw) {$g_{raw}$\\$[B, 128]$};

    \draw[connection] (ts_enc) -- (t_raw);
    \draw[connection] (gr_enc) -- (g_raw);

    \node[process, below=0.5cm of t_raw, minimum width=1.5cm, minimum height=0.6cm, fill=yellow!10, draw=orange!50] (norm_t) {Norm};
    \node[process, above=0.5cm of g_raw, minimum width=1.5cm, minimum height=0.6cm, fill=yellow!10, draw=orange!50] (norm_g) {Norm}; 
    
    \node[loss, below=0.8cm of ts_enc] (align_loss) {Align};

    \draw[connection] (t_raw) -- (norm_t);
    \draw[connection] (g_raw) -- (norm_g); 

    \draw[connection] (norm_t.west) to[out=180, in=0] (align_loss.east);
    \draw[connection] (norm_g.west) to[out=180, in=0] (align_loss.east);

    \coordinate[right=1cm of t_raw] (concat_pt_top);
    \coordinate[right=1cm of g_raw] (concat_pt_bot);
    
    \node[process, right=1.5cm of t_raw, yshift=-1.5cm] (posterior) {Posterior\\$q(z|ts, gr)$};
    
    \draw[connection] (t_raw.east) to[out=0,in=180] (posterior.west);
    \draw[connection] (g_raw.east) to[out=0,in=180] (posterior.west);

    \node[tensor, right=of posterior] (params) {$\mu, \log\sigma^2$};
    \node[loss, above=0.5cm of params, scale=0.8] (kl_loss) {KL\\Div};
    
    \draw[connection] (posterior) -- (params);
    \draw[connection] (params) -- (kl_loss);

    \node[circle, draw, fill=white, inner sep=2pt, right=1cm of params] (plus) {$+$};
    \node[above=0.5cm of plus] (noise) {$\epsilon \sim \mathcal{N}(0,I)$};
    \draw[connection] (noise) -- (plus);
    \draw[connection] (params) -- (plus) node[midway, above, font=\tiny] {$\sigma$};
    \draw[connection] (params) to[out=-20, in=180] node[midway, below, font=\tiny] {$\mu$} (plus);

    \node[tensor, right=0.5cm of plus] (z) {$z$\\Latent};
    \draw[connection] (plus) -- (z);

    \node[process, right=1.5cm of z, fill=purple!5, draw=purple!50!black] (decoder) {Decoder\\$p(ts|gr, z)$};
    
    \draw[connection] (z) -- (decoder);

    \draw[connection] (g_raw) to[out=0, in=210] (decoder.west);


    \node[input, right=of decoder] (ts_out) {$\hat{TS}$\\Output};
    
    \node[loss, above=0.4cm of ts_out, scale=1] (recon_dist_loss) {Recon \\ \& Dist};

    \draw[connection] (decoder) -- (ts_out);
    \draw[connection] (ts_out) -- (recon_dist_loss);
    \draw[connection] (ts_in.east) to[out=21,in=183] (recon_dist_loss.west);

    \begin{scope}[on background layer]
        \node[fit=(ts_in)(recon_dist_loss)(g_raw), fill=white, draw=black!20, rounded corners=10pt, inner sep=15pt] (bg) {};
    \end{scope}

\end{tikzpicture}
    }
    \caption{The Architecture of Our Proposed Graph2TS}
    \label{fig:graph2ts}
\end{figure*}
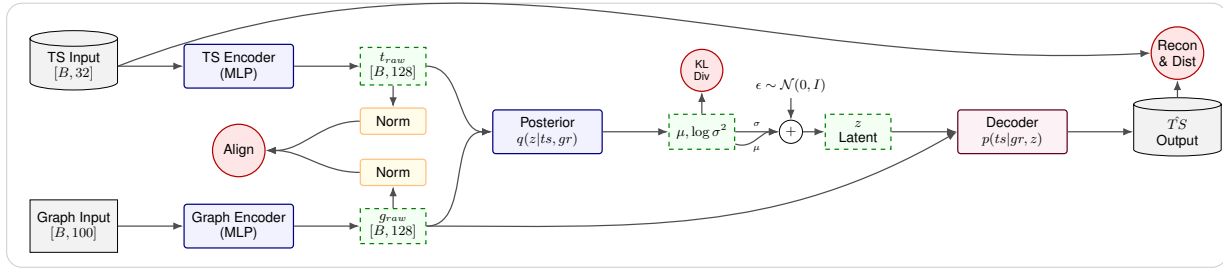
\section{Graph2TS: Structure-Conditioned Time Series Generation}

Building on the structure-conditioned formulation introduced in the previous section,
we developed Graph2TS, a conditional generative deep learning architecture. Graph2TS conditions a variational autoencoder on a quantile-based graph representation, which encodes global transition dynamics, while a latent variable captures residual stochastic variability.

\subsection{Graph2TS Architecture}


We illustrate the proposed Graph2TS architecture in Figure~\ref{fig:graph2ts}. Each input time series $x \in \mathbb{R}^T$ is first mapped to a quantile graph $g = \phi(x)$, which serves as a fixed-size structural condition. The graph is encoded into a continuous representation and combined with a latent variable to generate time-series samples. This design explicitly separates global structural information, captured by the graph, from stochastic residual variation, modeled by the latent variable. The implications of this structure-conditioned design are analyzed in the following sections.

\subsection{Generative Process}

Let $X \in \mathbb{R}^T$ denote a time-series random variable drawn from
$p_{\mathrm{data}}(x)$.
We define a deterministic structural encoder
\begin{equation}
G = \phi(X),
\end{equation}
where $\phi(\cdot)$ maps a time series to its quantile graph.

Given $G$, we introduce a latent variable $Z \in \mathbb{R}^d$
to model variability not captured by the structural representation:
\begin{align}
Z &\sim p(z) = \mathcal{N}(0, I), \\
X &\sim p_\theta(x \mid G, Z).
\end{align}

The resulting conditional distribution is
\begin{equation}
p_\theta(x \mid g)
=
\int p_\theta(x \mid g, z)\,p(z)\,dz,
\end{equation}
which allows multiple distinct samples to be generated
from the same structural graph.

\subsection{Learning Objective}

For our \textbf{Graph2TS}, we adopt a conditional variational autoencoder formulation,
where generation is conditioned on a structural representation $g$.
Let $q_\phi(z \mid x, g)$ denote the variational posterior.
The core learning objective is the conditional evidence lower bound (ELBO):
\begin{equation}
\begin{aligned}
\mathcal{L}_{\mathrm{ELBO}}
= {} &
\mathbb{E}_{q_\phi(z \mid x, g)}
\big[
\log p_\theta(x \mid g, z)
\big] \\
& -
\mathrm{KL}\big(
q_\phi(z \mid x, g)\,\|\,p(z)
\big).
\end{aligned}
\end{equation}

In practice, the conditional likelihood term is instantiated as a
mean-squared reconstruction loss, and the KL term is weighted by a
coefficient $\beta$ to control the strength of stochastic regularization.
The resulting training objective is augmented with additional
structure-preserving regularizers:
\begin{equation}
\begin{aligned}
\label{eq:full_objective}
\min_{\theta,\phi}\;
\mathcal{L}
=
\lambda_{\mathrm{rec}}\,\mathcal{L}_{\mathrm{rec}}
+
\beta\,\mathrm{KL}\big(q_\phi(z \mid x, g)\,\|\,p(z)\big) \\
+
\lambda_{\mathrm{align}}\,\mathcal{L}_{\mathrm{align}}
+
\lambda_{\mathrm{dist}}\,\mathcal{L}_{\mathrm{dist}} .
\end{aligned}
\end{equation}

The alignment loss $\mathcal{L}_{\mathrm{align}}$ is implemented as a
bidirectional InfoNCE objective that enforces consistency between
graph and time-series embeddings in a shared latent space.
This term improves structural correspondence but does not introduce
additional stochasticity into the generative process.

The distribution regularization term $\mathcal{L}_{\mathrm{dist}}$
matches the order statistics of generated and real sequences,
encouraging consistency in marginal amplitude distributions.
This stabilizes generation and mitigates distributional drift,
while preserving the conditional diversity provided by the latent variable $z$.

\subsection{Interpretation of Structure-Conditioned Generation}
\label{sec:structure_effect}

While the problem formulation motivates conditioning on structure, this section explains how such conditioning affects the learned generative behavior. We briefly interpret how conditioning on a structural representation $G$ shapes the behavior of Graph2TS.

This behavior can be decomposed as follows.

\begin{proposition}[Structure--residual decomposition]
Given a time-series random variable $X \in \mathbb{R}^T$ and its
deterministic structural representation $G=\phi(X)$,
define
\begin{equation}
f(G) \triangleq \mathbb{E}[X \mid G],
\qquad
R \triangleq X - f(G).
\end{equation}
Then $X$ admits the decomposition
\begin{equation}
X = f(G) + R,
\end{equation}
where the residual satisfies
\begin{equation}
\mathbb{E}[R \mid G] = 0.
\end{equation}
\end{proposition}

The function $f(G)$ represents a \emph{structural backbone}:
The underlying trajectory is consistent with the structural summary $G$.
The residual term $R$ captures admissible variations around this backbone,
including stochastic fluctuations that are not determined by $G$.

Unconditional generative models must simultaneously learn both
$f(G)$ and the residual distribution of $R$ from data,
which is difficult for noisy time series.
In contrast, conditioning on $G$ fixes the backbone $f(G)$
and restricts stochastic generation to the residual component.

This yields the following variance decomposition:
\begin{equation}
\mathrm{Var}(X)
=
\mathrm{Var}\!\left(\mathbb{E}[X \mid G]\right)
+
\mathbb{E}\!\left[\mathrm{Var}(X \mid G)\right],
\end{equation}
where the first term corresponds to structural variability
and the second to conditional stochastic variation.
By explicitly conditioning on $G$,
the proposed model controls these two sources of variability separately. A detailed derivation of this decomposition is provided in Appendix~\ref{sec:formalua_var}.

In the proposed \textbf{Graph2TS}, the decoder parameterizes a conditional distribution $p_\theta(x \mid g, z)$. This can be interpreted as
\begin{equation}
x = f_\theta(g) + r_\theta(g, z),
\end{equation}
where $f_\theta(g)$ approximates $\mathbb{E}[X \mid G=g]$ and $r_\theta(g,z)$ models the residual variation with
$\mathbb{E}_z[r_\theta(g,z)\mid g]\approx 0$.

As a result, structural conditioning suppresses noise-induced distortion while preserving global temporal patterns, and the latent variable $z$ reintroduces controlled diversity without violating structural consistency.

\section{Experiments}

\subsection{Setup}
\label{sec:exp-setup}

\textbf{Baselines.}
We compare against TimeGAN~\cite{DBLP:conf/nips/YoonJS19} and DiffusionTS~\cite{DBLP:conf/iclr/YuanQ24diffusionts}, which are the dominant paradigms for time-series generation. These baselines provide a rigorous benchmark for evaluating both distributional fidelity and temporal coherence. 

\textbf{Datasets.}
We evaluate the proposed method on four benchmark time-series datasets:
(i) the CHB--MIT Scalp EEG Database (\textbf{CHB-MIT})~\cite{PhysioNet-chbmit-1.0.0},
(ii) the Daily Sunspot Number dataset (\textbf{Sunspot})~\cite{sunspot},
(iii) the Electricity Load dataset (\textbf{Electricity})~\cite{electricityloaddiagrams20112014_321},
and (iv) the MIT-BIH Arrhythmia ECG dataset (\textbf{ECG})~\cite{932724_mitbih}.
Additional dataset details are provided in Appendix~\ref{sec-appendix-datasets}.

\textbf{Metrics.}
We assess generative performance using a compact set of metrics that capture \emph{distributional fidelity}, \emph{temporal structure}, and \emph{manifold coverage}. Distributional alignment is evaluated using the Wasserstein distance and the Kolmogorov--Smirnov (KS) statistic. Temporal consistency is measured via discrepancies in autocorrelation functions (ACF) and power spectral densities (PSD). To assess representativeness, we report prototype-based metrics including nearest-neighbor prototype error and coverage at fixed distance thresholds. Unlike discriminative or downstream-task metrics, which are task- and classifier-dependent, our metrics directly assess whether the generator matches the target distribution and dynamics. Formal metric definitions are deferred to Appendix~\ref{sec-appendix-metrics}.


\subsection{Overall Generation Quality}
\label{sec:overall_quality}
\begin{table*}[t]
\centering
\footnotesize 
\setlength{\tabcolsep}{2.5pt} 
\caption{Comparison of distributional fidelity, temporal structure, and representativeness across datasets. 
Metrics: \textbf{Wass}: Wasserstein distance, \textbf{KS}: KS statistic, \textbf{ACF}: ACF MAE, \textbf{PSD}: PSD L2 distance, \textbf{P-Err}: Prototype Error, \textbf{MDR}: Medoid Distance Ratio, \textbf{Cov}: Manifold Coverage. 
($\downarrow$: Lower is better, $\uparrow$: Higher is better).}
\label{tab:overall_metrics}
\begin{tabular}{l l c c c c c c c c c}
\toprule
\multirow{2}{*}{\textbf{Dataset}} & \multirow{2}{*}{\textbf{Model}} 
& \multicolumn{2}{c}{\textbf{Distribution} ($\downarrow$)} 
& \multicolumn{2}{c}{\textbf{Temporal} ($\downarrow$)} 
& \multicolumn{3}{c}{\textbf{Representativeness} ($\downarrow$)} 
& \multicolumn{2}{c}{\textbf{Coverage} ($\uparrow$)} \\
\cmidrule(lr){3-4} \cmidrule(lr){5-6} \cmidrule(lr){7-9} \cmidrule(lr){10-11}
& & \textbf{Wass} & \textbf{KS} & \textbf{ACF} & \textbf{PSD} & \textbf{P-Err (avg)} & \textbf{P-Err (med)} & \textbf{MDR} & \textbf{Cov@0.5} & \textbf{Cov@0.9} \\
\midrule

\multirow{3}{*}{\textbf{CHB-MIT}} 
& \textbf{Graph2TS (Ours)} & \textbf{5.07e-6} & \textbf{0.024} & \textbf{0.039} & \textbf{9.36e-10} & \textbf{2.329} & \textbf{2.098} & \textbf{0.561} & \textbf{0.987} & \textbf{0.993} \\
& DiffusionTS          & 1.10e-5          & 0.153          & 0.085          & 4.12e-9           & 2.647          & 2.231          & 0.637          & 0.972          & 0.988 \\
& TimeGAN              & 1.29e-5          & 0.134          & 0.085          & 3.81e-9           & 3.404          & 3.058          & 0.780          & 0.977          & 0.989 \\
\midrule

\multirow{3}{*}{\textbf{Sunspot}} 
& \textbf{Graph2TS (Ours)} & \textbf{2.329} & \textbf{0.113} & 0.077          & \textbf{8.02e2} & 4.546          & 3.744          & 0.513          & 0.828          & 0.965 \\
& DiffusionTS          & 35.49         & 0.320          & \textbf{0.021} & 3.74e3          & \textbf{3.765} & \textbf{3.307} & \textbf{0.366} & \textbf{0.929} & \textbf{0.995} \\
& TimeGAN              & 14.74         & 0.190          & 0.098          & 1.05e3          & 5.108          & 4.613          & 0.755          & 0.806          & 0.965 \\
\midrule

\multirow{3}{*}{\textbf{Electricity}} 
& \textbf{Graph2TS (Ours)} & 7.710          & 0.073          & 0.043          & \textbf{6.13e2} &4.369 & \textbf{3.540} & \textbf{0.427} & \textbf{0.946} & \textbf{0.998} \\
& DiffusionTS          & \textbf{4.030} & \textbf{0.059} & \textbf{0.027} & 3.55e3          &\textbf{3.793}           & 3.963          & 0.542          & 0.889          & 0.990 \\
& TimeGAN              & 21.11         & 0.235          & 0.064          & 2.16e3          & 4.951          & 4.523          & 0.730          & 0.860          & 0.960 \\
\midrule

\multirow{3}{*}{\textbf{ECG}} 
& \textbf{Graph2TS (Ours)} & \textbf{0.014} & \textbf{0.033} & \textbf{0.051} & \textbf{5.92e-3} & \textbf{4.268} & \textbf{3.760} & \textbf{0.498} & \textbf{0.895} & \textbf{0.961} \\
& DiffusionTS          & 0.114          & 0.201          & 0.147          & 0.198            & 596.0        & 596.3        & 549.9         & 0.000          & 0.000 \\
& TimeGAN              & 0.101          & 0.282          & 0.172          & 0.108            & 249.6        & 250.0        & 135.6         & 0.000          & 0.000 \\
\bottomrule
\end{tabular}
\end{table*}
Table~\ref{tab:overall_metrics} provides a unified evaluation of generation quality under equal sample size, covering distributional fidelity, temporal structure, representativeness, and manifold coverage across all datasets. Overall, the proposed structure-conditioned \textbf{Graph2TS} demonstrates strong and well-balanced performance across all criteria and metrics.

In terms of \emph{distributional fidelity}, our method achieves the lowest Wasserstein distance and KS statistic on CHB-MIT, ECG, and Sunspot, and remains competitive on Electricity.
These results indicate accurate alignment with the real marginal distributions, despite the absence of adversarial training.
For \emph{temporal structure}, the proposed model attains low ACF MAE and PSD discrepancy across datasets, demonstrating faithful preservation of temporal dependencies and dominant spectral characteristics.
Compared to diffusion-based models, which tend to oversmooth sharp dynamics, and GAN-based models, which often distort frequency content, our approach achieves a more balanced temporal representation.

Beyond distributional and temporal metrics, the proposed method consistently
exhibits strong \emph{representativeness} and \emph{manifold coverage}. Prototype-based errors are among the lowest across CHB-MIT, Electricity, and ECG, indicating that real samples are well approximated by nearby synthetic instances.
At the same time, high coverage scores show that the generated samples broadly span the support of the real data manifold, avoiding both mode collapse and excessive dispersion.
Notably, on challenging biomedical signals such as ECG, the proposed method maintains non-zero coverage and stable prototype distances, whereas competing methods fail to adequately cover the real data distribution.

Taken together, these results demonstrate that conditioning generation on quantile-based structural representations enables a favorable trade-off between fidelity, diversity, and stability.
The proposed \textbf{Graph2TS} consistently produces time series that are distributionally accurate, temporally coherent, and representative of the underlying data manifold across diverse domains.

These quantitative findings are further supported by qualitative visualizations of temporal and distributional structure.
As shown in Appendix~\ref{sec:appendix_qualitative_analysis}, mean ACF and PSD curves demonstrate close alignment between real and generated signals across datasets, while t-SNE embeddings indicate strong overlap of real and synthetic manifolds without collapse or excessive dispersion. Additional tables and visualizations supporting these results are deferred to Appendix~\ref{sec:appendix-examples},


\textbf{Limitations of Baseline Models on Heavy Tailed Signals}
We observe markedly different behavior of diffusion- and GAN-based models on Sunspot and ECG.
Sunspot signals are dominated by smooth, low-frequency, and weakly heavy-tailed dynamics, which align well with the Gaussian perturbation and distribution-matching assumptions underlying DifussionTS and TimeGAN.
In this regime, directly modeling the full data distribution is relatively well-conditioned, explaining their competitive or superior performance on Sunspot.

As empirically demonstrated in Appendix~\ref{sec:appendix-sunspot-ecg} and Figure~\ref{fig:sunspot_ecg_hist}, ECG signals exhibit extremely heavy tails in their increments, which fundamentally violates the Gaussian assumption of diffusion models, leading to over-smoothing, spectral distortion, or mode collapse.
By conditioning generation on a quantile-based transition graph, Graph2TS emphasizes global structural relations rather than raw amplitude density, making it more robust to heavy-tailed and highly variable dynamics.
In contrast, Graph2TS achieves performance comparable to DifussionTS and TimeGAN models on smooth signals such as Sunspot, while substantially outperforming them on challenging biomedical signals such as ECG.

\section{Ablation Study}

\begin{table*}[t!]
\centering
\caption{Ablation study on ECG dataset.
We compare the full model with (i) identity graph conditioning (w/o graph)
and (ii) a deterministic Graph2TS model without stochastic latent variables.
Lower is better ($\downarrow$), higher is better ($\uparrow$).}
\label{tab:ablation_ecg_structure}
\begin{tabular}{lcccccc}
\toprule
\textbf{Model}
& \textbf{Wass.} $\downarrow$
& \textbf{KS} $\downarrow$
& \textbf{ACF MAE} $\downarrow$
& \textbf{PSD L2} $\downarrow$
& \textbf{ProtoErr} $\downarrow$
& \textbf{Coverage@$\tau_{0.5}$} $\uparrow$ \\
\midrule

\textbf{Full (Graph + Graph2TS)}
& \textbf{0.014}
& \textbf{0.033}
& \textbf{0.051}
& \textbf{0.0059}
& \textbf{4.27}
& \textbf{0.895} \\

w/o graph 
& 0.049
& 0.125
& 0.088
& 0.0179
& 4.41
& 0.875 \\

w/o stochasticity 
& 0.078
& 0.228
& 0.084
& 0.0310
& 5.26
& 0.752 \\

\bottomrule
\end{tabular}
\end{table*}

\begin{figure*}[t!]
    \centering

    \resizebox{0.9\textwidth}{!}{
    \begin{minipage}{\textwidth}   

        \newcommand{\rowlabelwidth}{0.04\linewidth}
        \newcommand{\subfigcontainer}{0.95\linewidth}
        \newcommand{\subfigwidth}{0.31\linewidth}

        \begin{minipage}[c]{\rowlabelwidth}
            \centering
            \rotatebox{90}{\textbf{\large ACF}}
        \end{minipage}%
        \hfill
        \begin{minipage}[c]{\subfigcontainer}
            \begin{subfigure}{\subfigwidth}
                \centering
                \includegraphics[width=\linewidth]{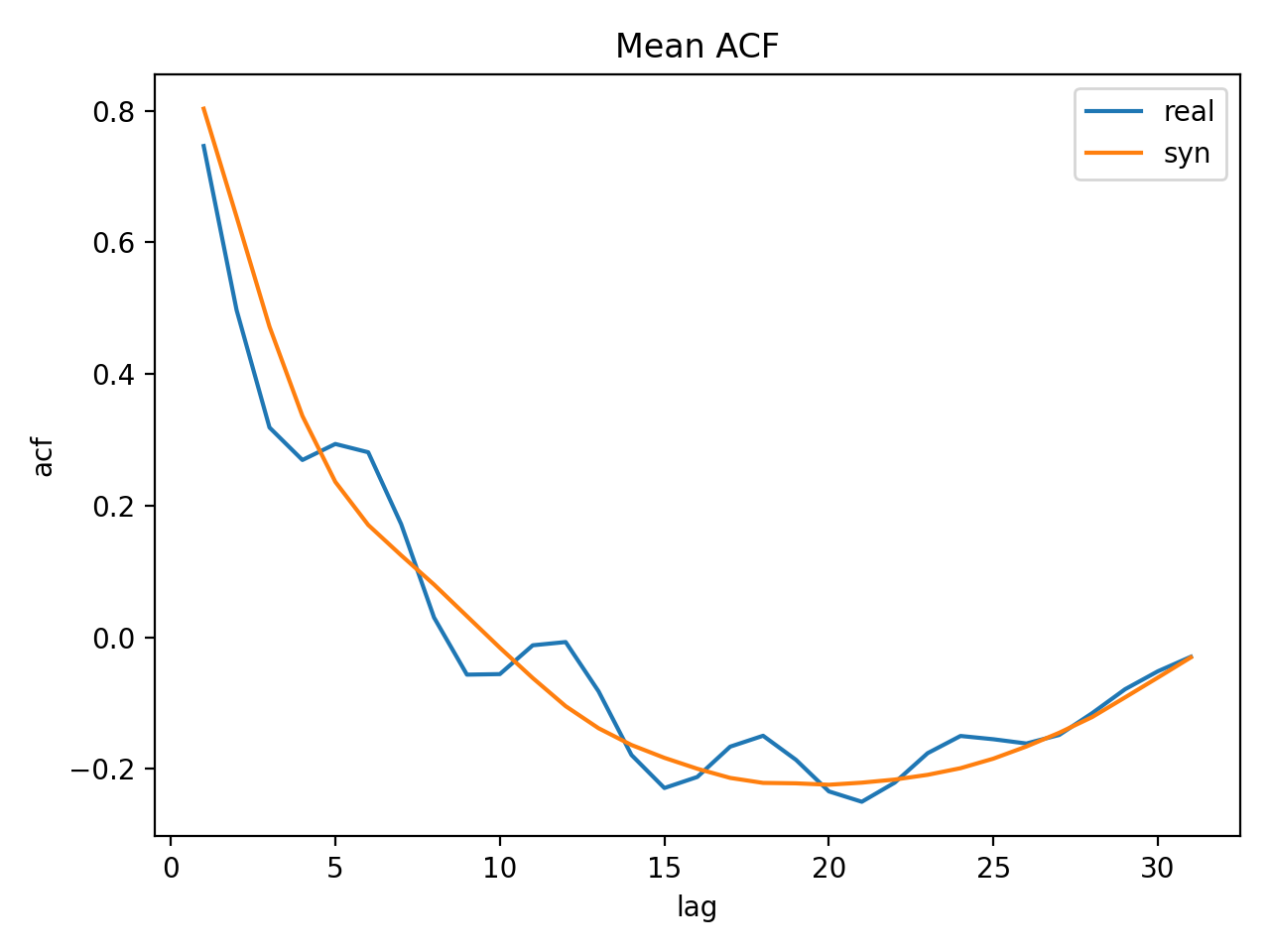}
                \caption{}
                \label{fig:acf_full}
            \end{subfigure}
            \hfill
            \begin{subfigure}{\subfigwidth}
                \centering
                \includegraphics[width=\linewidth]{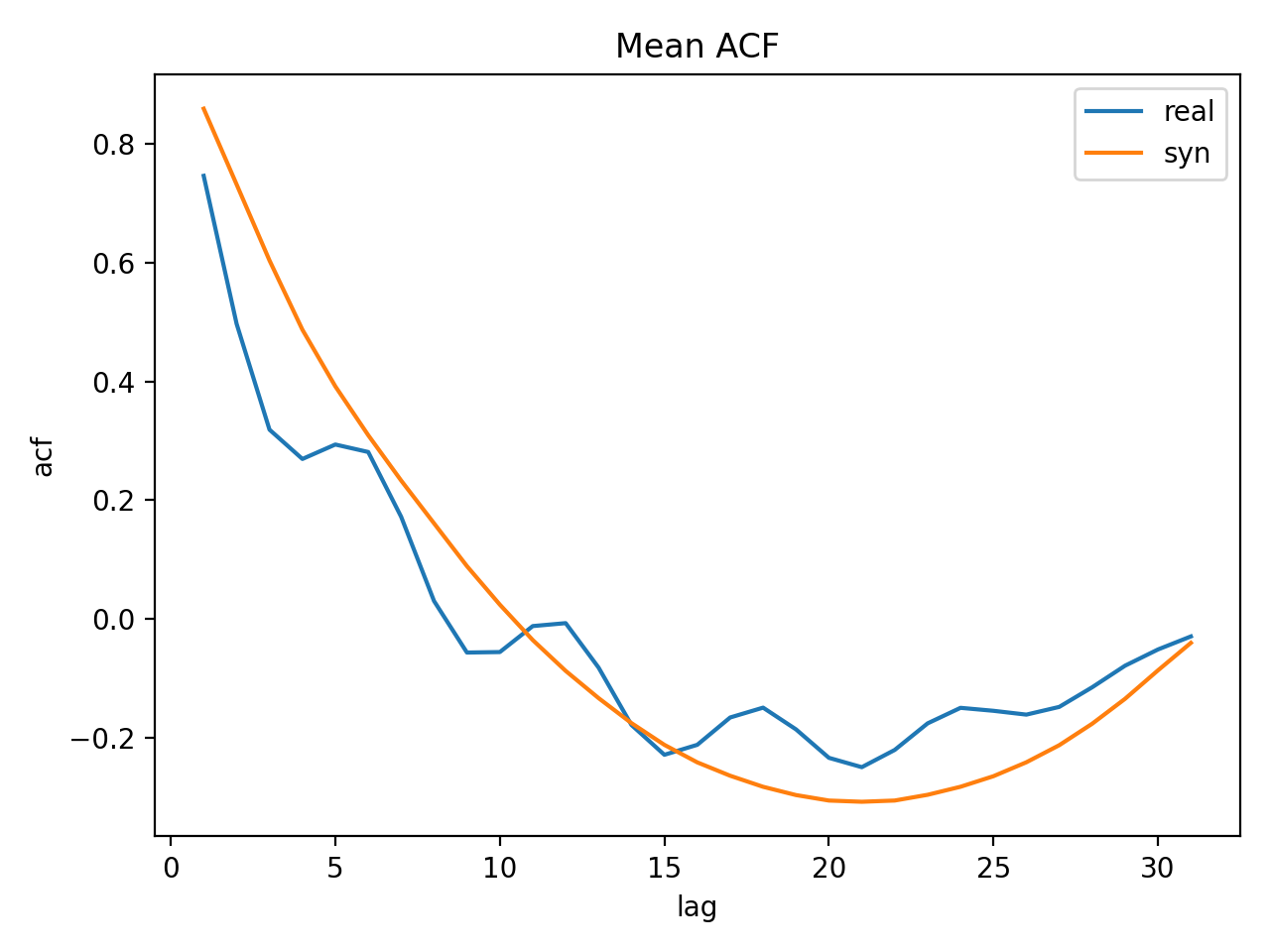}
                \caption{}
                \label{fig:acf_no_graph}
            \end{subfigure}
            \hfill
            \begin{subfigure}{\subfigwidth}
                \centering
                \includegraphics[width=\linewidth]{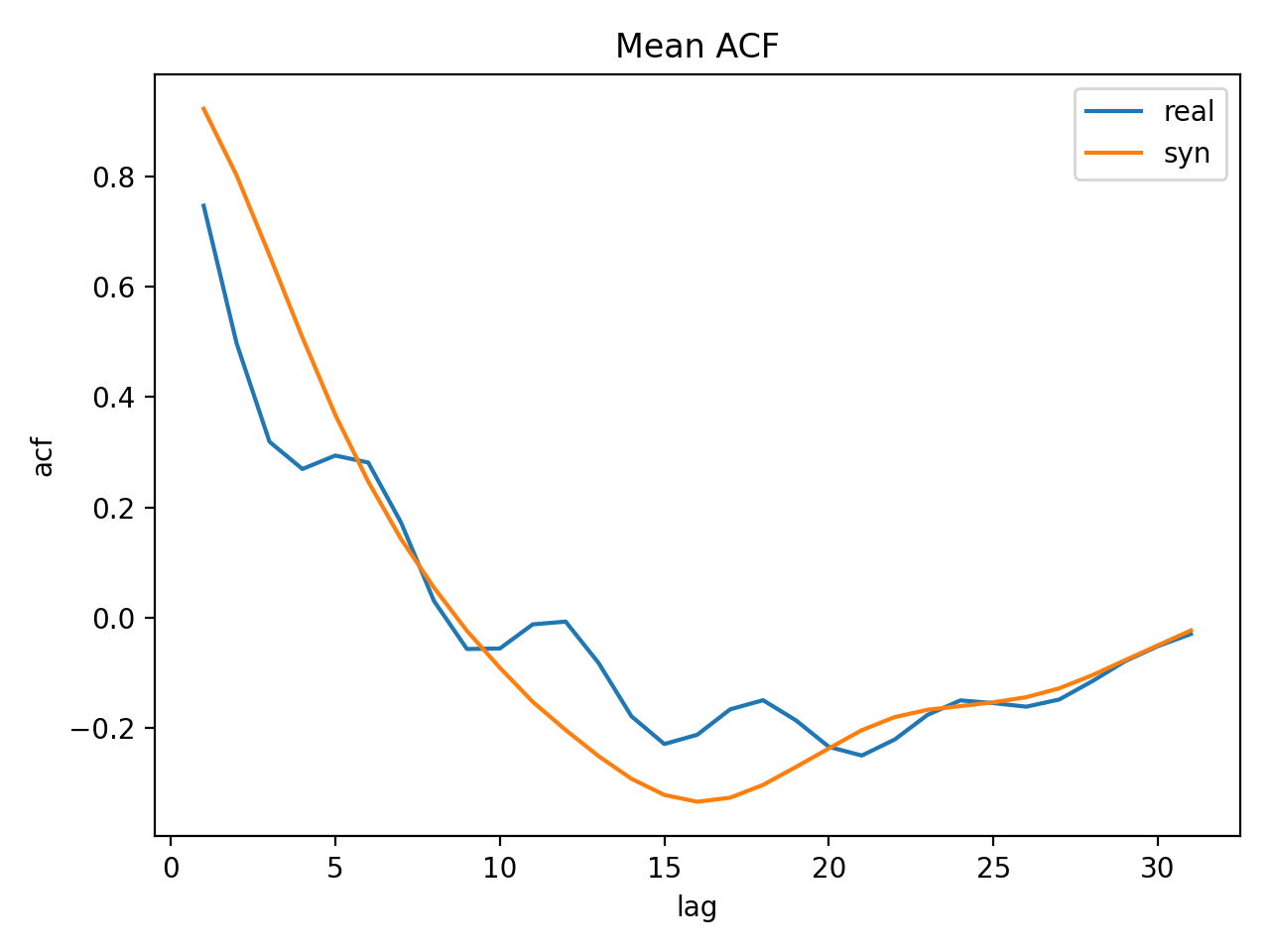}
                \caption{}
                \label{fig:acf_no_stoch}
            \end{subfigure}
        \end{minipage}

        \vspace{0.6em} 

        \begin{minipage}[c]{\rowlabelwidth}
            \centering
            \rotatebox{90}{\textbf{\large PSD}}
        \end{minipage}%
        \hfill
        \begin{minipage}[c]{\subfigcontainer}
            \begin{subfigure}{\subfigwidth}
                \centering
                \includegraphics[width=\linewidth]{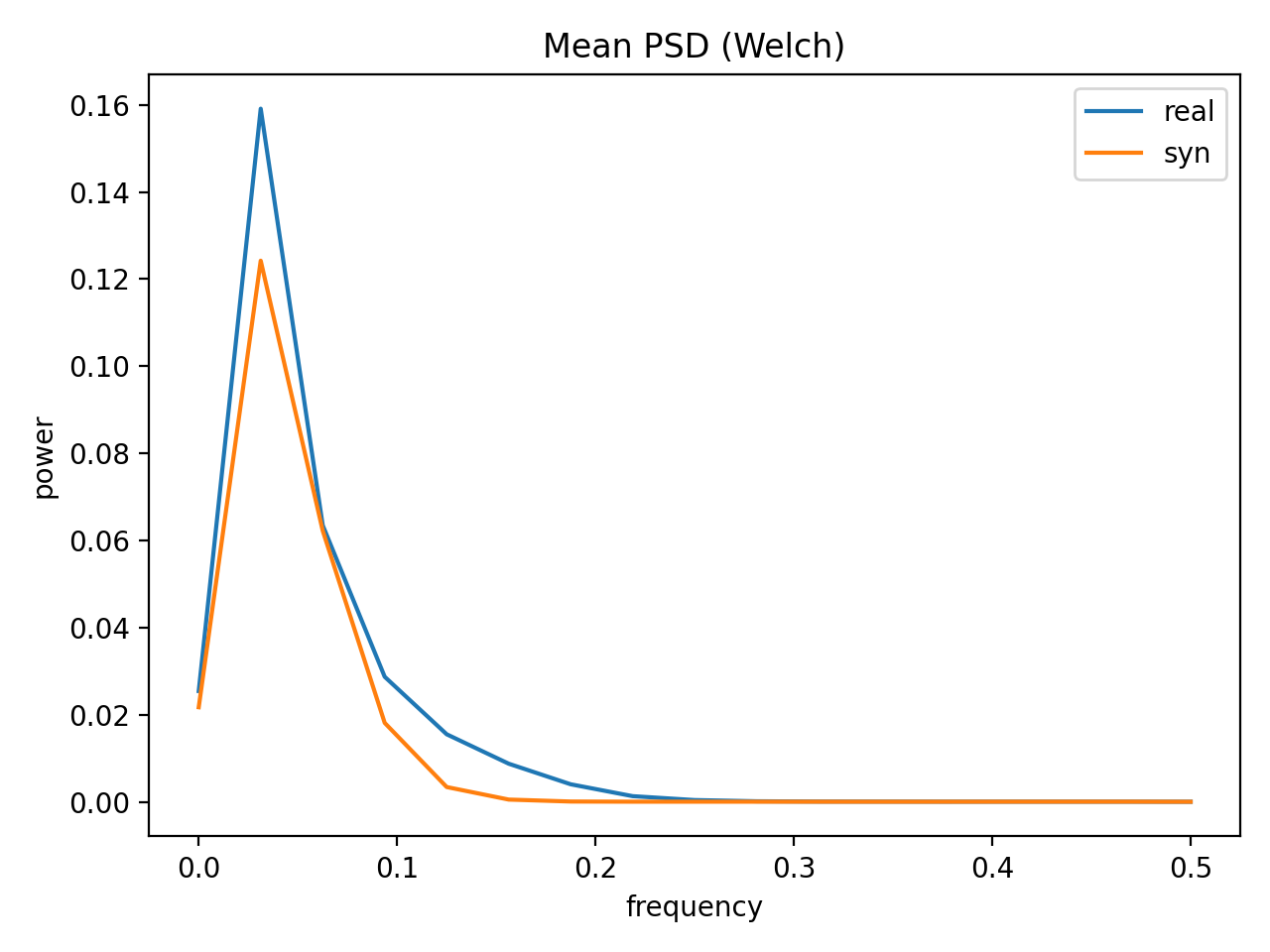}
                \caption{}
                \label{fig:psd_full}
            \end{subfigure}
            \hfill
            \begin{subfigure}{\subfigwidth}
                \centering
                \includegraphics[width=\linewidth]{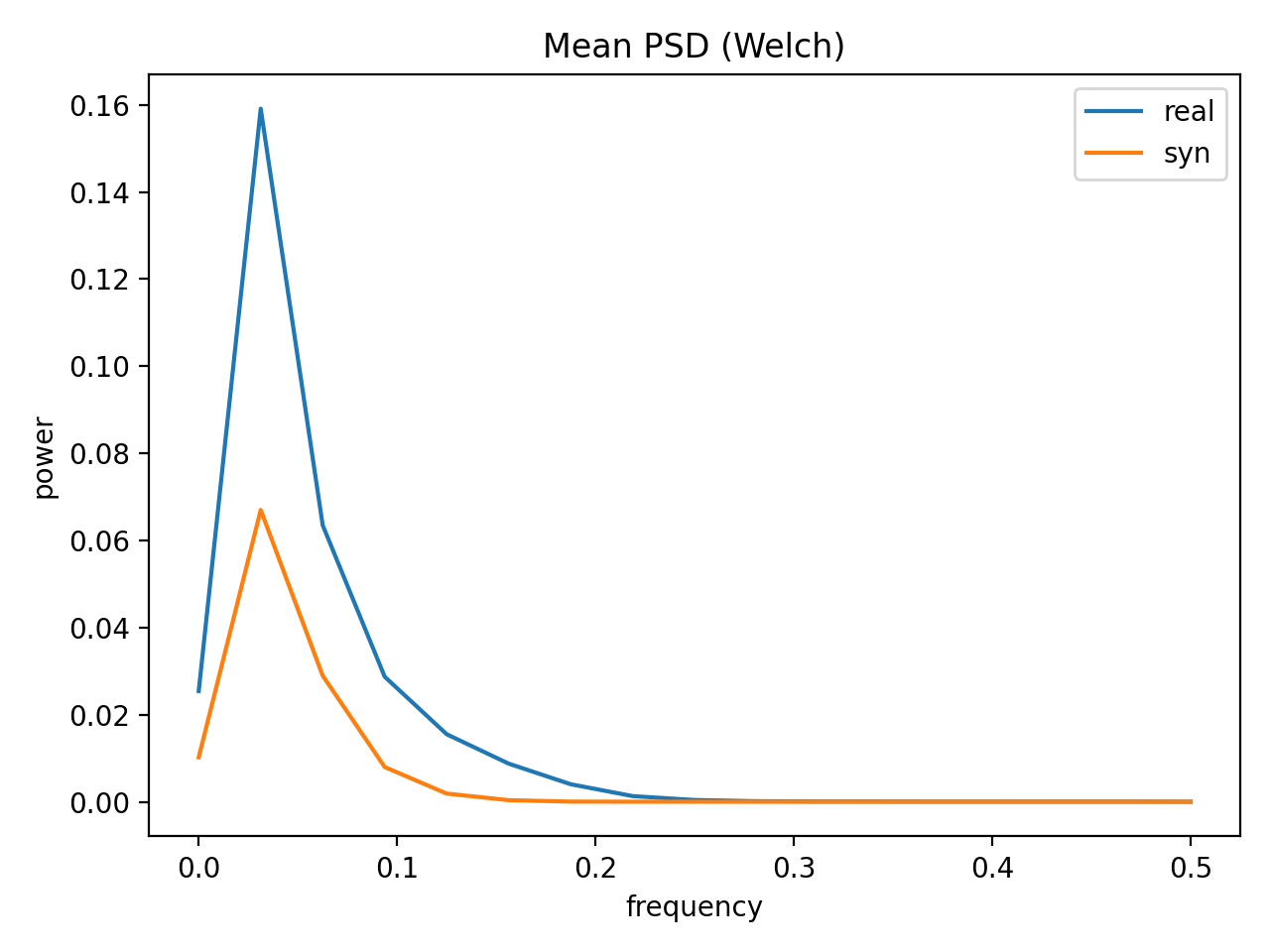}
                \caption{}
                \label{fig:psd_no_graph}
            \end{subfigure}
            \hfill
            \begin{subfigure}{\subfigwidth}
                \centering
                \includegraphics[width=\linewidth]{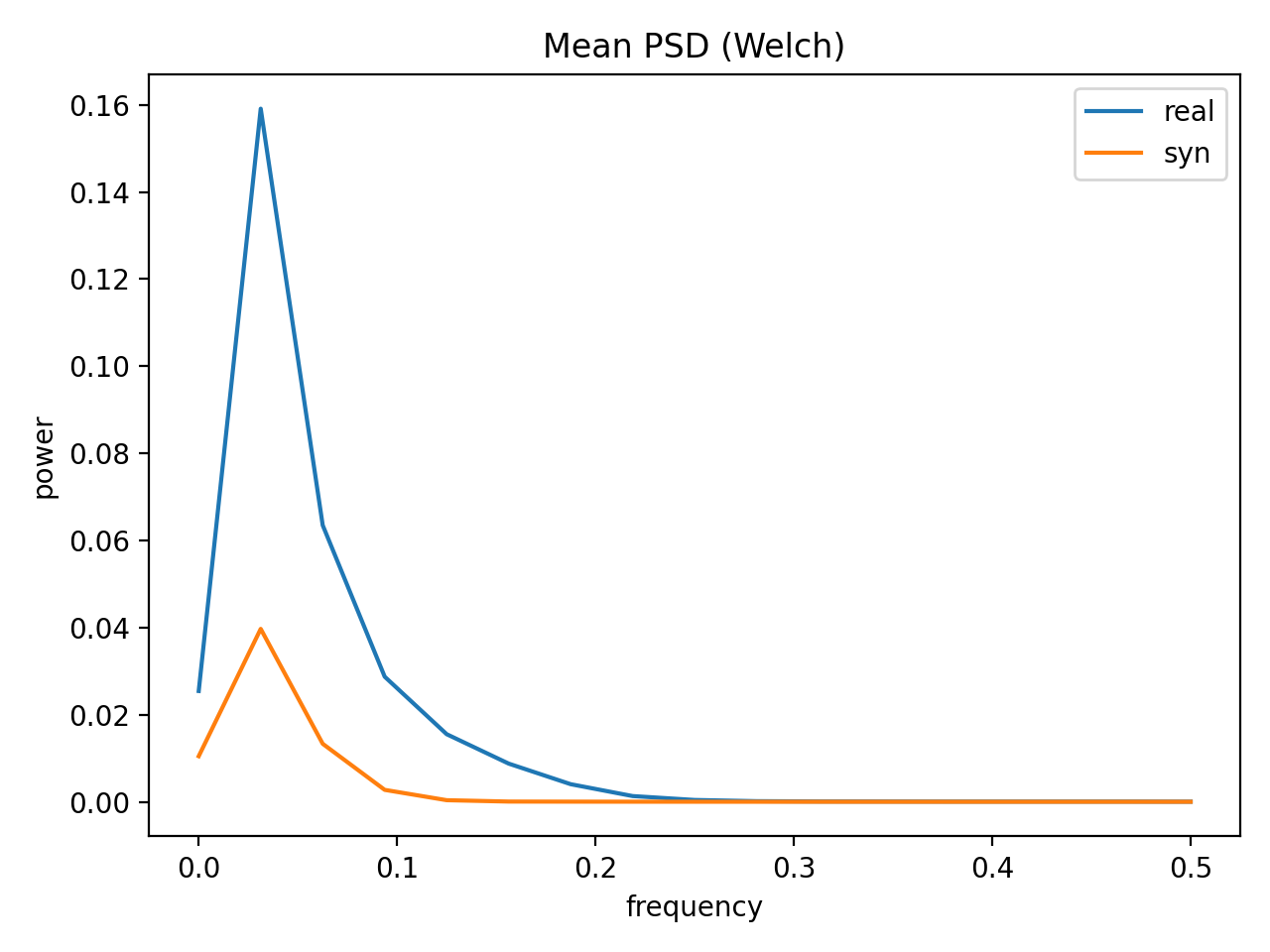}
                \caption{}
                \label{fig:psd_no_stoch}
            \end{subfigure}
        \end{minipage}

        \vspace{0.6em}

        \begin{minipage}[c]{\rowlabelwidth}
            \centering
            \rotatebox{90}{\textbf{\large t-SNE}}
        \end{minipage}%
        \hfill
        \begin{minipage}[c]{\subfigcontainer}
            \begin{subfigure}{\subfigwidth}
                \centering
                \includegraphics[width=\linewidth]{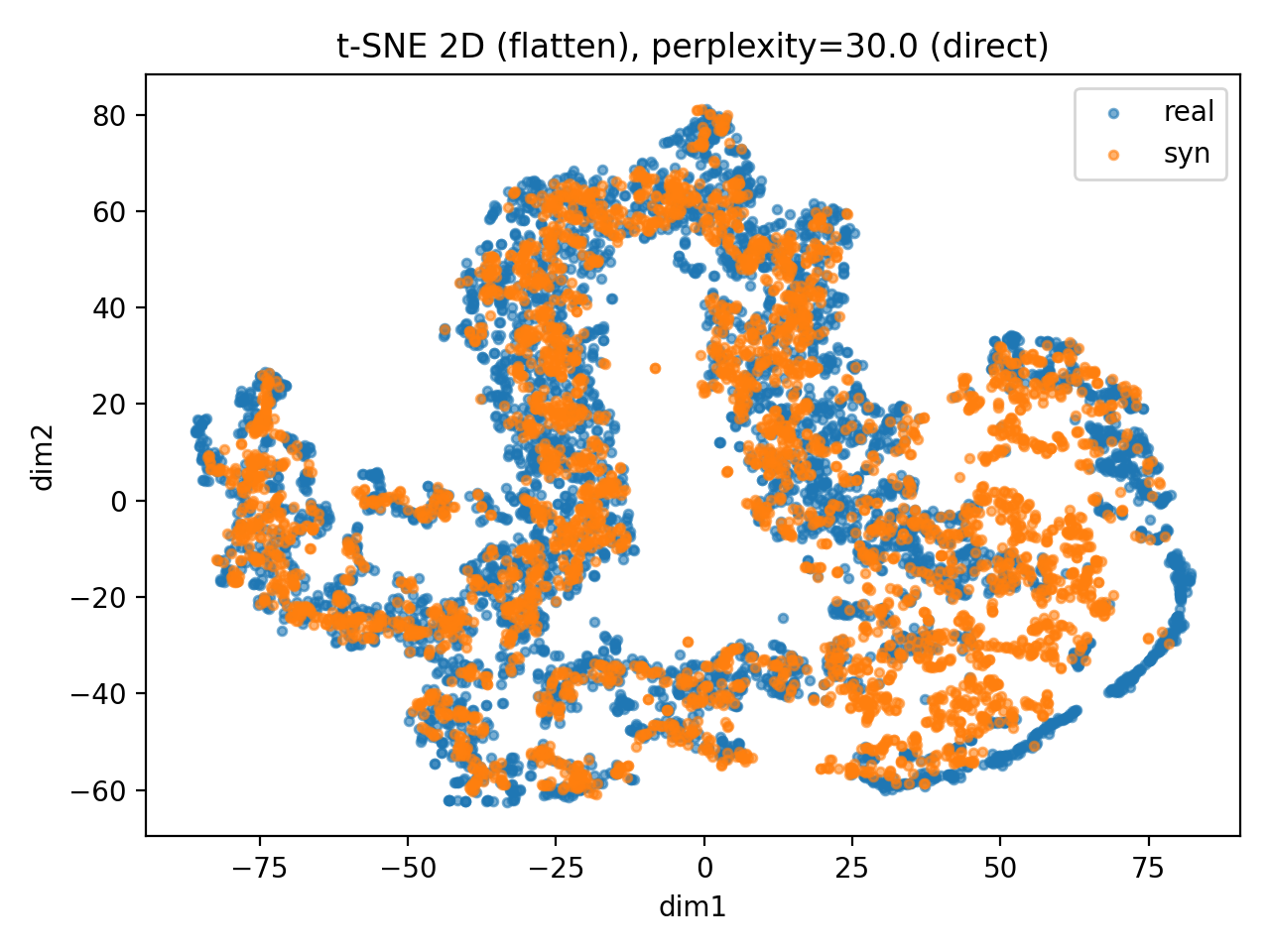}
                \caption{}
                \label{fig:tsne_full}
            \end{subfigure}
            \hfill
            \begin{subfigure}{\subfigwidth}
                \centering
                \includegraphics[width=\linewidth]{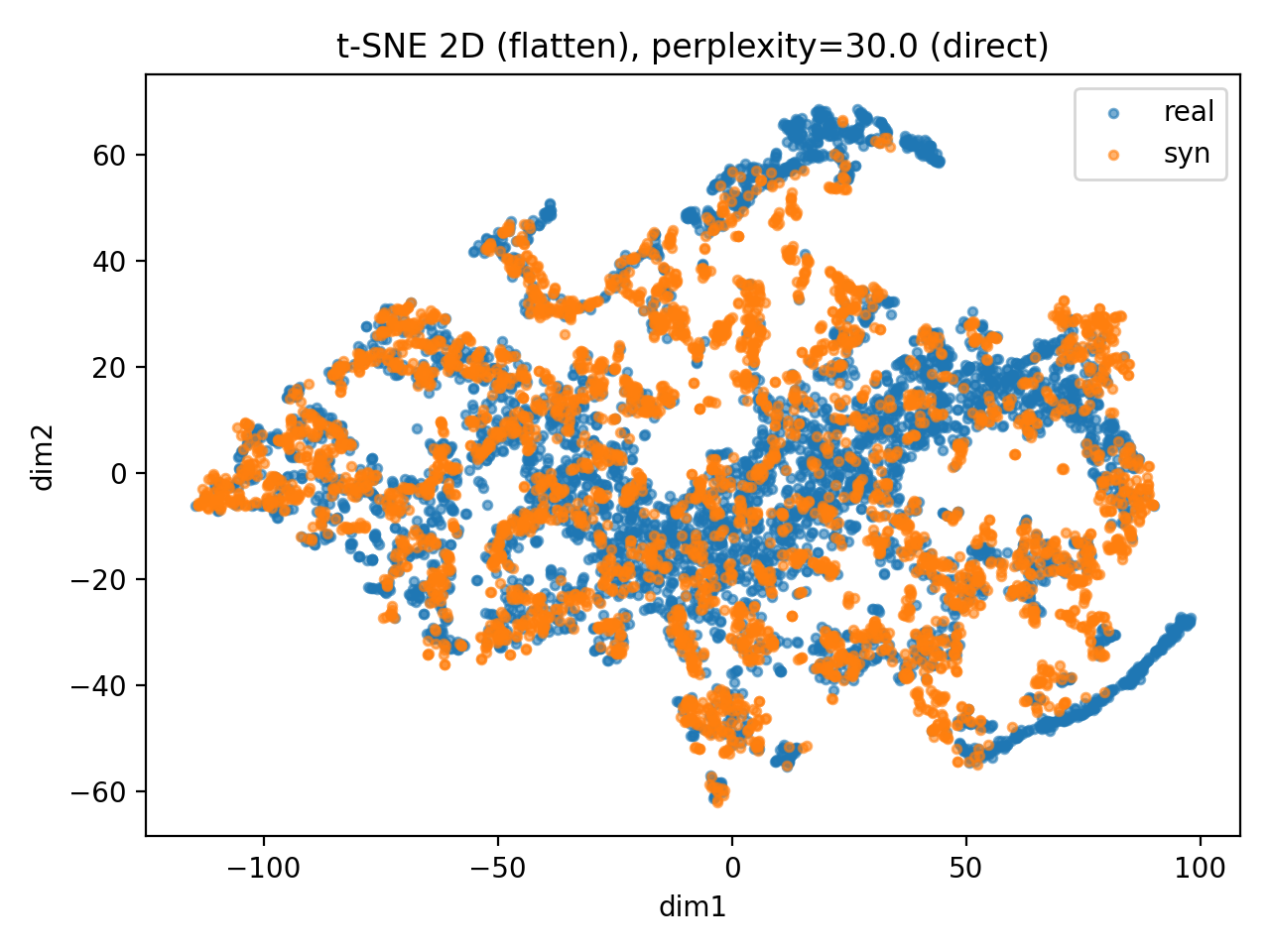}
                \caption{}
                \label{fig:tsne_no_graph}
            \end{subfigure}
            \hfill
            \begin{subfigure}{\subfigwidth}
                \centering
                \includegraphics[width=\linewidth]{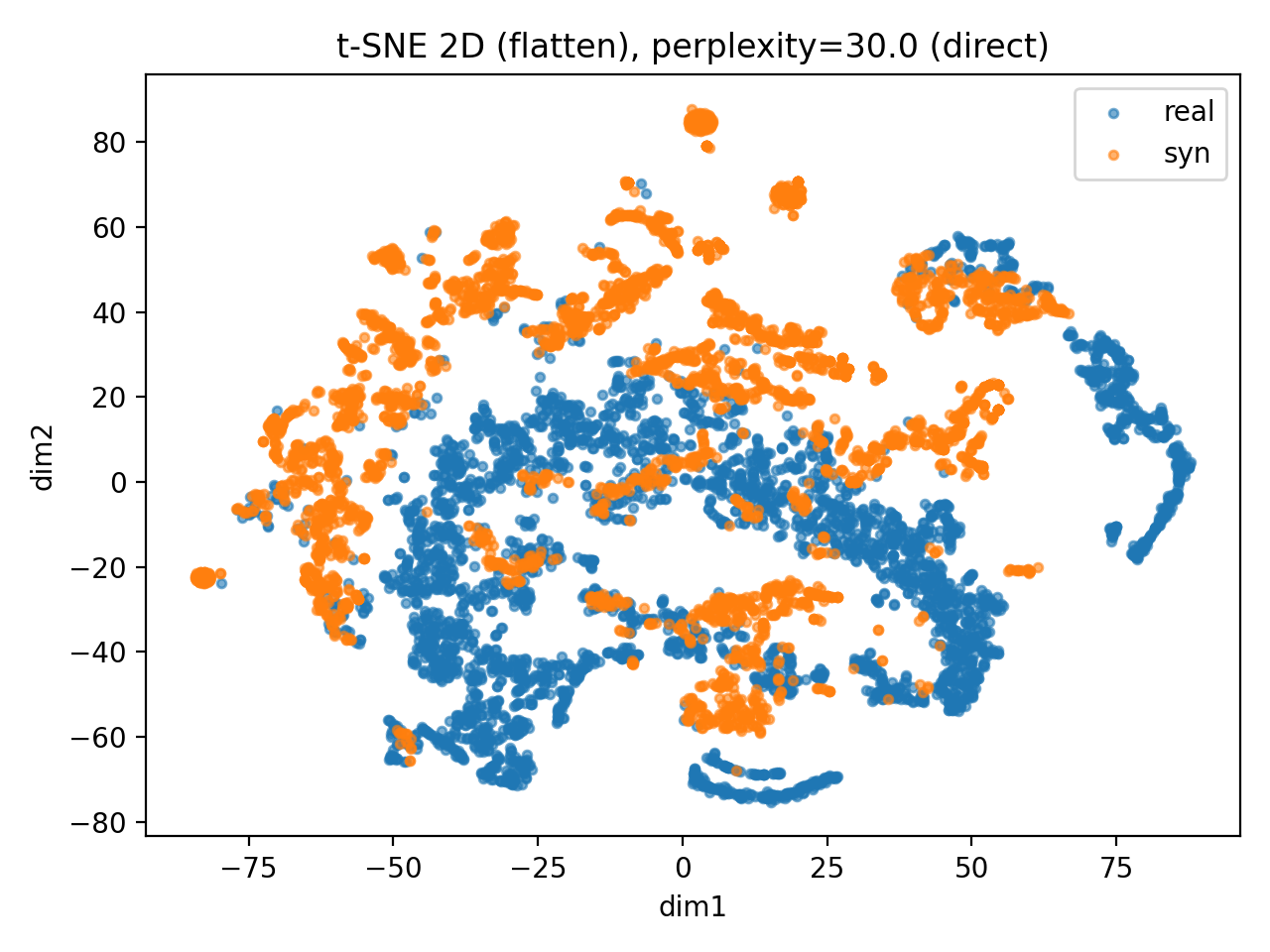}
                \caption{}
                \label{fig:tsne_no_stoch}
            \end{subfigure}
        \end{minipage}

        \vspace{0.5em}

        \begin{minipage}[c]{\rowlabelwidth}
            \quad
        \end{minipage}%
        \hfill
        \begin{minipage}[c]{\subfigcontainer}
            \begin{minipage}{\subfigwidth}
                \centering
                \textbf{(a) Full Model}
            \end{minipage}%
            \hfill
            \begin{minipage}{\subfigwidth}
                \centering
                \textbf{(b) w/o Graph}
            \end{minipage}%
            \hfill
            \begin{minipage}{\subfigwidth}
                \centering
                \textbf{(c) w/o Stochasticity}
            \end{minipage}
        \end{minipage}

    \end{minipage} 
    } 

    \caption{
    Ablation study visualizing the impact of structural conditioning and stochastic residual modeling.
    \textbf{Rows (top to bottom):} Autocorrelation Function (ACF), Power Spectral Density (PSD), and t-SNE visualization.
    \textbf{Columns (left to right):} The proposed Full Model, the model without the graph module (w/o Graph), and the model without the stochastic component (w/o Stochasticity).
    Removing stochasticity leads to pronounced spectral energy collapse and mode contraction, while removing the graph weakens and destroy low-frequency temporal structure.
    }
    \label{fig:ablation_component}
\end{figure*}

\begin{table*}[t!]
\centering
\footnotesize
\caption{Ablation study on CHB-MIT and Sunspot datasets. Each row removes or modifies a single loss component.
Metrics: \textbf{Wass}: Wasserstein distance, \textbf{KS}: KS statistic, \textbf{ACF}: ACF MAE, \textbf{PSD}: PSD L2 distance, \textbf{P-Err}: ProtoErr, \textbf{Cov}: Coverage@$\tau_{0.5}$.
($\downarrow$: Lower is better, $\uparrow$: Higher is better).}
\label{tab:ablation_loss}

\begin{tabular}{l l c c c c c c}
\toprule
\multirow{2}{*}{\textbf{Dataset}} & \multirow{2}{*}{\textbf{Ablation}} 
& \multicolumn{2}{c}{\textbf{Distribution} ($\downarrow$)} 
& \multicolumn{2}{c}{\textbf{Temporal} ($\downarrow$)} 
& \textbf{Representativeness} ($\downarrow$)
& \textbf{Coverage} ($\uparrow$) \\
\cmidrule(lr){3-4} \cmidrule(lr){5-6} \cmidrule(lr){7-7} \cmidrule(lr){8-8}
& & \textbf{Wass} & \textbf{KS} & \textbf{ACF} & \textbf{PSD} & \textbf{P-Err(avg)} & \textbf{Cov@0.5} \\
\midrule

\multirow{5}{*}{\textbf{CHB-MIT}}
& \textbf{Full model}           & 5.07e-6           & 0.024          & \textbf{0.039} & 9.36e-10          & 2.329          & 0.987 \\
& $\mathcal{L}_{recon}=0$       & \textbf{2.73e-6}  & \textbf{0.013} & 0.203          & 1.00e-9           & 4.478          & 0.711 \\
& $\mathcal{L}_{align}=0$       & 3.91e-6  & 0.024          & 0.044          & \textbf{7.08e-10} & 2.310          & \textbf{0.990} \\
& $\mathcal{L}_{dist}=0$        & 4.22e-6           & 0.021 & 0.051          & 9.50e-10          & \textbf{2.302} & \textbf{0.990} \\
& $\beta_{\mathrm{KL}}=0$       & 5.64e-5           & 0.265          & 0.115          & 2.18e-8           & 6.288          & 0.382 \\
\midrule

\multirow{5}{*}{\textbf{Sunspot}}
& \textbf{Full model}           & 2.329          & 0.113          & 0.077          & 8.02e2          & 4.546          & 0.828 \\
& $\mathcal{L}_{recon}=0$       & \textbf{1.709} & \textbf{0.082} & 0.179          & 3.69e3          & 8.300          & 0.282 \\
& $\mathcal{L}_{align}=0$       & 2.605          & 0.112          & \textbf{0.066} & 9.21e2          & \textbf{4.482} & \textbf{0.831} \\
& $\mathcal{L}_{dist}=0$        & 2.510          & 0.107          & 0.074          & \textbf{4.88e2} & 4.594          & 0.825 \\
& $\beta_{\mathrm{KL}}=0$       & 29.78         & 0.219          & 0.117          & 1.09e4          & 7.297          & 0.398 \\

\bottomrule
\end{tabular}
\end{table*}

\subsection{Effect of Structural Conditioning and Stochasticity Residual}
\label{sec:ablation_structure}


Table~\ref{tab:ablation_ecg_structure} and Fig.~\ref{fig:ablation_component} provide empirical support for the proposed structure--residual decomposition. By ablating either the structural component (graph conditioning) or the stochastic residual component (latent variables), we observe consistent performance degradation, confirming that both components are essential for faithful and diverse time-series generation.

\textbf{Effect of structural conditioning.}
Replacing the graph with an identity matrix removes explicit structural information while keeping the model architecture unchanged. Although the model does not collapse, it exhibits consistent degradation in distributional fidelity, temporal alignment, and manifold coverage. As shown in Fig.\ref{fig:psd_no_graph}, without a graph, the low-frequency component of the generated time series degrades, indicating that the global structure is destroyed.
This indicates that the graph provides a strong inductive bias that stabilizes the learned conditional mean $f(G)=\mathbb{E}[X\mid G]$.
When $G$ carries little information, $f(G)$ degenerates into a weak, averaged backbone, increasing the burden on stochastic modeling and leading to systematic performance degradation.

\textbf{Effect of stochastic residual modeling and its interaction with structure.}
Removing stochastic latent variables yields a deterministic \textbf{Graph2TS}
variant (see Appendix~\ref{sec:model_without_stochasticity}), leading to severe
mode contraction, increased prototype error, and reduced coverage.
As shown in Fig.~\ref{fig:ablation_component}, deterministic generation
strongly attenuates low- and mid-frequency energy, producing a flattened power
spectrum.
Although graph conditioning enforces structural consistency, the absence of
stochasticity forces the decoder to collapse all admissible realizations into a
single conditional average $x \approx f_\theta(g)$, suppressing spectral energy
and resulting in over-smoothed signals.

In contrast, the full model combines structural conditioning with stochastic residual modeling: the graph anchors a stable structural backbone $f(G)$, while the latent variable models the residual variability $R$ with
$\mathbb{E}[R \mid G]\approx 0$.
Together, they enable accurate distributional matching, temporal consistency, and faithful manifold coverage, particularly for high-variability biomedical signals such as ECG.

\subsection{Loss Components}

Table~\ref{tab:ablation_loss} reports ablation results on the CHB-MIT and Sunspot datasets, respectively, where each row removes or modifies a single loss component while keeping all other settings fixed. These experiments aim to clarify the functional role of each term in the proposed objective.

\textbf{Effect of reconstruction and KL regularization.}
Removing the reconstruction loss ($\mathcal{L}_{recon}=0$) leads to deceptively improved marginal distribution metrics, such as lower Wasserstein distance and KS statistic, while causing severe degradation in temporal alignment, prototype error, and coverage on both datasets.
This indicates that distribution matching alone is insufficient to ensure structurally coherent and representative time-series generation.
In contrast, removing KL regularization ($\beta_{\mathrm{KL}}=0$) results in a systematic collapse across all metrics, highlighting its essential role in stabilizing the latent space and enabling meaningful conditional sampling.

\textbf{Effect of auxiliary alignment and distance losses.}
Removing either the alignment loss ($\mathcal{L}_{align}=0$) or the distance loss ($\mathcal{L}_{dist}=0$) results in only marginal performance changes.
While certain temporal or spectral metrics may slightly improve or degrade, overall distributional fidelity and manifold coverage remain largely stable.
These results suggest that $\mathcal{L}_{align}$ and $\mathcal{L}_{dist}$ act as auxiliary regularizers that refine structural details rather than determining the model's core generative capability.

\textbf{Summary.}
Overall, these results indicate that the primary performance gains of Graph2TS originate from structure-conditioned generation via the quantile graph. Auxiliary losses provide incremental refinements but do not account for the core improvements over diffusion- and GAN-based baselines.




\section{Conclusion}




This work presents a novel perspective for time-series generation, successfully addressing the challenge of preserving global temporal organization and modeling stochastic local variations. We argue that many real-world signals are more naturally described by a structural backbone with admissible residual variability, and reformulate time-series generation as a cross-modal problem that maps compact structural representations to sequences via quantile-based transition graphs.

Based on this formulation, we propose \textbf{Graph2TS}, a quantile-graph conditioned variational autoencoder that decouples structural conditioning from stochastic generation. Graph conditioning stabilizes the learned conditional mean and mitigates noise amplification and over-smoothing, while latent variables model controlled residual variability. Extensive experiments across diverse datasets demonstrate improved distributional fidelity, temporal alignment, and manifold coverage, particularly on highly variable signals where diffusion- and GAN-based methods struggle. These results highlight structure-controlled and cross-modal generation as a promising direction for time-series modeling.

To the best of our knowledge, \textbf{Graph2TS} is the first machine learning approach to generate synthetic time series from graphs. Future work will explore extensions to multivariate time series, richer structural graph representations, and broader cross-modal generative settings.

\clearpage

\section*{Impact Statement}


This work contributes to the development of controllable time-series generation by introducing a graph-based, structure-conditioned framework. A primary benefit of this approach is its applicability in settings where realistic synthetic time series are required but access to real data is limited, such as biomedical signal analysis, energy systems, and privacy-sensitive domains.
Furthermore, while the current state of the art has focused on generating time series with certain constraints, to the best of our knowledge, no method has focused on generating time series data from graphs.
By enabling the generation of data that preserves global temporal structure while allowing controlled variability, the proposed method may facilitate model development, benchmarking, and data sharing without exposing sensitive raw measurements.

At the same time, as with most generative models, the ability to synthesize realistic time series may carry potential risks if misused, for example, in deceptive or fraudulent scenarios involving financial or sensor data. These risks are not unique to the proposed method and reflect broader challenges in generative modeling. We believe that the benefits of structure-controlled synthetic data generation, particularly for privacy-preserving research and data accessibility, outweigh these concerns, and that responsible use and domain-specific safeguards are essential when deploying such models in practice.


\bibliography{icml2026/sec8-references}
\bibliographystyle{icml2026/icml2026}

\newpage
\appendix
\onecolumn

\section{Additional Experimental Results and Analysis}
\label{sec:appendix-extra-results}

\subsection{Additional Results on Limitations of Baseline Models on Heavy Tailed Signals}
\label{sec:appendix-sunspot-ecg}
\begin{table}[h!]
\centering
\caption{Distribution statistics in \textbf{z-score space} for Sunspot vs ECG windows (length $T=32$).
We report summary statistics for raw values $x$ and first-order differences $\Delta x$.
Excess kurtosis indicates tail-heaviness (larger $\Rightarrow$ heavier tails).}
\label{tab:sunspot_ecg_stats}
\resizebox{\linewidth}{!}{
\begin{tabular}{lcccccccc}
\toprule
\textbf{Dataset} & $B$ & \multicolumn{4}{c}{\textbf{$x$ statistics}} & \multicolumn{3}{c}{\textbf{$\Delta x$ statistics}} \\
\cmidrule(lr){3-6}\cmidrule(lr){7-9}
 &  & mean & std & excess kurt. & quantile range (0.1\% $\rightarrow$ 99.9\%) & mean & std & excess kurt. \\
\midrule
Sunspot & 1758  & -0.0360 & 0.9668 & 0.5321 & [-1.0741,\; 3.5320] & -7.5e-05 & 0.1994 & 2.2466 \\
ECG     & 20312 & 0.2148  & 0.8283 & 7.3562 & [-1.6078,\; 5.1106] & 4.4e-05  & 0.1819 & 41.5587 \\
\bottomrule
\end{tabular}}
\end{table}
\begin{figure}[h!]
\centering
\begin{subfigure}[t]{0.49\linewidth}
  \centering
  \includegraphics[width=\linewidth]{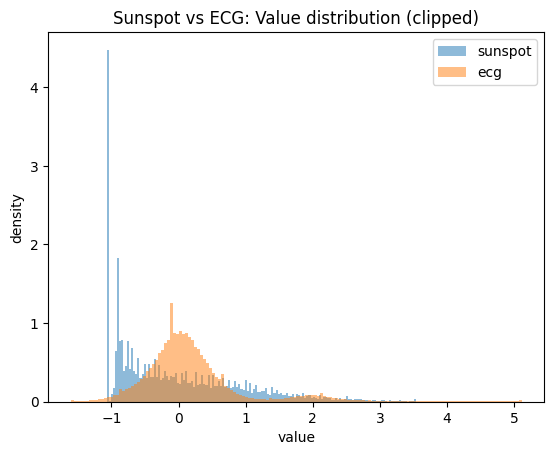}
  \caption{Value distribution $p(x)$ (clipped for visualization).}
  \label{fig:sunspot_ecg_value_hist}
\end{subfigure}\hfill
\begin{subfigure}[t]{0.49\linewidth}
  \centering
  \includegraphics[width=\linewidth]{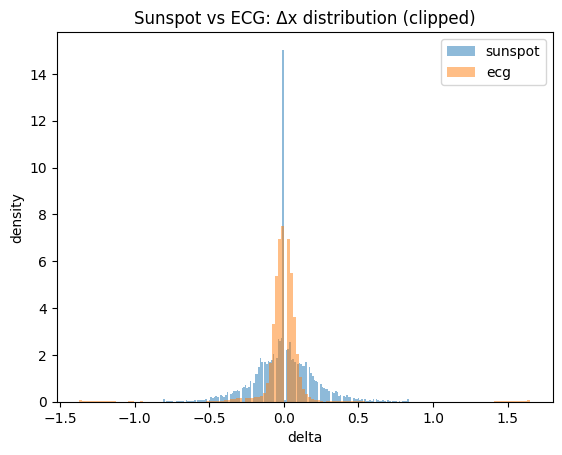}
  \caption{First-order difference distribution $p(\Delta x)$ (clipped).}
  \label{fig:sunspot_ecg_delta_hist}
\end{subfigure}
\caption{Sunspot vs ECG in \textbf{z-score space}. ECG exhibits substantially heavier tails, especially in $\Delta x$,
indicating more frequent abrupt transitions compared to the smoother Sunspot dynamics.}
\label{fig:sunspot_ecg_hist}
\end{figure}

Table~\ref{tab:sunspot_ecg_stats} and Fig.~\ref{fig:sunspot_ecg_hist} summarize the empirical distributions in the
\textbf{same input space used by models} (z-score normalized windows).
Sunspot is close to a light-tailed process: $x$ has low excess kurtosis ($0.53$) and its increment distribution
$\Delta x$ is only mildly heavy-tailed (excess kurtosis $2.25$).
In contrast, ECG is strongly non-Gaussian: $x$ exhibits heavy tails (excess kurtosis $7.36$) and, more importantly,
its increment distribution is extremely heavy-tailed (excess kurtosis $41.56$), revealing frequent abrupt changes.

\textbf{Implication for generative modeling.}
Diffusion-based time-series generators implicitly assume that the target distribution in the normalized space is
reasonably smooth and well-approximated by iterative Gaussian perturbations. This assumption aligns well with the
smooth, low-frequency dynamics of Sunspot, but is violated by ECG where sharp transitions dominate $\Delta x$.
Consequently, DiffusionTS/TimeGAN may over-smooth sharp events or introduce spurious high-frequency artifacts,
leading to noticeable mismatches in ACF/PSD.

\textbf{Why Graph2TS can be advantageous on ECG.}
Our \textbf{Graph2TS} conditions generation on a quantile-transition graph, which emphasizes \emph{structural and relational}
patterns rather than raw amplitude density. This inductive bias is robust to heavy tails and non-stationary increments,
making it not only good at smooth series but also substantially effective for high-variability biomedical signals such as ECG.

\subsection{Qualitative Comparison of Real and Generated Samples}
\label{sec:appendix-examples}

\begin{figure}[t!]
    \centering

    \resizebox{0.9\linewidth}{!}{%
    \begin{minipage}{\linewidth}
        \centering

        \begin{subfigure}[b]{1.0\linewidth}
            \centering
            \includegraphics[width=\linewidth]{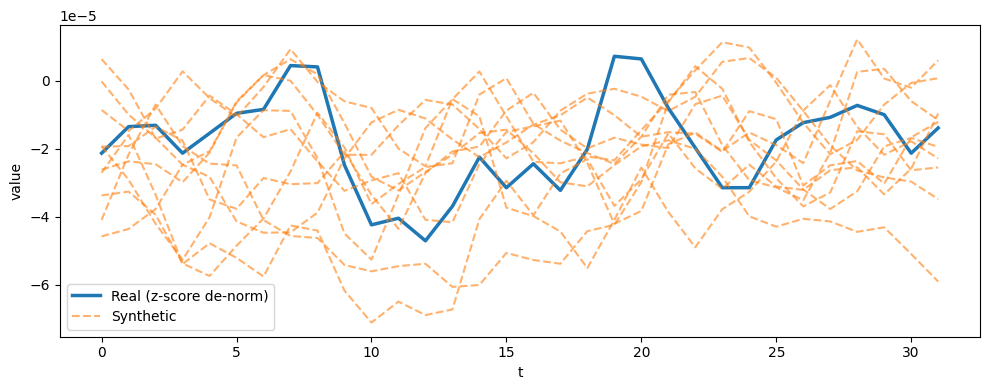}
            \caption{Example 1.}
            \label{fig:sub_top}
        \end{subfigure}

        \par\bigskip

        \begin{subfigure}[b]{1.0\linewidth}
            \centering
            \includegraphics[width=\linewidth]{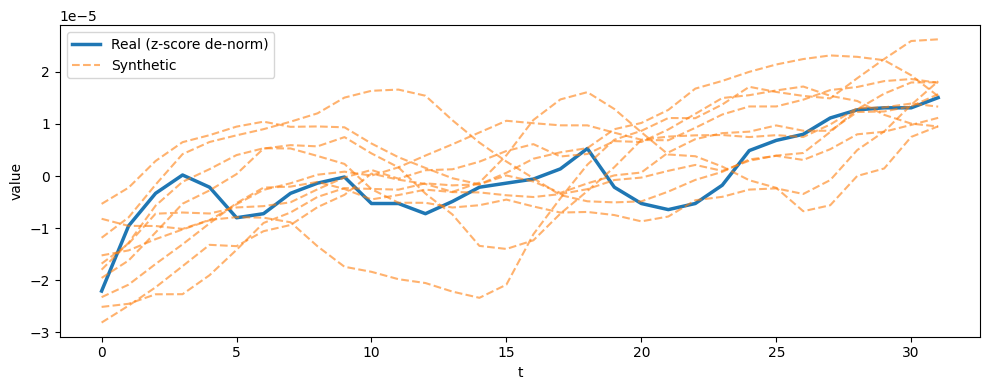}
            \caption{Example 2.}
            \label{fig:sub_mid}
        \end{subfigure}

        \par\bigskip

        \begin{subfigure}[b]{1.0\linewidth}
            \centering
            \includegraphics[width=\linewidth]{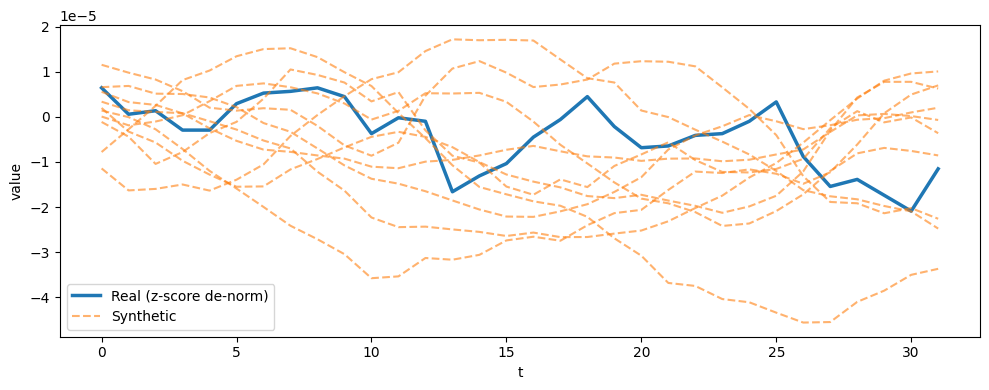}
            \caption{Example 3.}
            \label{fig:sub_bot}
        \end{subfigure}

    \end{minipage}%
    } 

    \caption{Visualization of real time series and corresponding generated time series. Each row (a, b, c) displays 10 samples generated from a single, distinct graph topology, illustrating the model's ability to synthesize data corresponding to different graphs.}
    \label{fig:generated_samples_3graphs}
\end{figure}

\begin{figure}[t!] 
    \centering

    \resizebox{0.9\linewidth}{!}{%
    \begin{minipage}{\linewidth}
        \centering

        \begin{subfigure}[b]{1.0\linewidth}
            \centering
            \includegraphics[width=\linewidth]{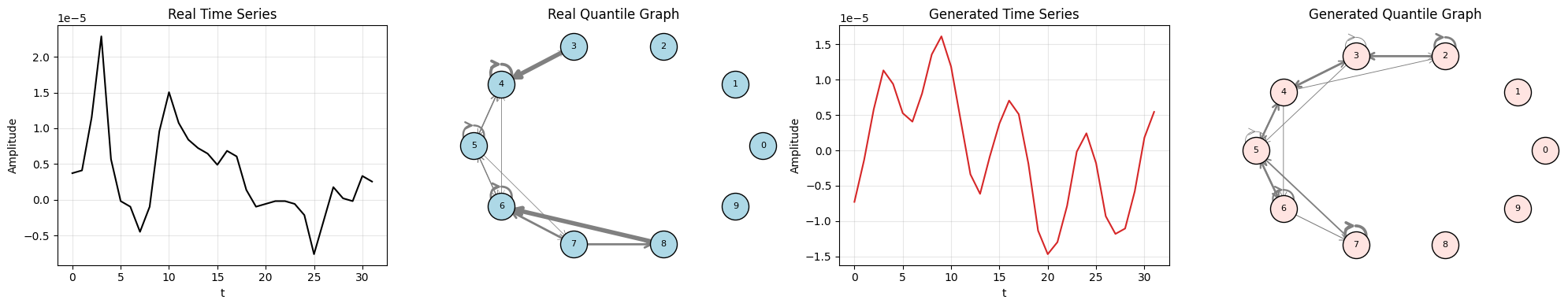}
            \caption{Sample 1: Comparison of real (left) and generated (right) time series and graphs.}
            \label{fig:sample_comparison_1}
        \end{subfigure}

        \par\bigskip

        \begin{subfigure}[b]{1.0\linewidth}
            \centering
            \includegraphics[width=\linewidth]{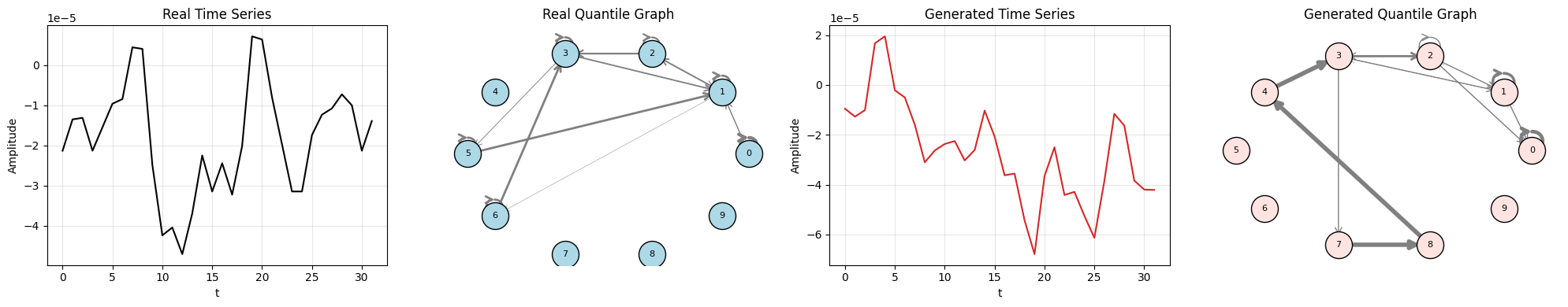}
            \caption{Sample 2: Comparison of real (left) and generated (right) time series and graphs.}
            \label{fig:sample_comparison_2}
        \end{subfigure}

        \par\bigskip

        \begin{subfigure}[b]{1.0\linewidth}
            \centering
            \includegraphics[width=\linewidth]{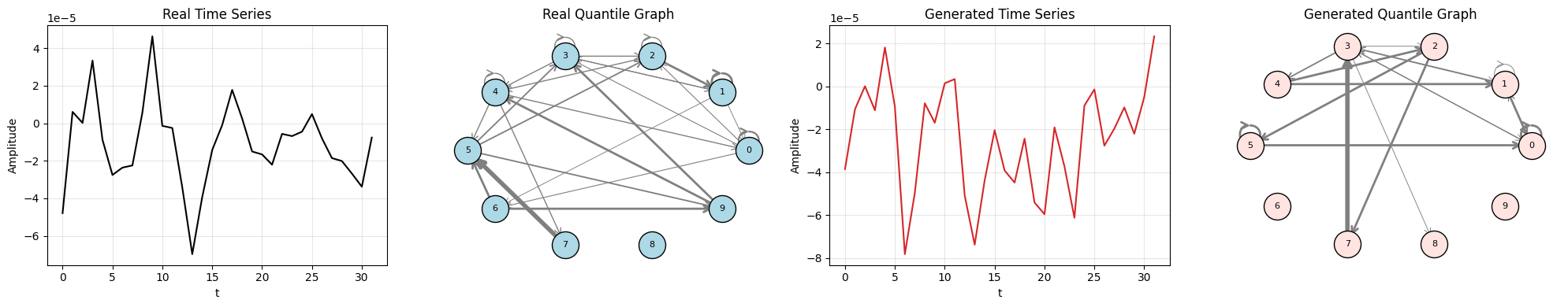}
            \caption{Sample 3: Comparison of real (left) and generated (right) time series and graphs.}
            \label{fig:sample_comparison_3}
        \end{subfigure}

    \end{minipage}%
    } 

    \caption{Qualitative comparison between real and generated data across three distinct samples. Each row visualizes the Real Time Series and Real Quantile Graph (left panels) alongside the Generated Time Series and Generated Quantile Graph (right panels), demonstrating the model's capability to capture both temporal dynamics and structural dependencies.}
    \label{fig:full_comparison_3x1}
\end{figure}

Figure~\ref{fig:generated_samples_3graphs} visualizes real and generated time series across three distinct structural conditions. Each row corresponds to a single, specific quantile graph derived from the "Real" time series shown in that row. Alongside the real sample, we display 10 samples synthesized by our model conditioned on that extracted graph structure.

Fig.~\ref{fig:full_comparison_3x1} provides a qualitative comparison between real and generated data across three distinct samples. The visual alignment between the ground truth (left) and the generated outputs (right) demonstrates the model's ability to capture both temporal dynamics and underlying graph structures.

Concerns may arise regarding the sparsity of graphs derived from short windows ($T=32$).
However, as shown in Fig.~\ref{fig:generated_samples_3graphs} and Fig.~\ref{fig:full_comparison_3x1}, the resulting quantile graphs are not randomly sparse; instead, they exhibit consistent topological structures (e.g., localized transitions and loops) that correlate with specific temporal behaviors.
This confirms that even short windows contain sufficient signal to construct meaningful structural priors, rather than mere noise.

Furthermore, it is worth noting that the graph derived from the generated time series (``Generated Quantile Graph'' in Fig.~\ref{fig:full_comparison_3x1}) is topologically consistent with, yet not identical to, the conditioning ``Real Quantile Graph.'' These minor structural deviations are expected and desirable: they arise from the stochastic residual $z$ introduced by the latent variable.
Since the quantile mapping is sensitive to local amplitude variations, the injected stochasticity naturally induces slight shifts in exact transition probabilities.
Crucially, the global structural backbone (e.g., dominant cycles and connectivity) is preserved, demonstrating that Graph2TS learns a robust conditional distribution $p(x|G)$ that supports diverse realizations, rather than simply memorizing the input transition matrix.

\subsection{Qualitative Analysis of Temporal and Distributional Structure}
\label{sec:appendix_qualitative_analysis}

\begin{figure}[h!]
    \centering

    \resizebox{0.90\textwidth}{!}{%
    \begin{minipage}{\textwidth} 
    \centering

        \newcommand{\rowlabelwidth}{0.04\linewidth}
        \newcommand{\subfigcontainer}{0.95\linewidth}
        \newcommand{\subfigwidth}{0.32\linewidth} 

        \begin{minipage}[c]{\rowlabelwidth}
            \centering
            \rotatebox{90}{\textbf{\large CHB-MIT}}
        \end{minipage}%
        \hfill
        \begin{minipage}[c]{\subfigcontainer}
            \begin{subfigure}{\subfigwidth}
                \centering
                \includegraphics[width=\linewidth]{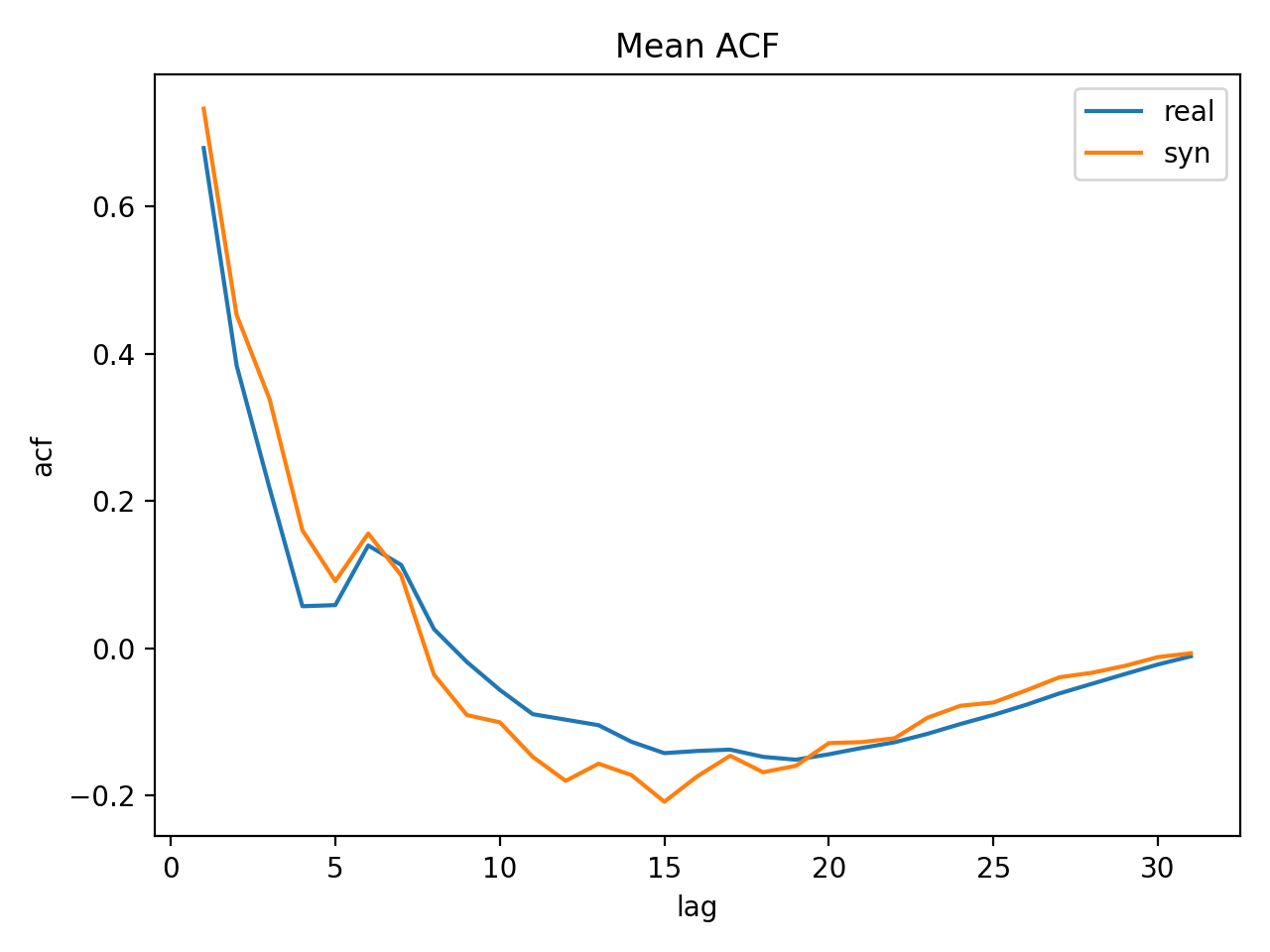}
                \label{fig:acf_chb_cvae}
            \end{subfigure}
            \hfill
            \begin{subfigure}{\subfigwidth}
                \centering
                \includegraphics[width=\linewidth]{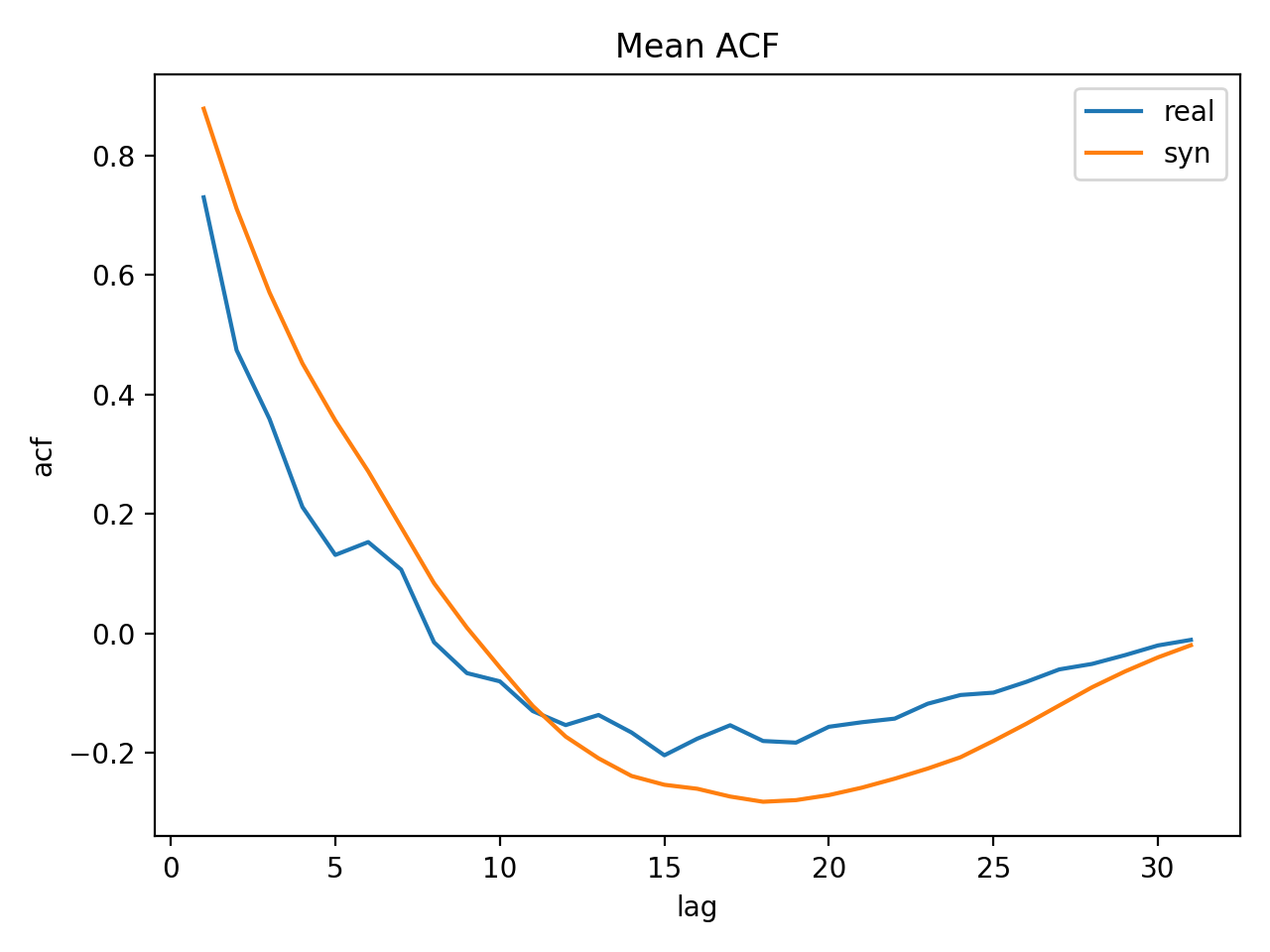}
                \label{fig:acf_chb_diffusion}
            \end{subfigure}
            \hfill
            \begin{subfigure}{\subfigwidth}
                \centering
                \includegraphics[width=\linewidth]{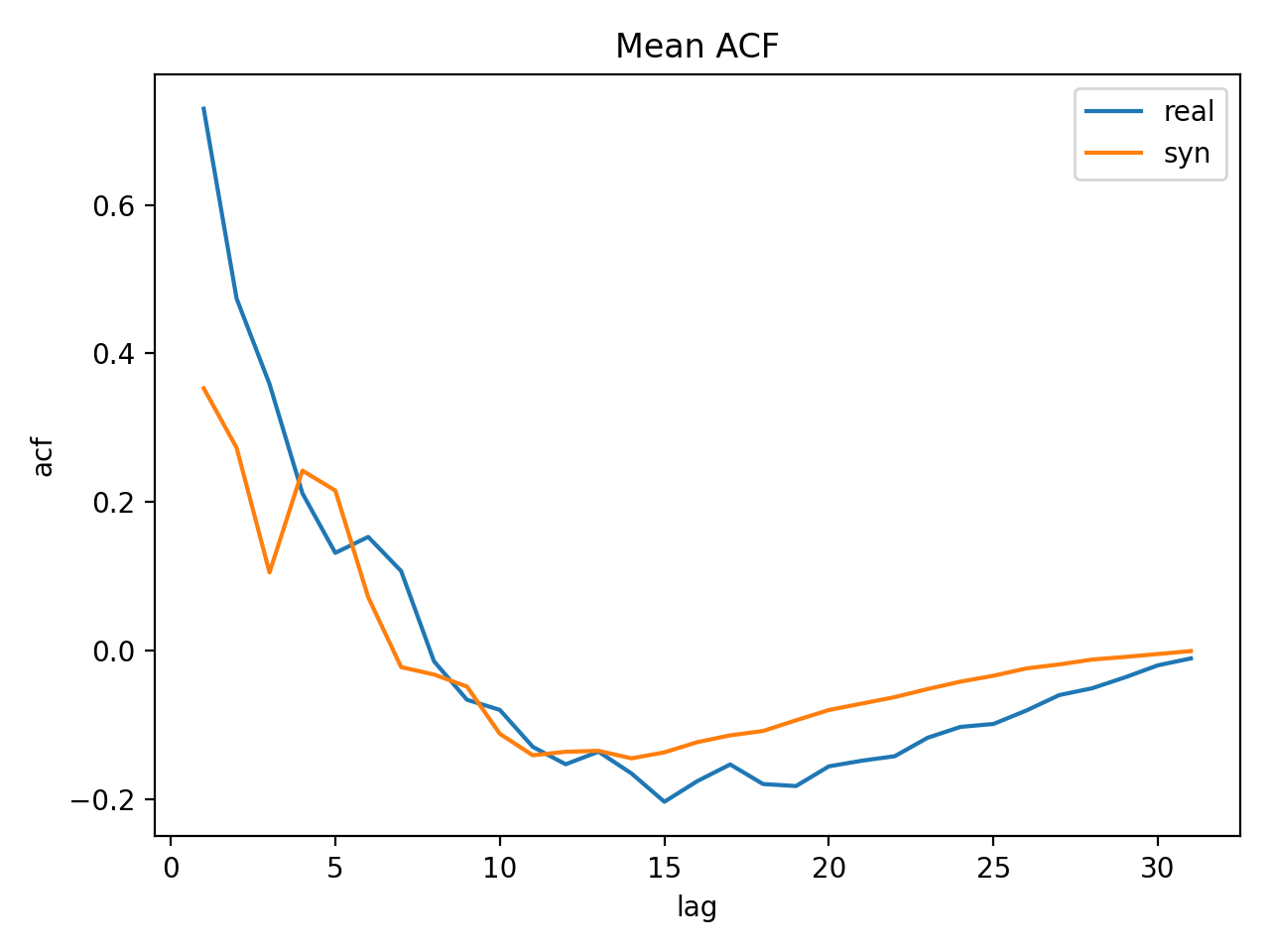}
                \label{fig:acf_chb_timegan}
            \end{subfigure}
        \end{minipage}

        \vspace{0.1em}

        \begin{minipage}[c]{\rowlabelwidth}
            \centering
            \rotatebox{90}{\textbf{\large Sunspot}}
        \end{minipage}%
        \hfill
        \begin{minipage}[c]{\subfigcontainer}
            \begin{subfigure}{\subfigwidth}
                \centering
                \includegraphics[width=\linewidth]{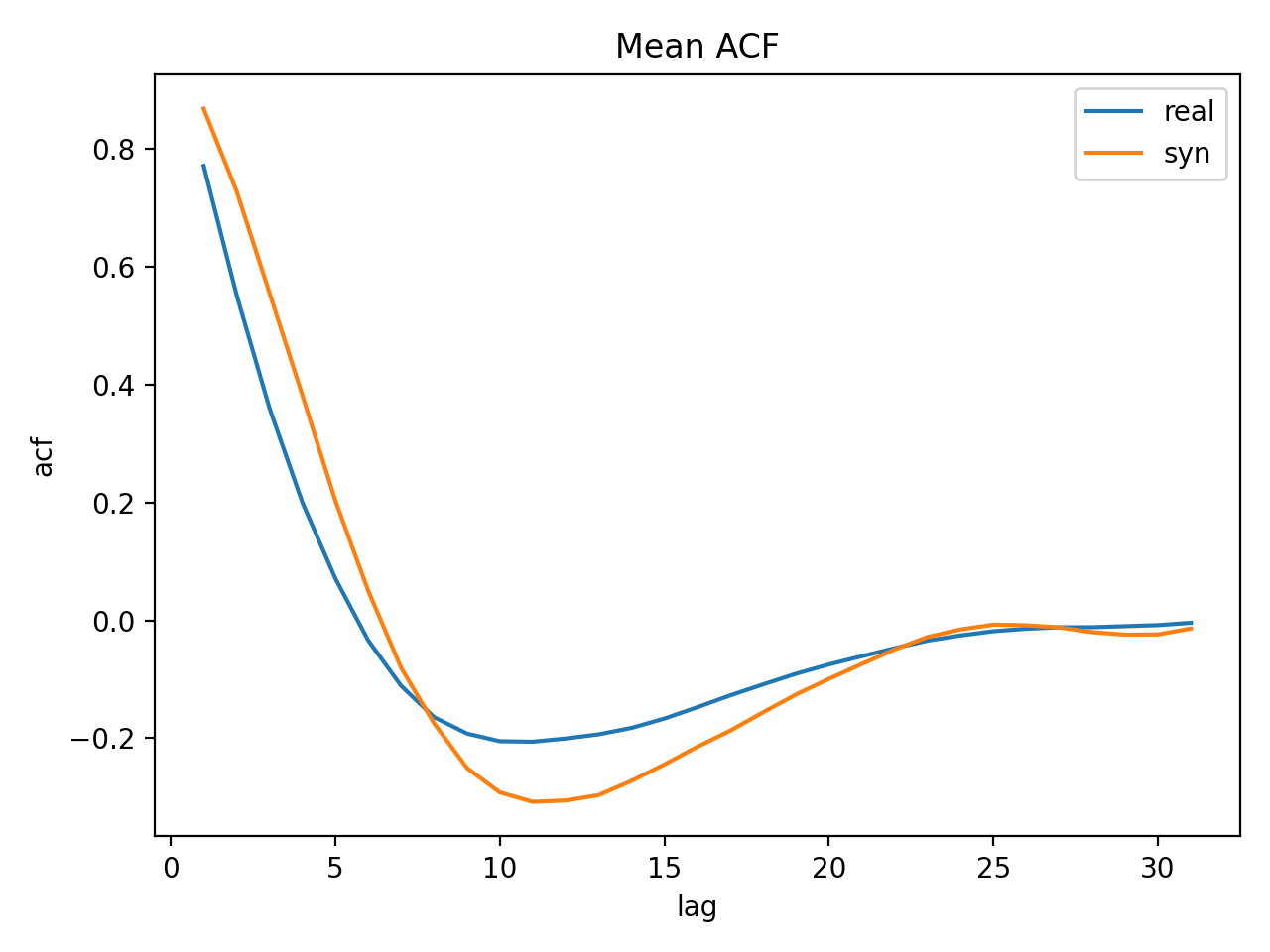}
                \label{fig:acf_sunspot_cvae}
            \end{subfigure}
            \hfill
            \begin{subfigure}{\subfigwidth}
                \centering
                \includegraphics[width=\linewidth]{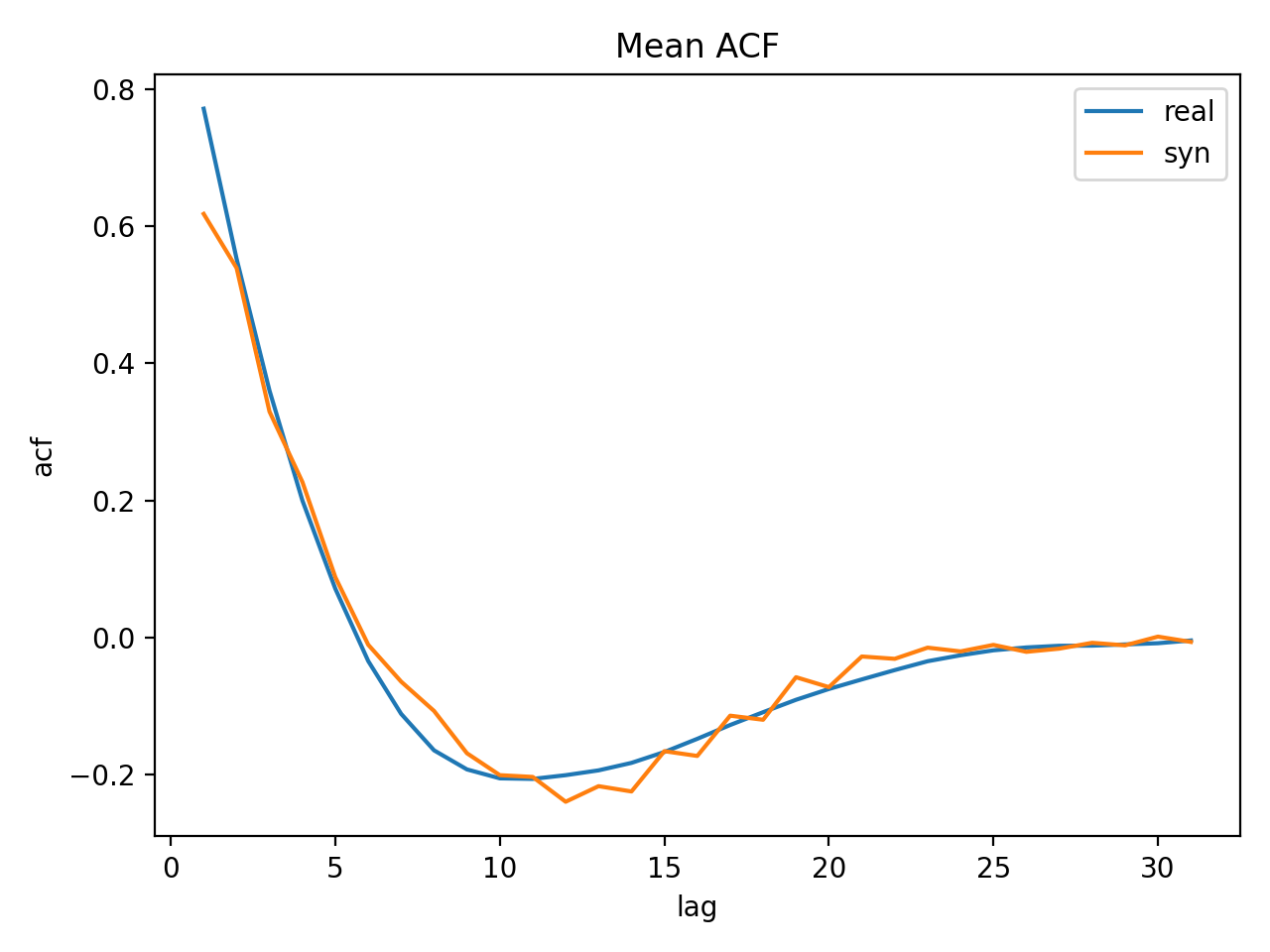}
                \label{fig:acf_sunspot_diffusion}
            \end{subfigure}
            \hfill
            \begin{subfigure}{\subfigwidth}
                \centering
                \includegraphics[width=\linewidth]{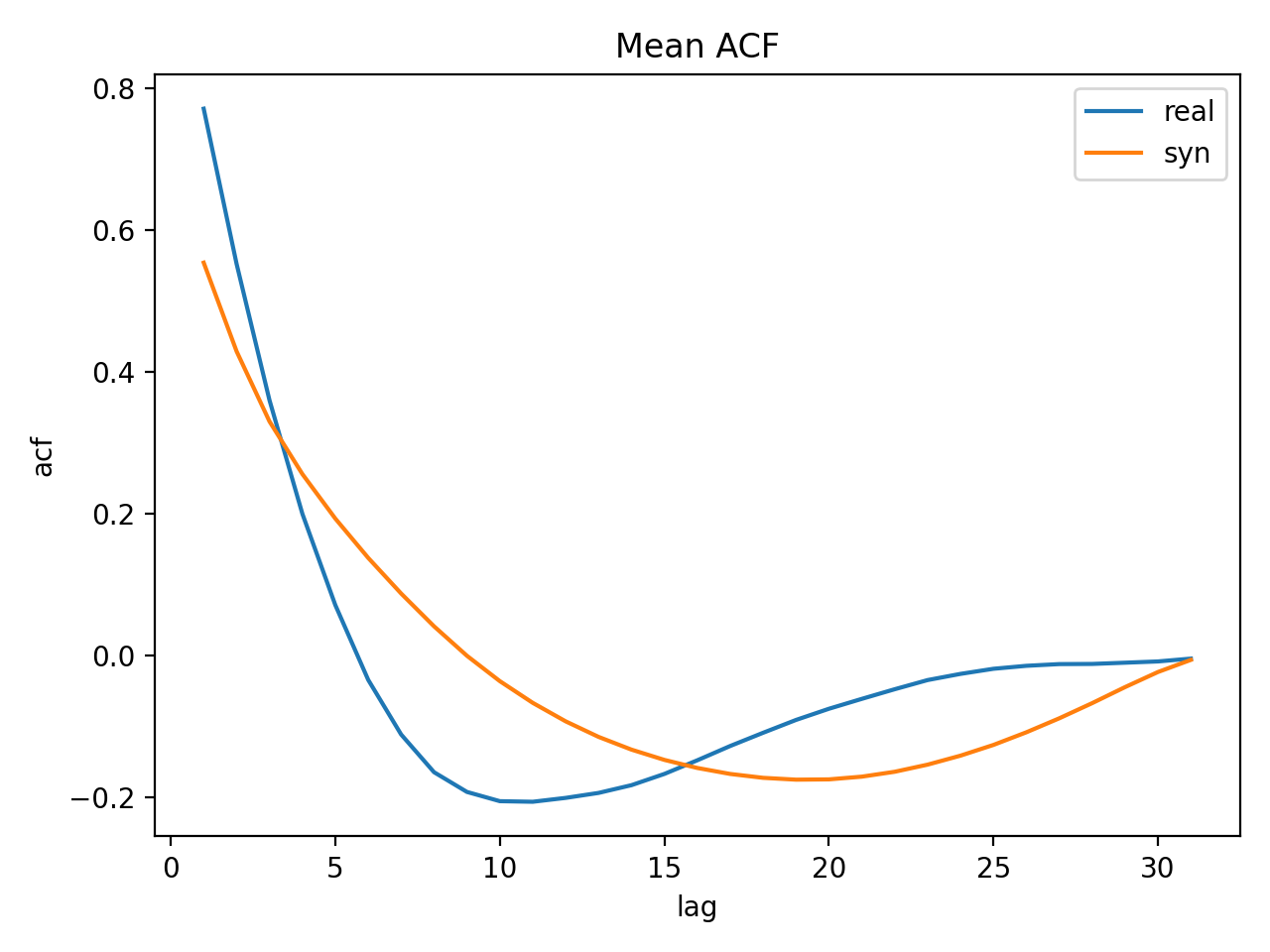}
                \label{fig:acf_sunspot_timegan}
            \end{subfigure}
        \end{minipage}

        \vspace{0.1em}

        \begin{minipage}[c]{\rowlabelwidth}
            \centering
            \rotatebox{90}{\textbf{\large Electricity}}
        \end{minipage}%
        \hfill
        \begin{minipage}[c]{\subfigcontainer}
            \begin{subfigure}{\subfigwidth}
                \centering
                \includegraphics[width=\linewidth]{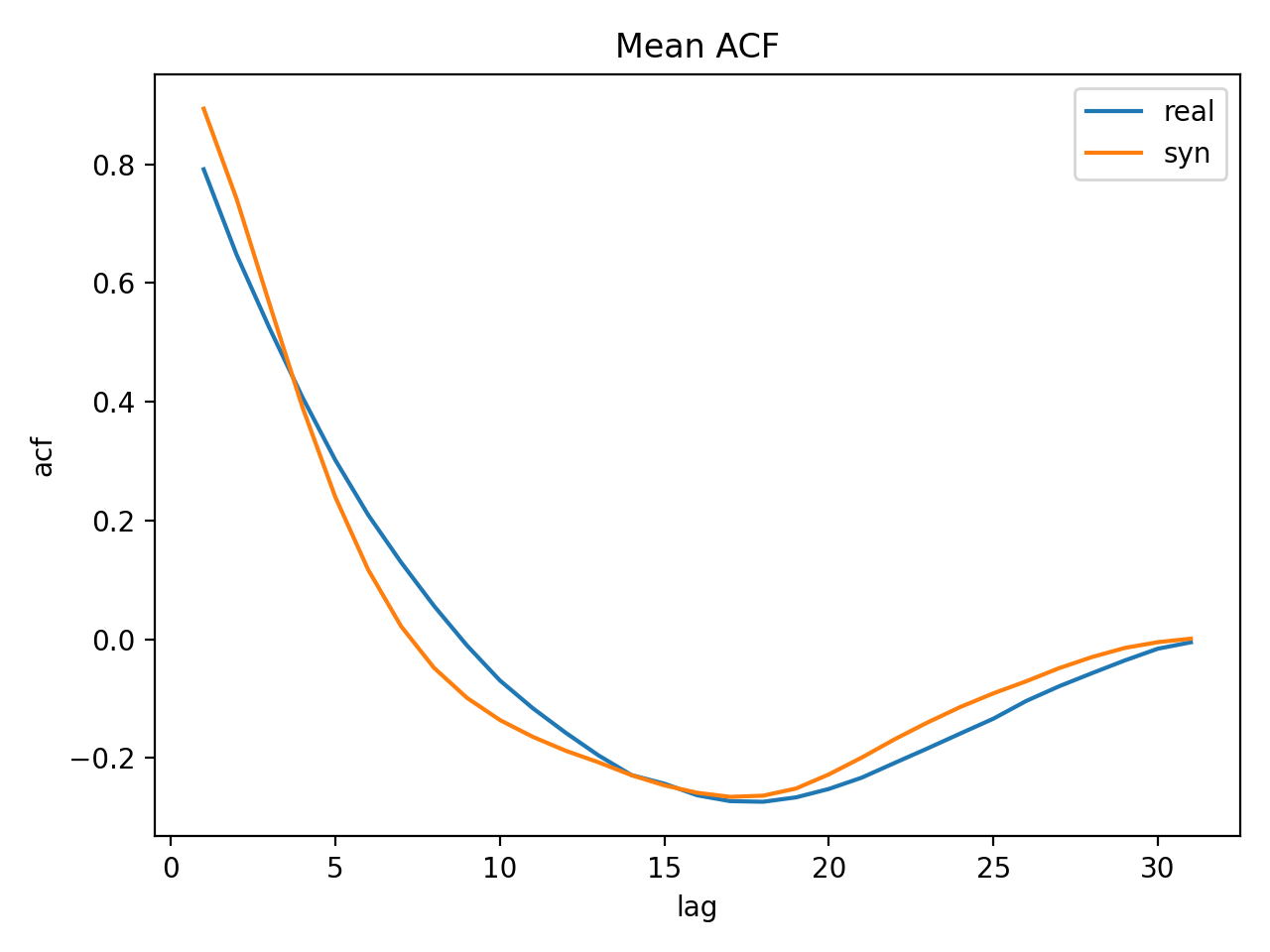}
                \label{fig:acf_elec_cvae}
            \end{subfigure}
            \hfill
            \begin{subfigure}{\subfigwidth}
                \centering
                \includegraphics[width=\linewidth]{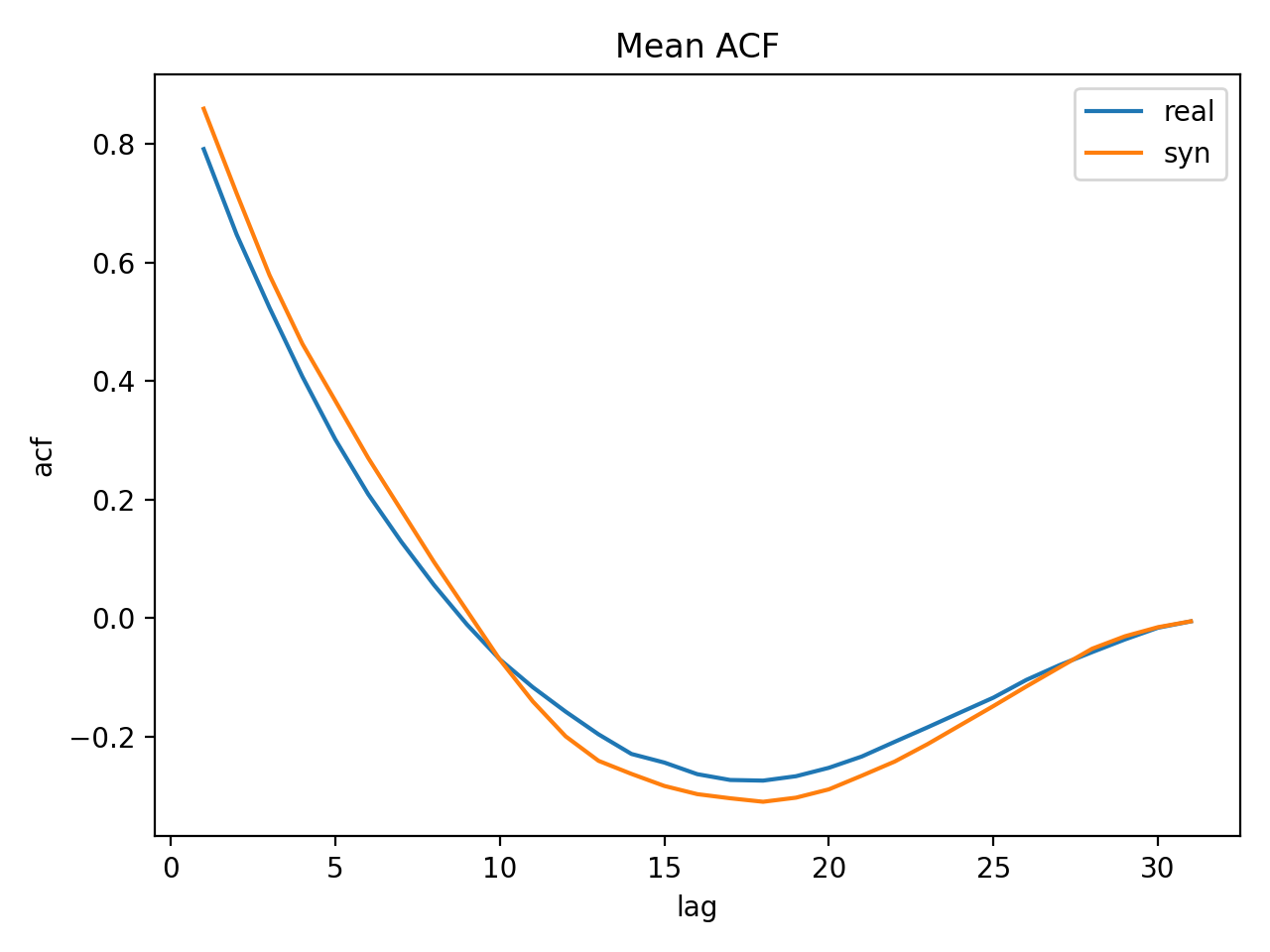}
                \label{fig:acf_elec_diffusion}
            \end{subfigure}
            \hfill
            \begin{subfigure}{\subfigwidth}
                \centering
                \includegraphics[width=\linewidth]{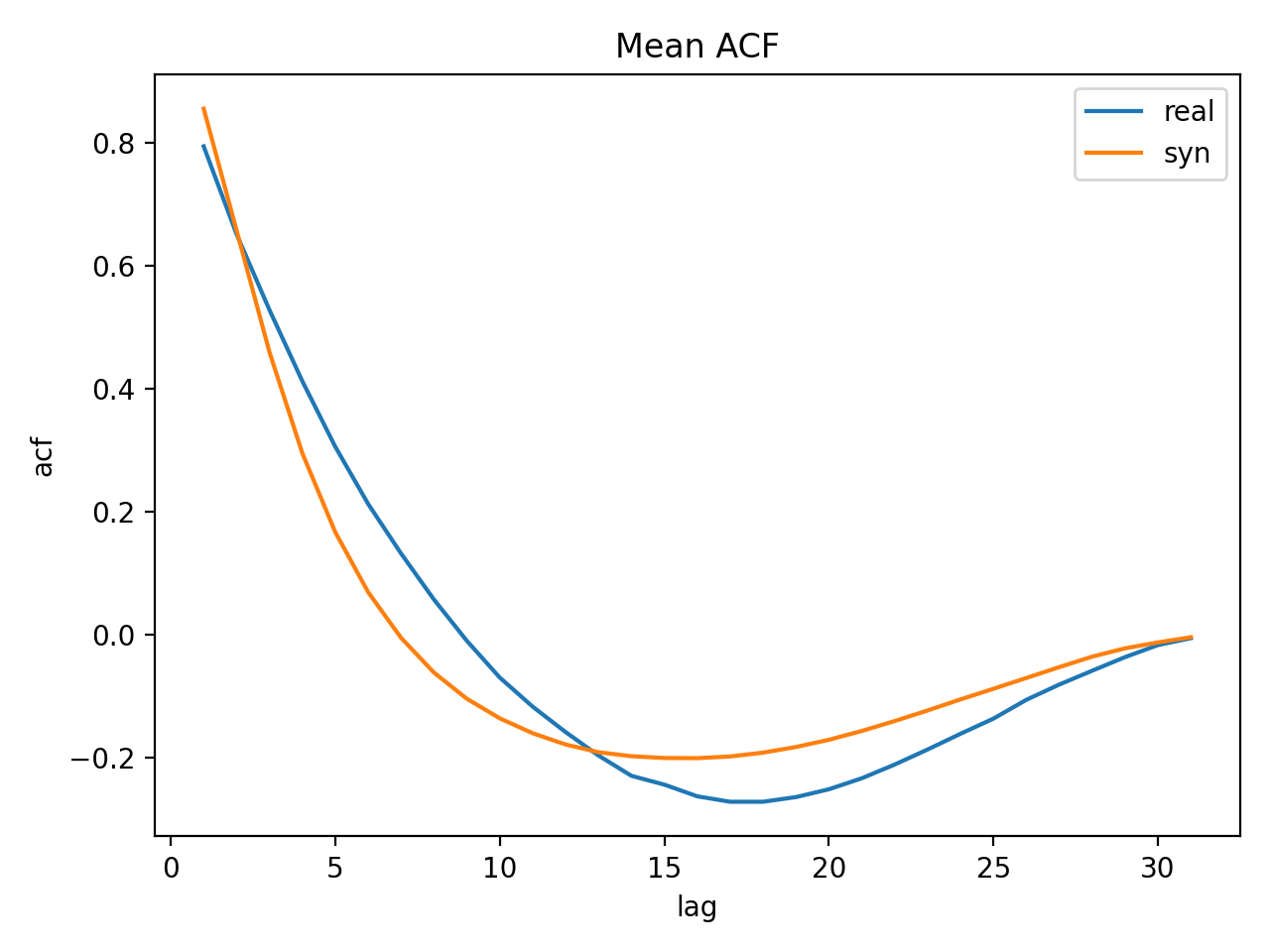}
                \label{fig:acf_elec_timegan}
            \end{subfigure}
        \end{minipage}

        \vspace{0.1em}

        \begin{minipage}[c]{\rowlabelwidth}
            \centering
            \rotatebox{90}{\textbf{\large ECG}}
        \end{minipage}%
        \hfill
        \begin{minipage}[c]{\subfigcontainer}
            \begin{subfigure}{\subfigwidth}
                \centering
                \includegraphics[width=\linewidth]{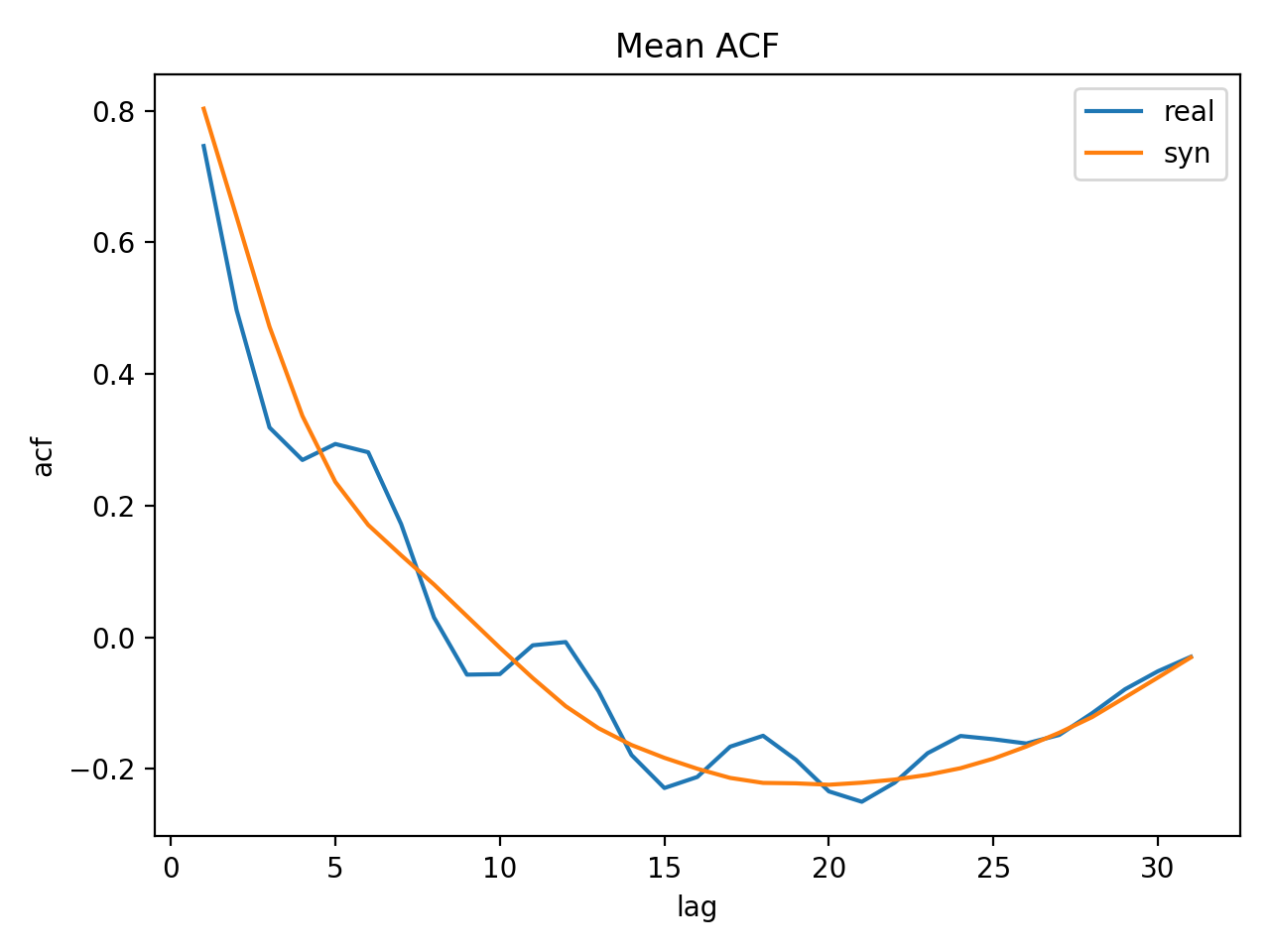}
                \label{fig:acf_ecg_cvae}
            \end{subfigure}
            \hfill
            \begin{subfigure}{\subfigwidth}
                \centering
                \includegraphics[width=\linewidth]{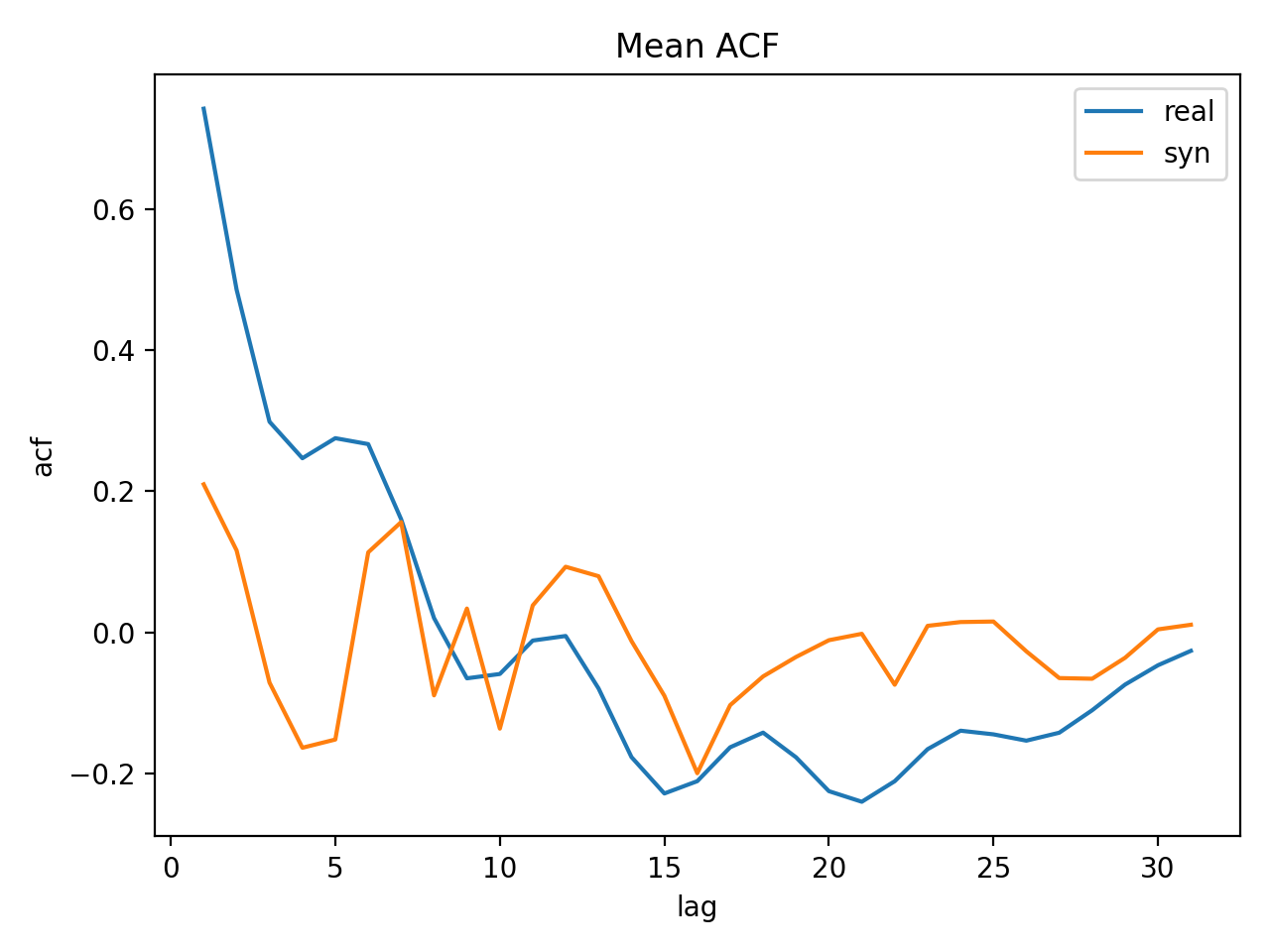}
                \label{fig:acf_ecg_diffusion}
            \end{subfigure}
            \hfill
            \begin{subfigure}{\subfigwidth}
                \centering
                \includegraphics[width=\linewidth]{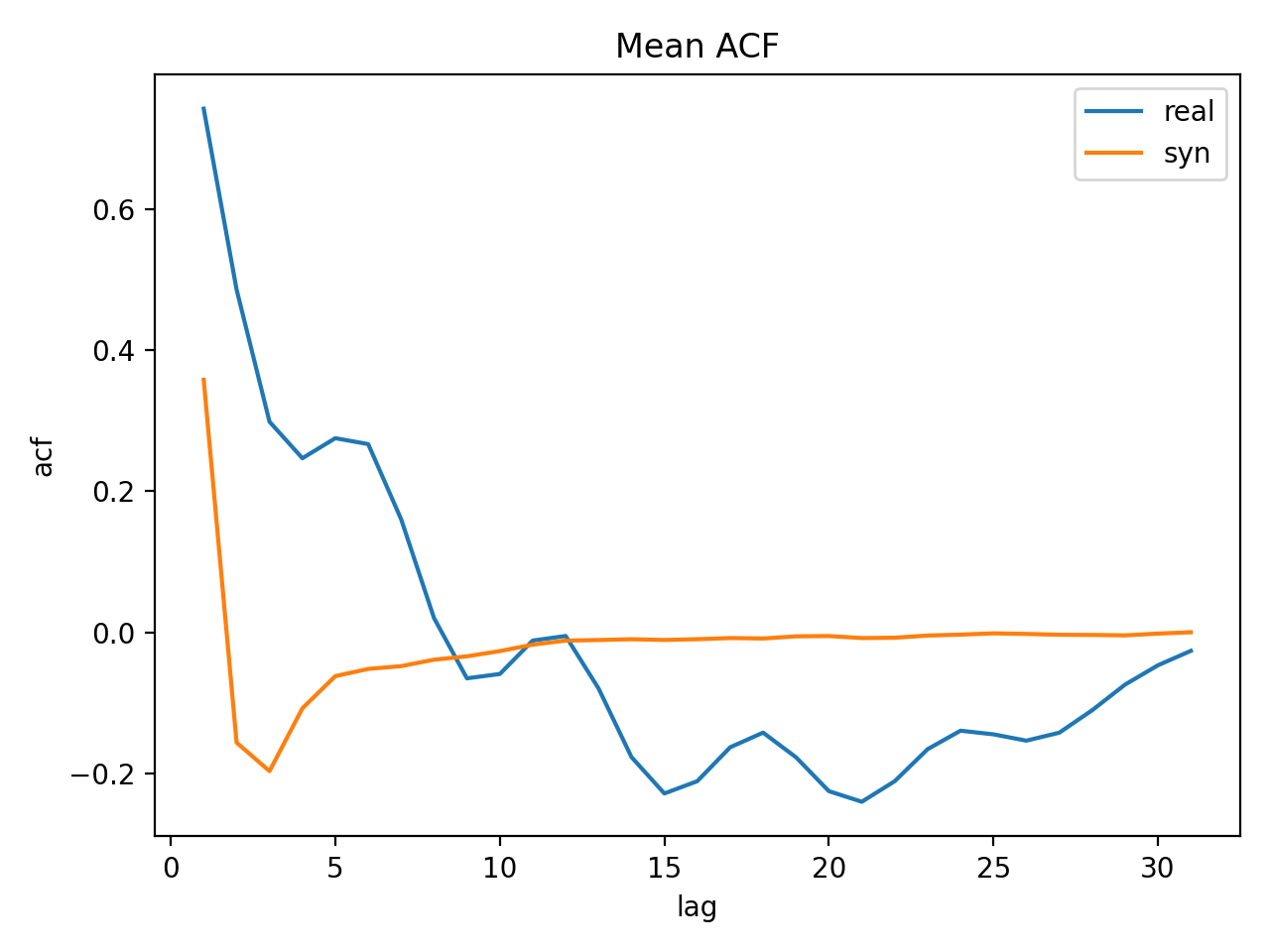}
                \label{fig:acf_ecg_timegan}
            \end{subfigure}
        \end{minipage}

        \vspace{0.2em}

        \begin{minipage}[c]{\rowlabelwidth}
            \quad
        \end{minipage}%
        \hfill
        \begin{minipage}[c]{\subfigcontainer}
            \begin{minipage}{\subfigwidth}
                \centering
                \textbf{(a) Graph2TS (ours)}
            \end{minipage}%
            \hfill
            \begin{minipage}{\subfigwidth}
                \centering
                \textbf{(b) DiffusionTS}
            \end{minipage}%
            \hfill
            \begin{minipage}{\subfigwidth}
                \centering
                \textbf{(c) TimeGAN}
            \end{minipage}
        \end{minipage}

    \end{minipage}%
    } 

    \caption{
    Mean autocorrelation functions (ACF) of generated time series on four datasets.
    \textbf{Rows (top to bottom):} CHB-MIT, Sunspot, Electricity, and ECG.
    \textbf{Columns:} Models.
    The proposed \textbf{Graph2TS} consistently preserves temporal dependency patterns of the real data across all datasets,
    while DiffusionTS exhibits over-smoothing and TimeGAN shows increased structural deviation.
    }
    \label{fig:acf_multi}
\end{figure}

\begin{figure}[h!]
    \centering

    \resizebox{0.90\textwidth}{!}{%
    \begin{minipage}{\textwidth}
    \centering

        \newcommand{\rowlabelwidth}{0.04\linewidth}
        \newcommand{\subfigcontainer}{0.95\linewidth}
        \newcommand{\subfigwidth}{0.31\linewidth}

        \begin{minipage}[c]{\rowlabelwidth}
            \centering
            \rotatebox{90}{\textbf{\large CHB-MIT}}
        \end{minipage}%
        \hfill
        \begin{minipage}[c]{\subfigcontainer}
            \begin{subfigure}{\subfigwidth}
                \centering
                \includegraphics[width=\linewidth]{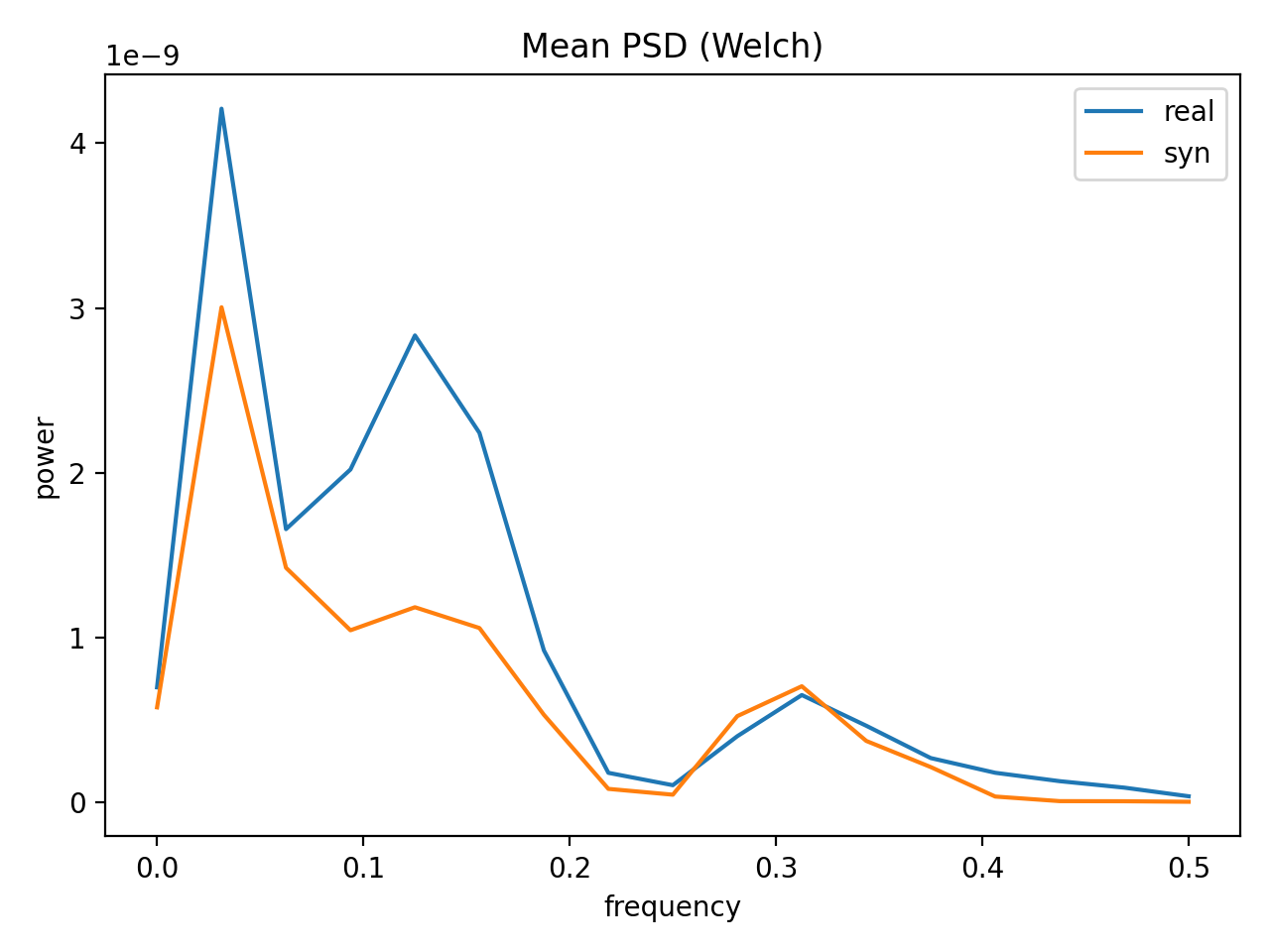}
                \label{fig:psd_chb_cvae}
            \end{subfigure}
            \hfill
            \begin{subfigure}{\subfigwidth}
                \centering
                \includegraphics[width=\linewidth]{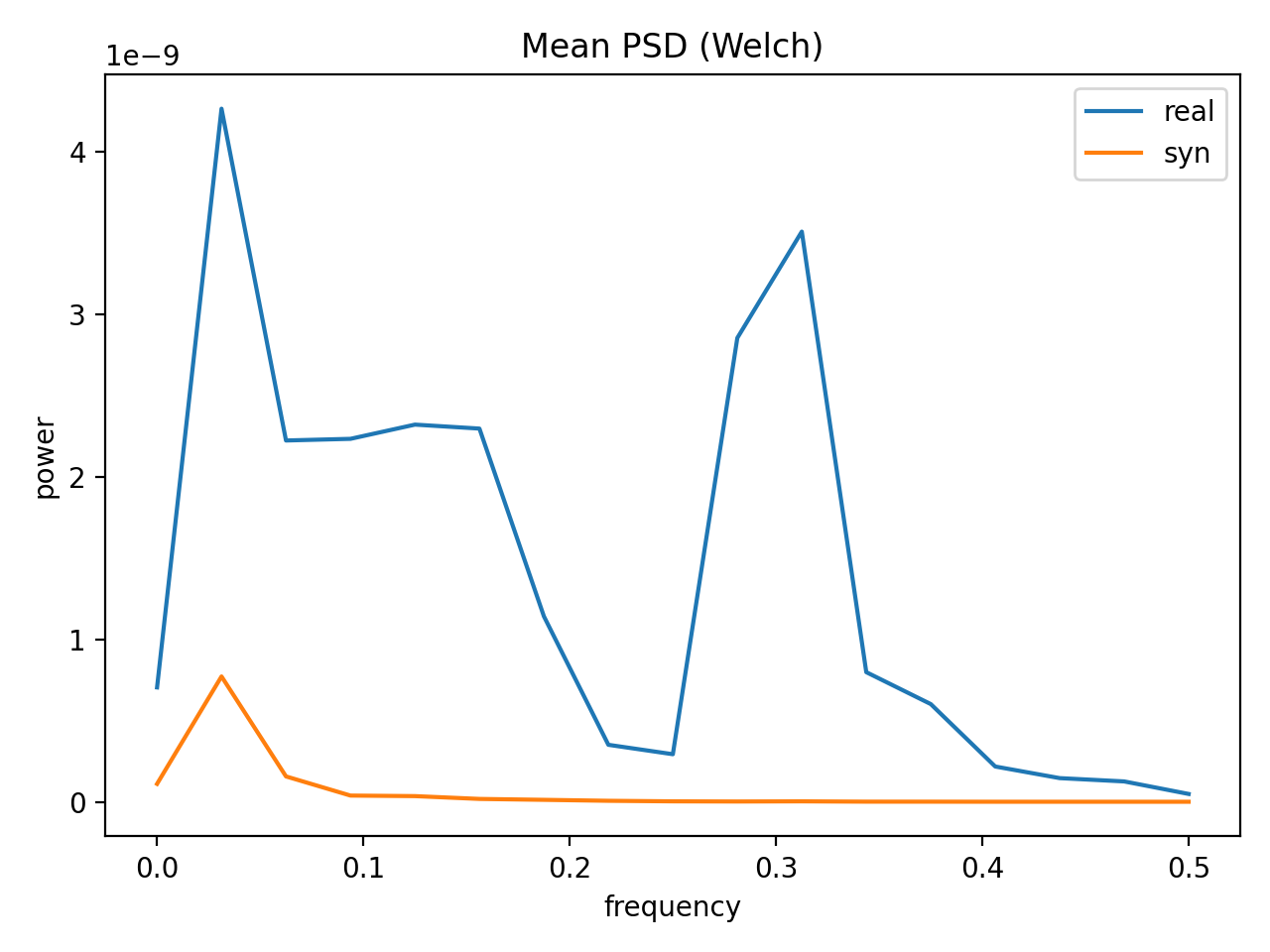}
                \label{fig:psd_chb_diffusion}
            \end{subfigure}
            \hfill
            \begin{subfigure}{\subfigwidth}
                \centering
                \includegraphics[width=\linewidth]{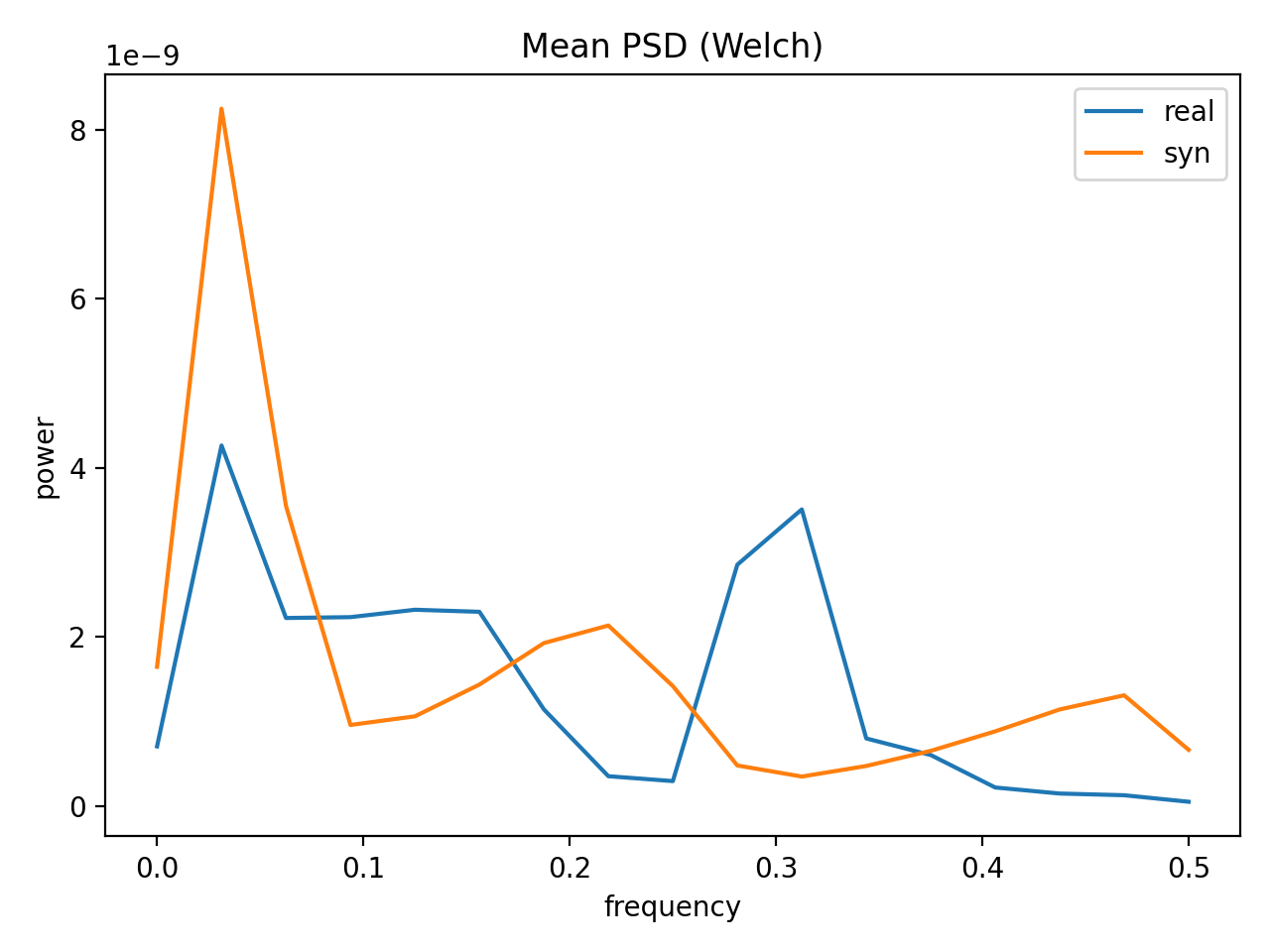}
                \label{fig:psd_chb_timegan}
            \end{subfigure}
        \end{minipage}

        \vspace{0.6em}

        \begin{minipage}[c]{\rowlabelwidth}
            \centering
            \rotatebox{90}{\textbf{\large Sunspot}}
        \end{minipage}%
        \hfill
        \begin{minipage}[c]{\subfigcontainer}
            \begin{subfigure}{\subfigwidth}
                \centering
                \includegraphics[width=\linewidth]{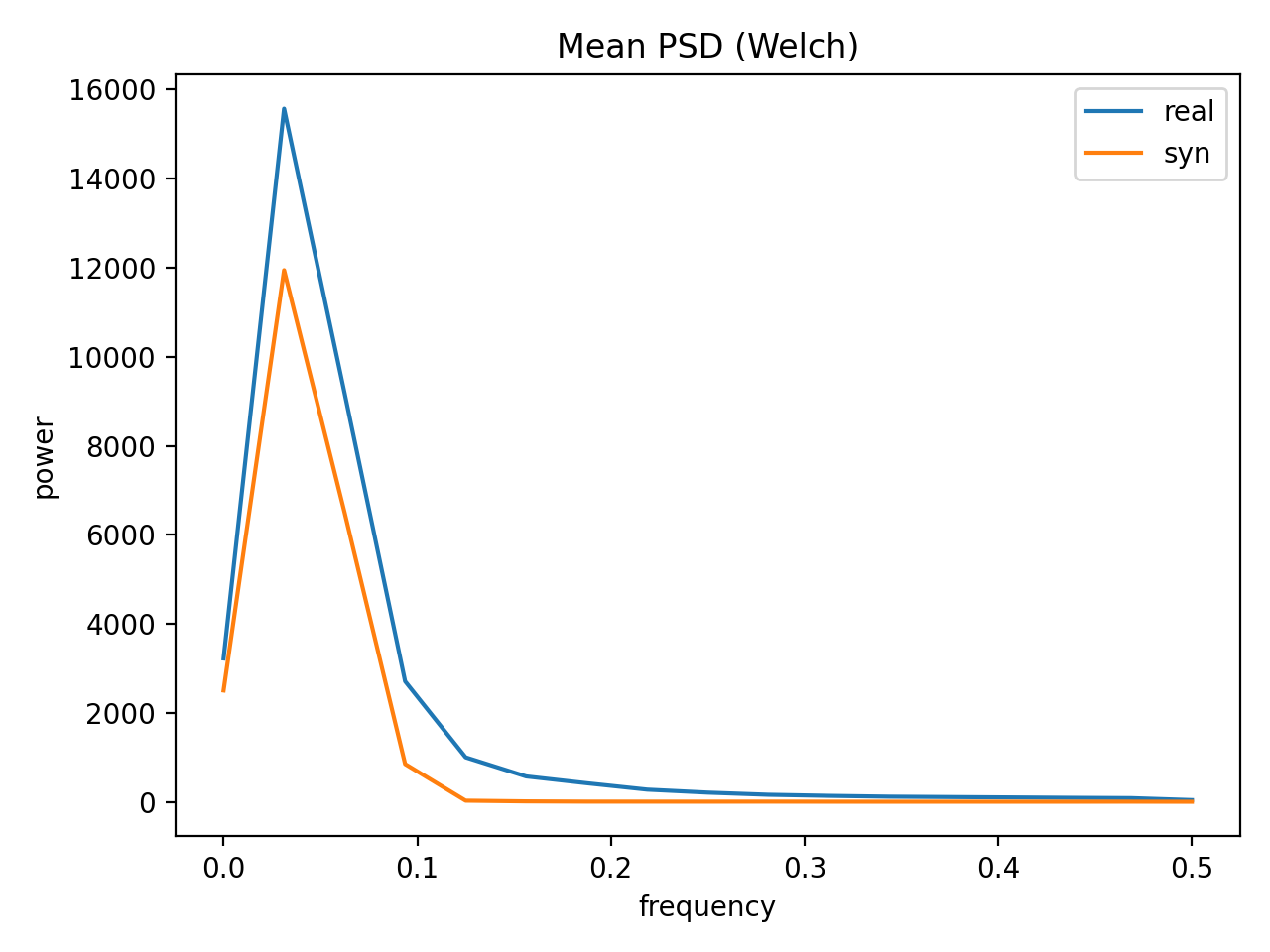}
                \label{fig:psd_sunspot_cvae}
            \end{subfigure}
            \hfill
            \begin{subfigure}{\subfigwidth}
                \centering
                \includegraphics[width=\linewidth]{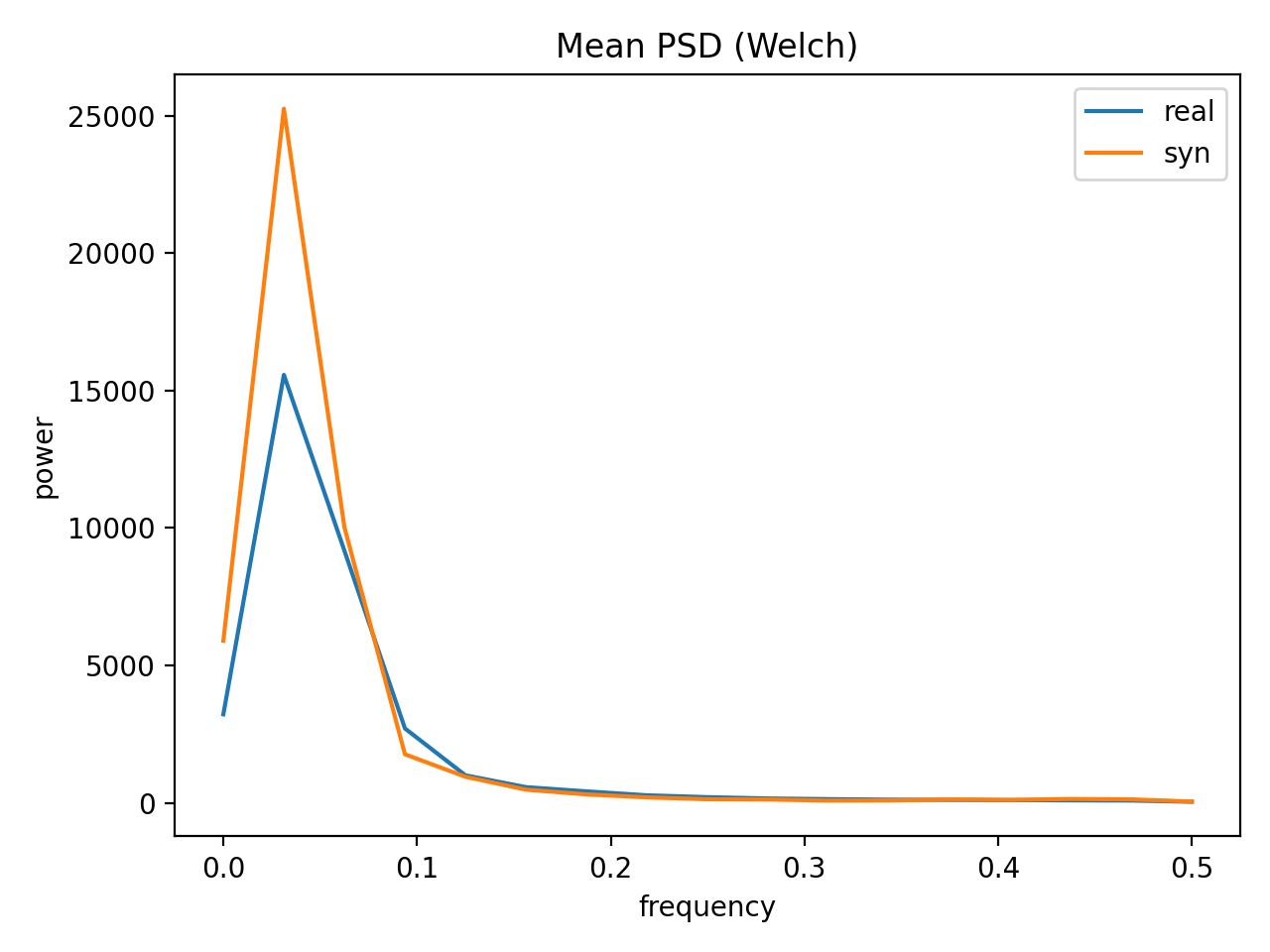}
                \label{fig:psd_sunspot_diffusion}
            \end{subfigure}
            \hfill
            \begin{subfigure}{\subfigwidth}
                \centering
                \includegraphics[width=\linewidth]{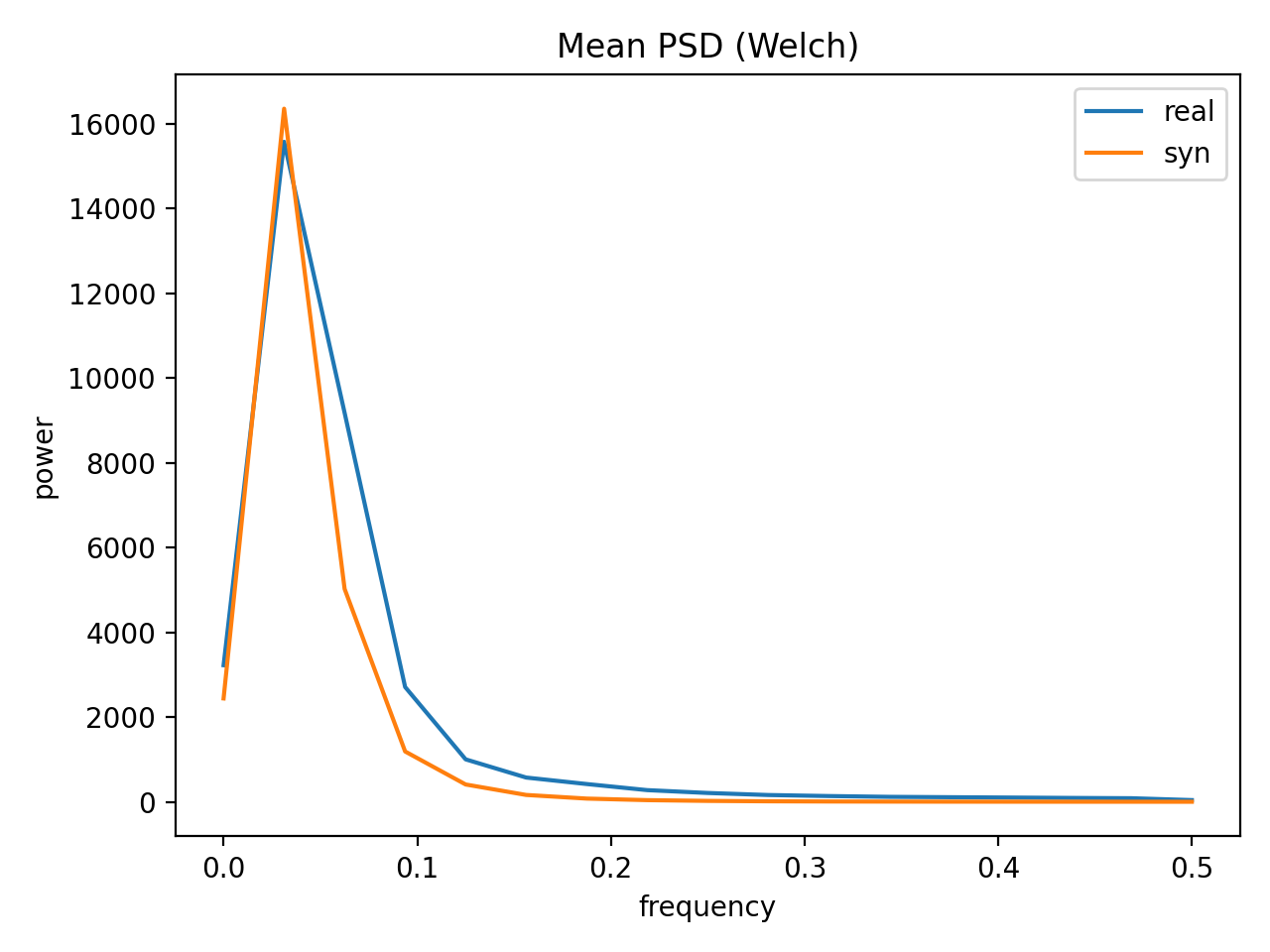}
                \label{fig:psd_sunspot_timegan}
            \end{subfigure}
        \end{minipage}

        \vspace{0.6em}

        \begin{minipage}[c]{\rowlabelwidth}
            \centering
            \rotatebox{90}{\textbf{\large Electricity}}
        \end{minipage}%
        \hfill
        \begin{minipage}[c]{\subfigcontainer}
            \begin{subfigure}{\subfigwidth}
                \centering
                \includegraphics[width=\linewidth]{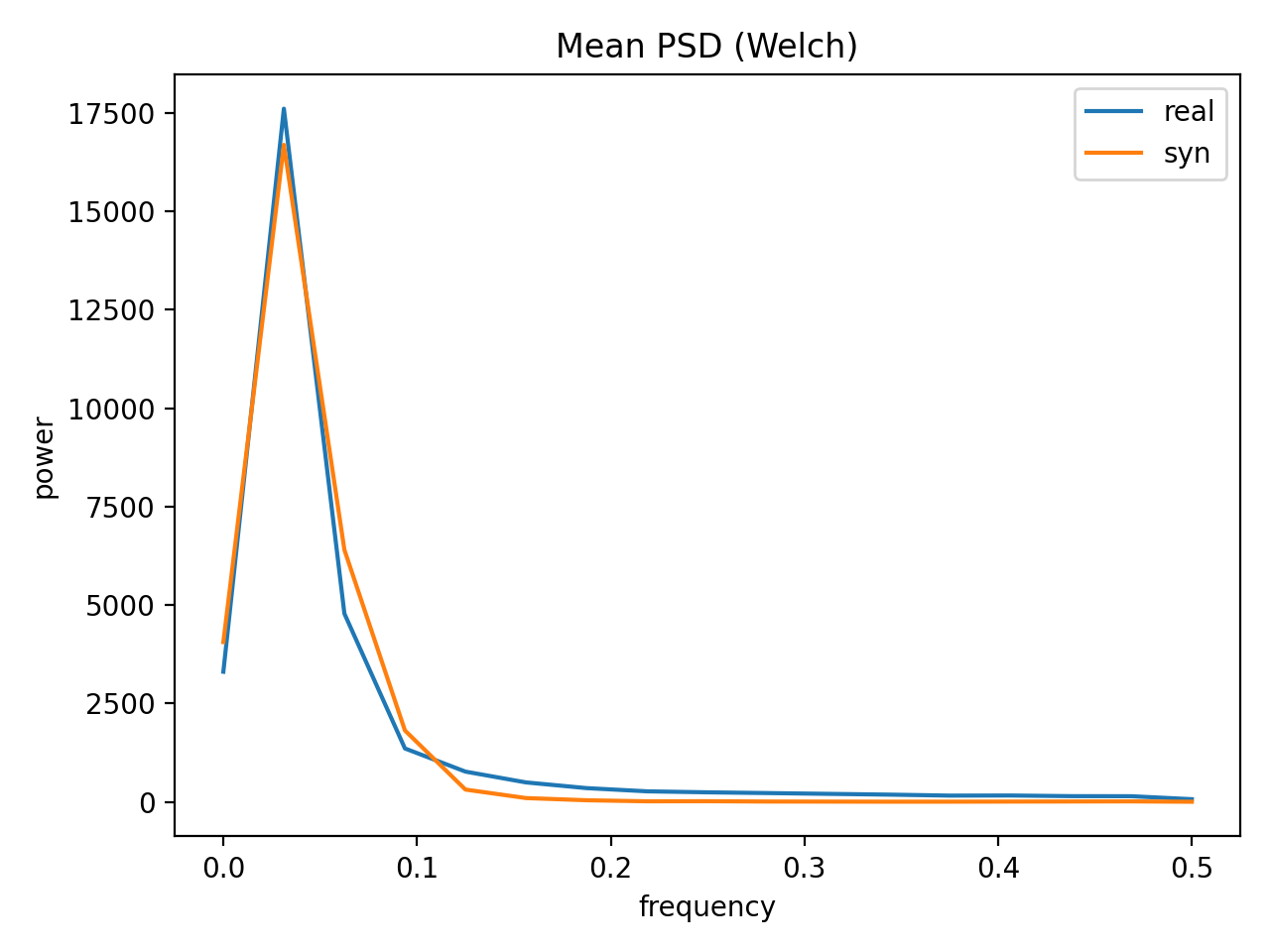}
                \label{fig:psd_elec_cvae}
            \end{subfigure}
            \hfill
            \begin{subfigure}{\subfigwidth}
                \centering
                \includegraphics[width=\linewidth]{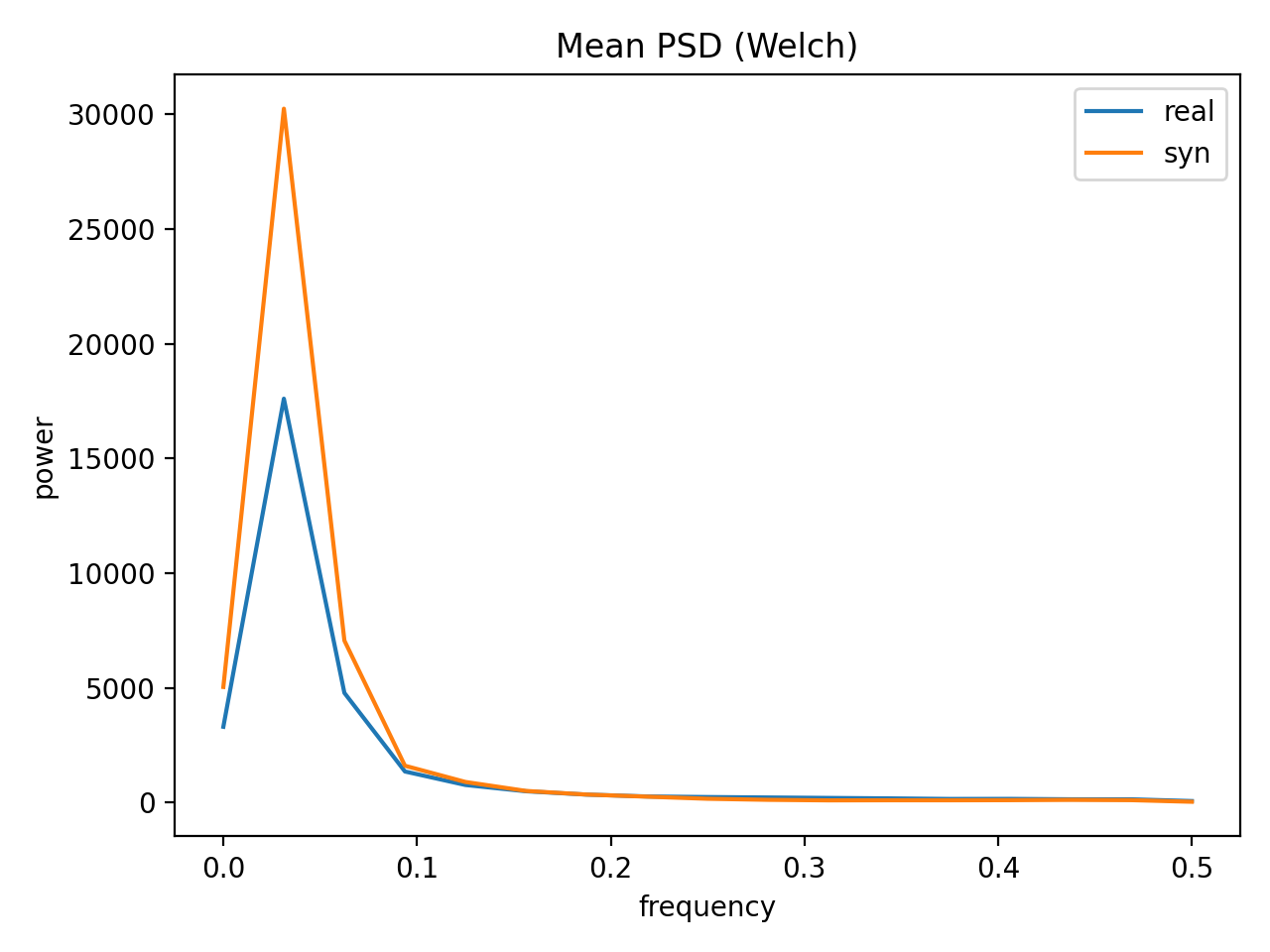}
                \label{fig:psd_elec_diffusion}
            \end{subfigure}
            \hfill
            \begin{subfigure}{\subfigwidth}
                \centering
                \includegraphics[width=\linewidth]{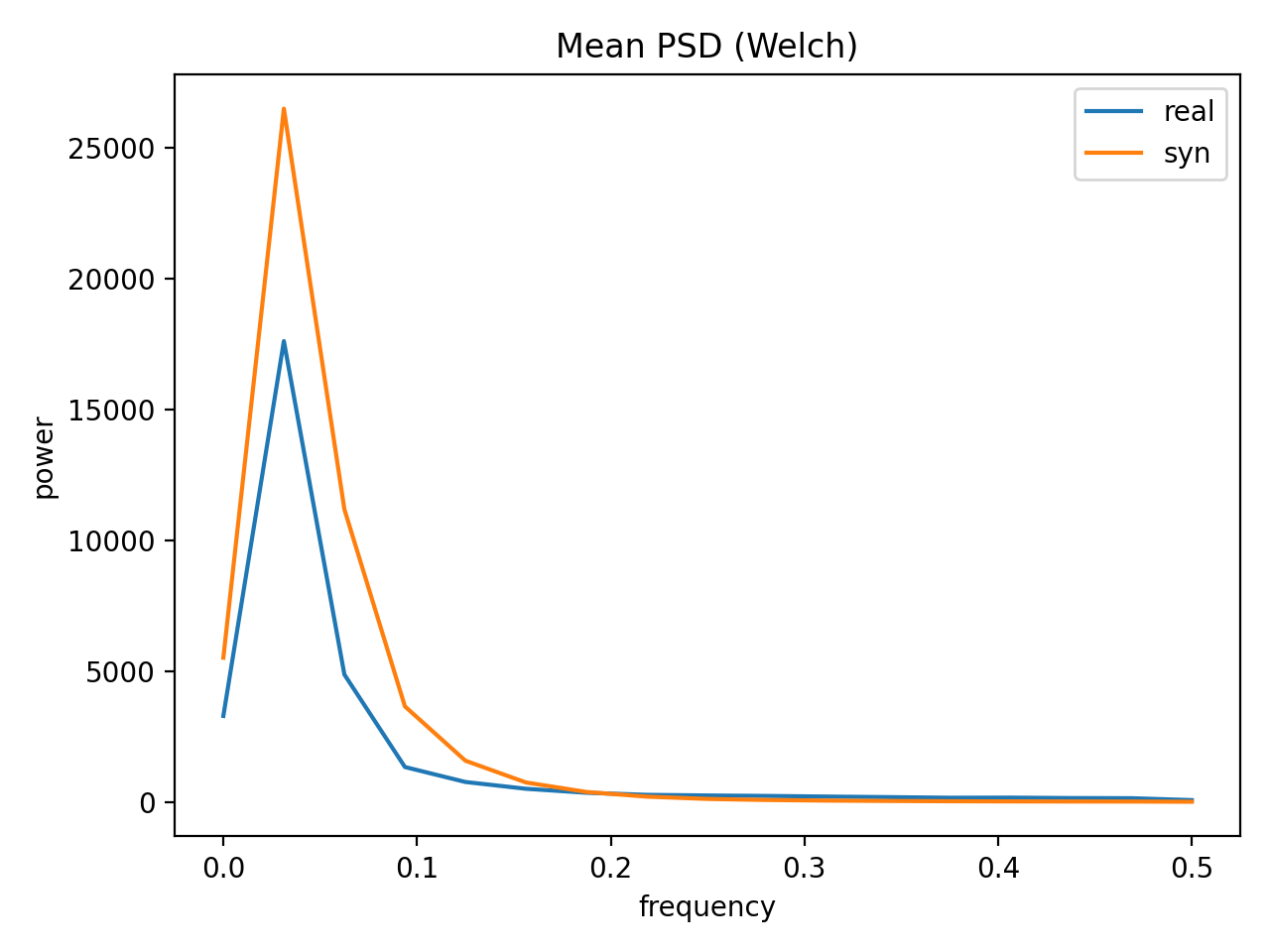}
                \label{fig:psd_elec_timegan}
            \end{subfigure}
        \end{minipage}

        \vspace{0.6em}

        \begin{minipage}[c]{\rowlabelwidth}
            \centering
            \rotatebox{90}{\textbf{\large ECG}}
        \end{minipage}%
        \hfill
        \begin{minipage}[c]{\subfigcontainer}
            \begin{subfigure}{\subfigwidth}
                \centering
                \includegraphics[width=\linewidth]{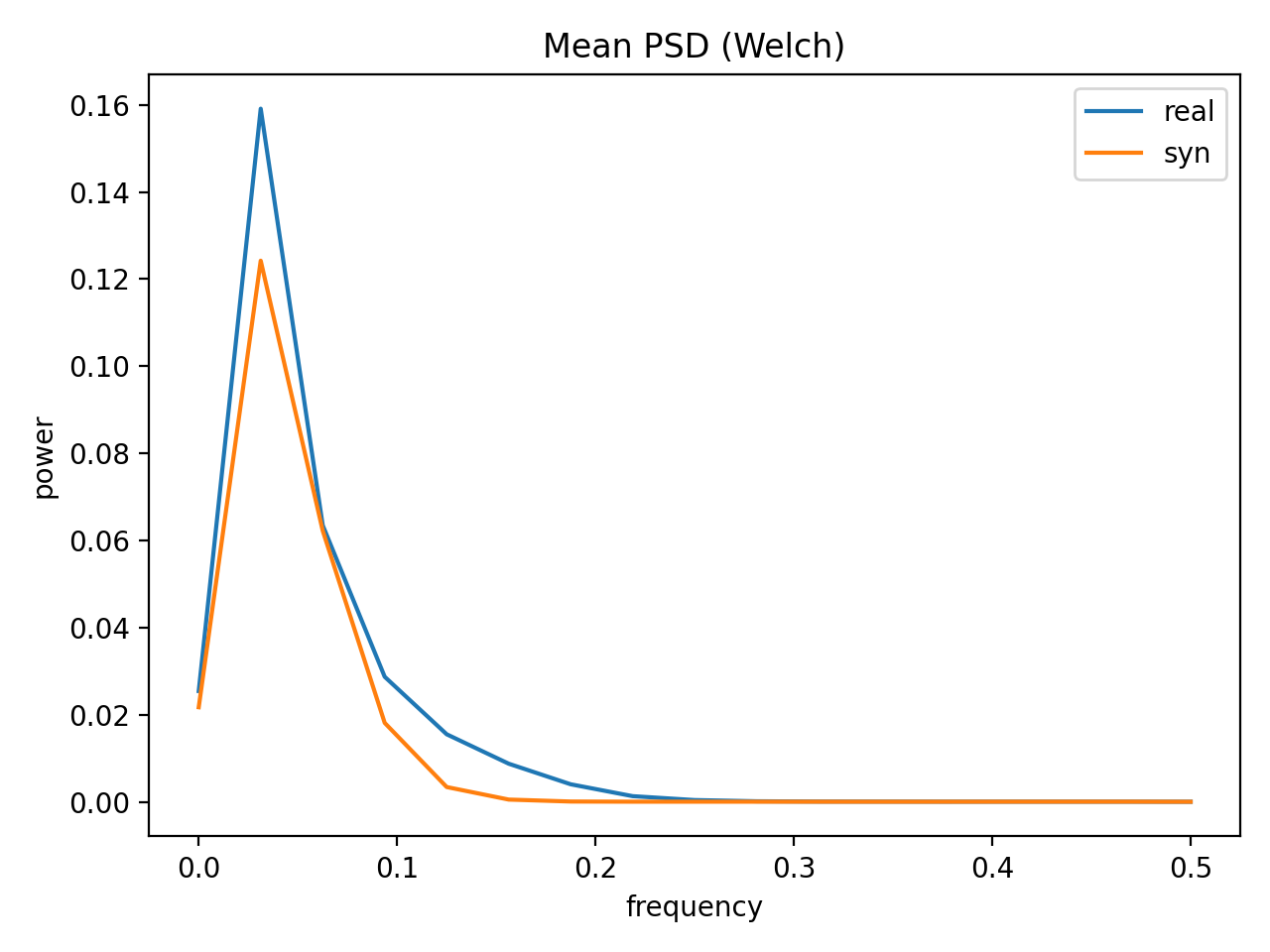}
                \label{fig:psd_ecg_cvae}
            \end{subfigure}
            \hfill
            \begin{subfigure}{\subfigwidth}
                \centering
                \includegraphics[width=\linewidth]{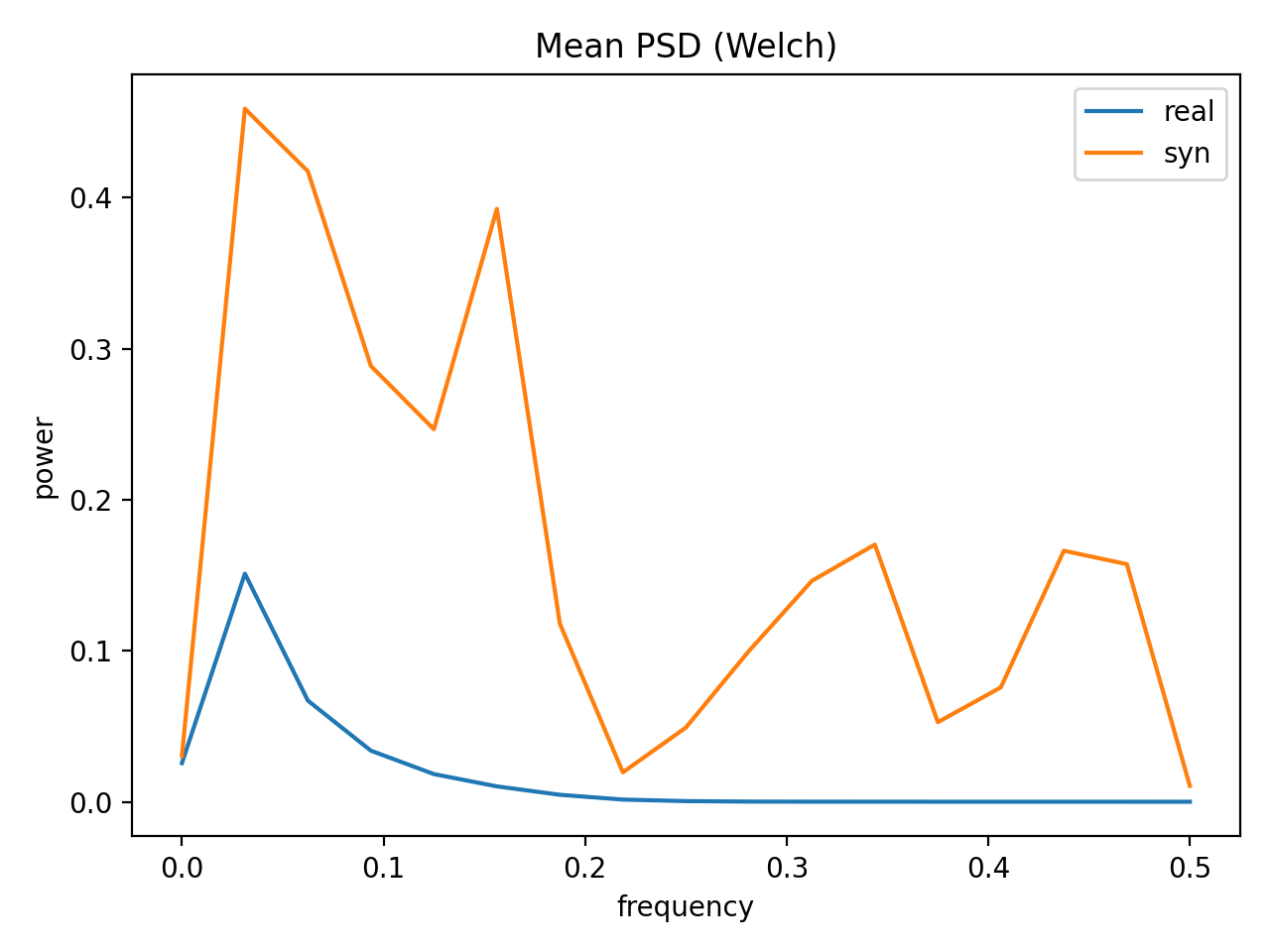}
                \label{fig:psd_ecg_diffusion}
            \end{subfigure}
            \hfill
            \begin{subfigure}{\subfigwidth}
                \centering
                \includegraphics[width=\linewidth]{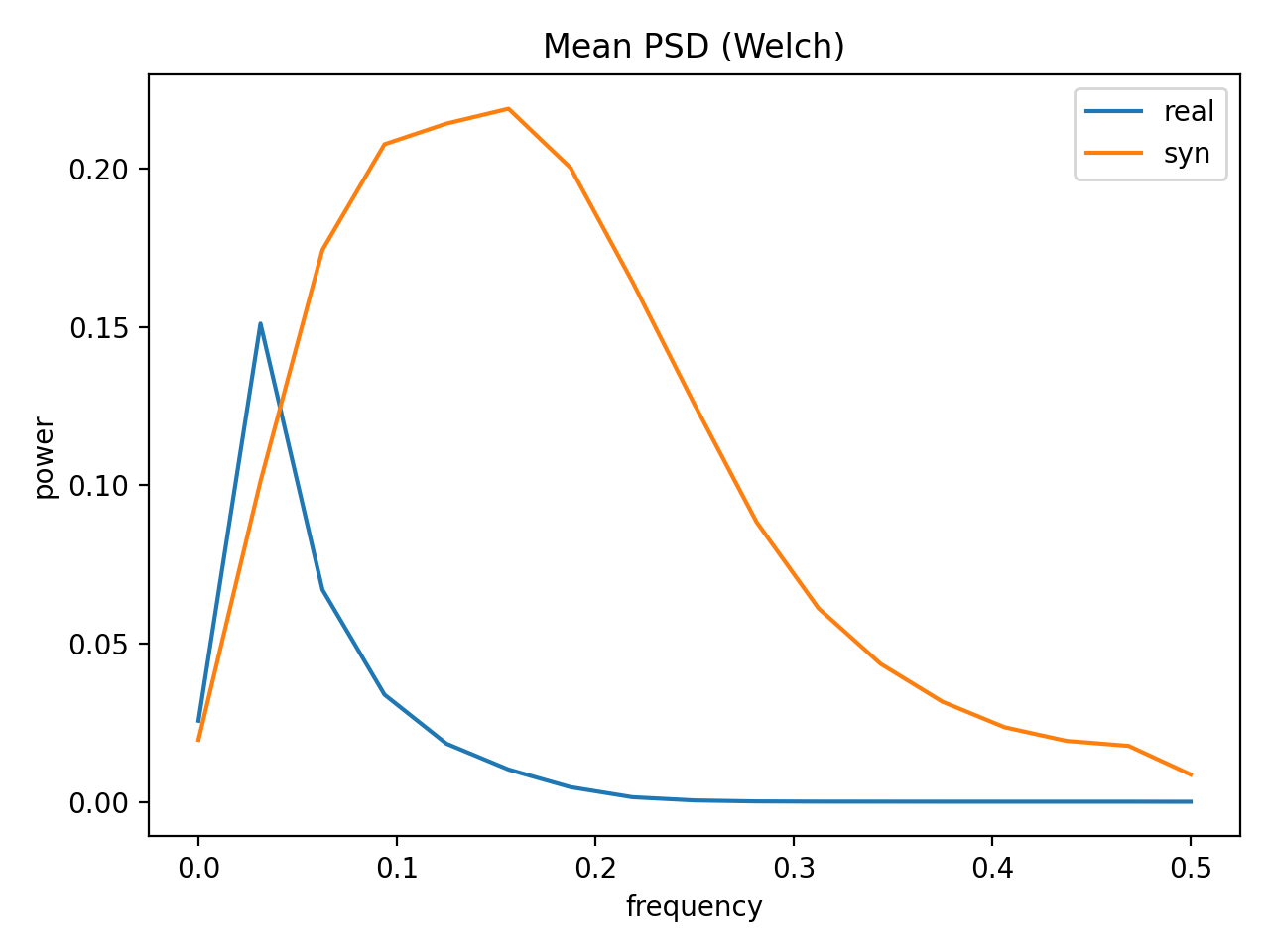}
                \label{fig:psd_ecg_timegan}
            \end{subfigure}
        \end{minipage}

        \vspace{0.5em}

        \begin{minipage}[c]{\rowlabelwidth}
            \quad
        \end{minipage}%
        \hfill
        \begin{minipage}[c]{\subfigcontainer}
            \begin{minipage}{\subfigwidth}
                \centering
                \textbf{(a) Graph2TS (ours)}
            \end{minipage}%
            \hfill
            \begin{minipage}{\subfigwidth}
                \centering
                \textbf{(b) DiffusionTS}
            \end{minipage}%
            \hfill
            \begin{minipage}{\subfigwidth}
                \centering
                \textbf{(c) TimeGAN}
            \end{minipage}
        \end{minipage}

    \end{minipage}%
    } 

    \caption{
    Mean power spectral density (PSD) of generated time series on four datasets.
    \textbf{Rows (top to bottom):} CHB-MIT, Sunspot, Electricity, and ECG.
    The proposed model \textbf{Graph2TS} closely matches the spectral profile of real data across all datasets,
    whereas DiffusionTS tends to attenuate high-frequency components and
    TimeGAN exhibits spectral distortion.
    }
    \label{fig:psd_multi}
\end{figure}
\begin{figure}[h!]
    \centering

    \resizebox{0.98\textwidth}{!}{%
    \begin{minipage}{\textwidth}
    \centering

        \newcommand{\rowlabelwidth}{0.04\linewidth}
        \newcommand{\subfigcontainer}{0.95\linewidth}
        \newcommand{\subfigwidth}{0.31\linewidth}

        \begin{minipage}[c]{\rowlabelwidth}
            \centering
            \rotatebox{90}{\textbf{\large CHB-MIT}}
        \end{minipage}%
        \hfill
        \begin{minipage}[c]{\subfigcontainer}
            \begin{subfigure}{\subfigwidth}
                \centering
                \includegraphics[width=\linewidth]{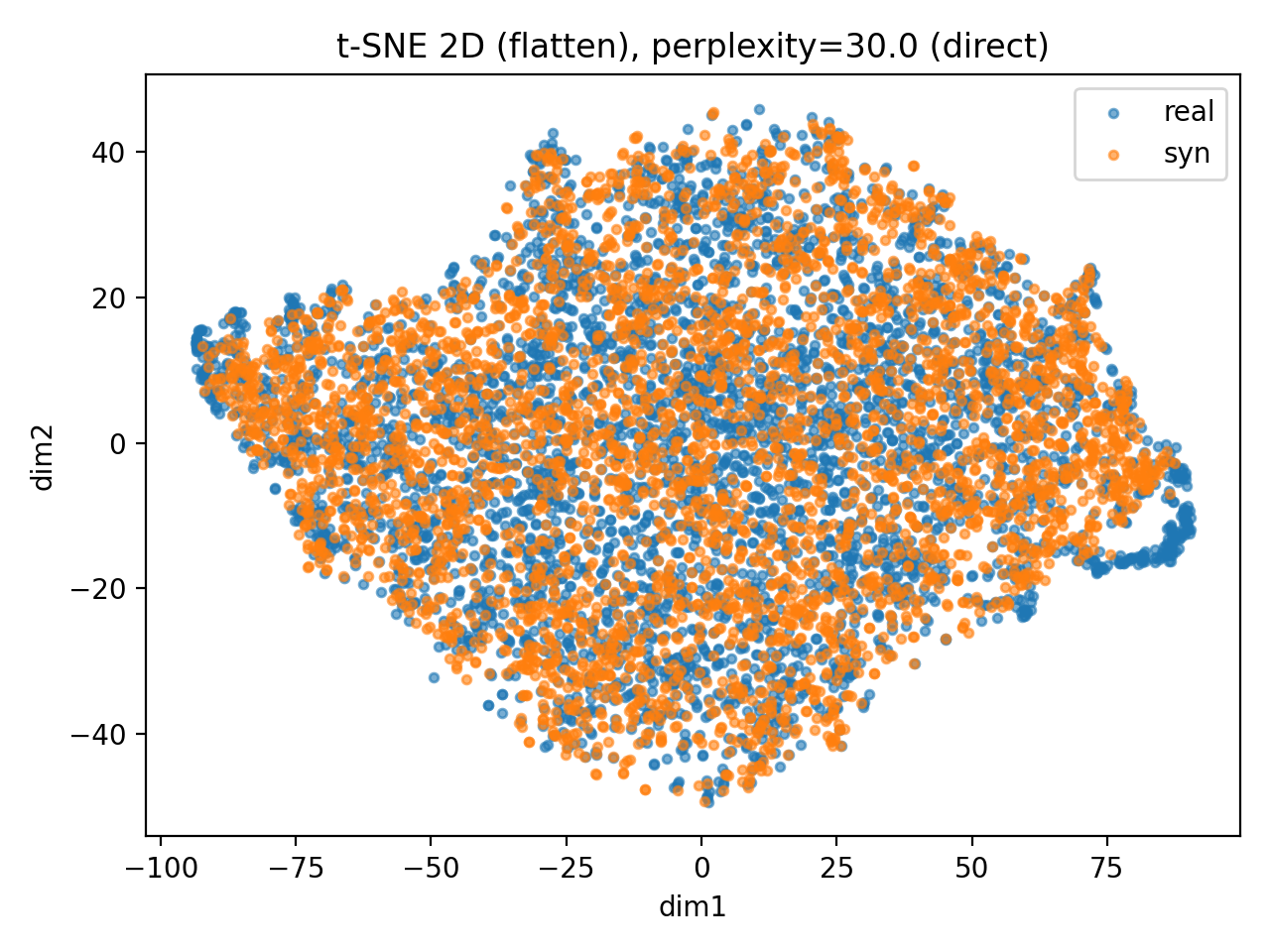}
                \label{fig:tsne_chb_cvae}
            \end{subfigure}
            \hfill
            \begin{subfigure}{\subfigwidth}
                \centering
                \includegraphics[width=\linewidth]{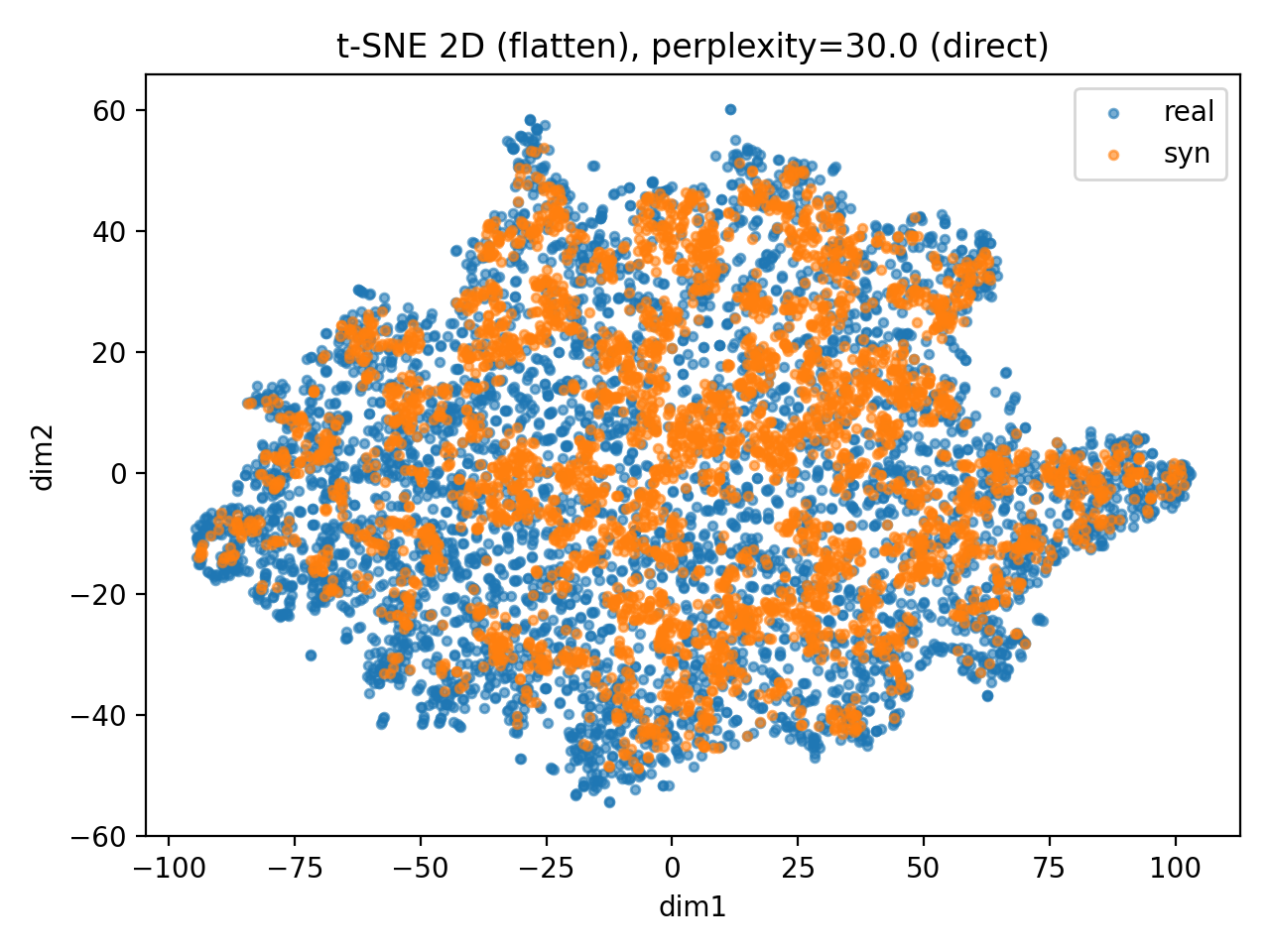}
                \label{fig:tsne_chb_diffusion}
            \end{subfigure}
            \hfill
            \begin{subfigure}{\subfigwidth}
                \centering
                \includegraphics[width=\linewidth]{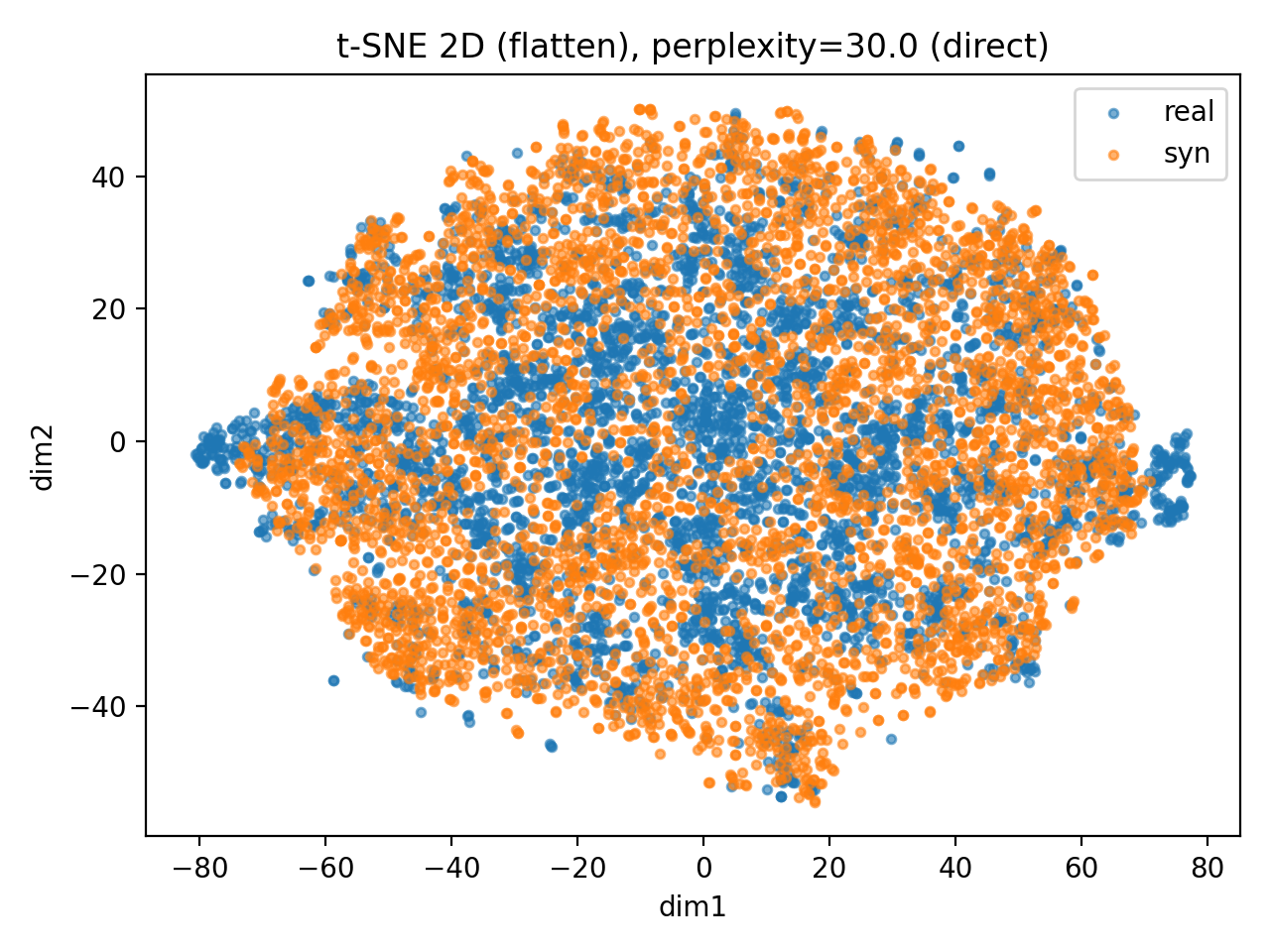}
                \label{fig:tsne_chb_timegan}
            \end{subfigure}
        \end{minipage}

        \vspace{0.6em}

        \begin{minipage}[c]{\rowlabelwidth}
            \centering
            \rotatebox{90}{\textbf{\large Sunspot}}
        \end{minipage}%
        \hfill
        \begin{minipage}[c]{\subfigcontainer}
            \begin{subfigure}{\subfigwidth}
                \centering
                \includegraphics[width=\linewidth]{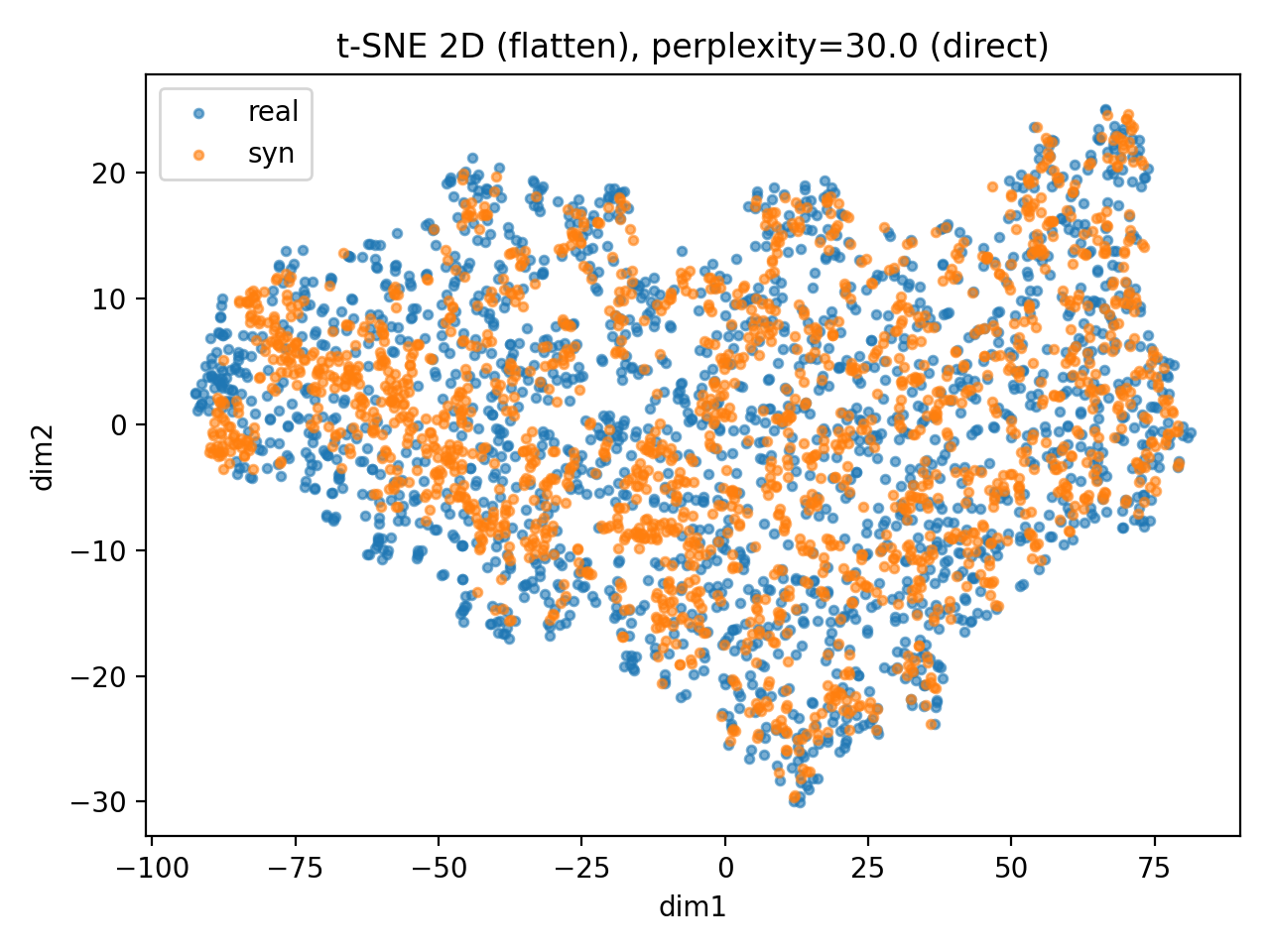}
                \label{fig:tsne_sunspot_cvae}
            \end{subfigure}
            \hfill
            \begin{subfigure}{\subfigwidth}
                \centering
                \includegraphics[width=\linewidth]{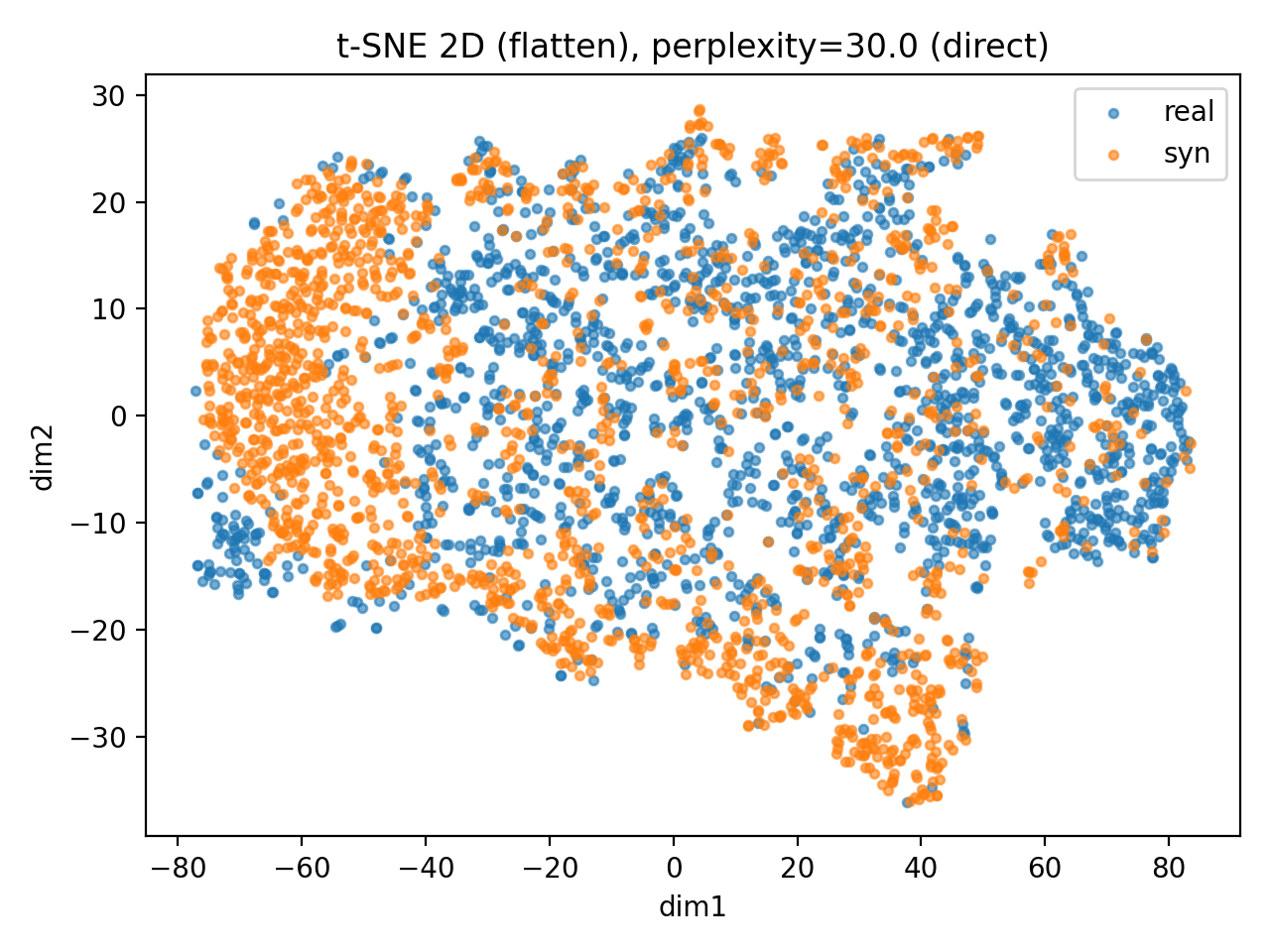}
                \label{fig:tsne_sunspot_diffusion}
            \end{subfigure}
            \hfill
            \begin{subfigure}{\subfigwidth}
                \centering
                \includegraphics[width=\linewidth]{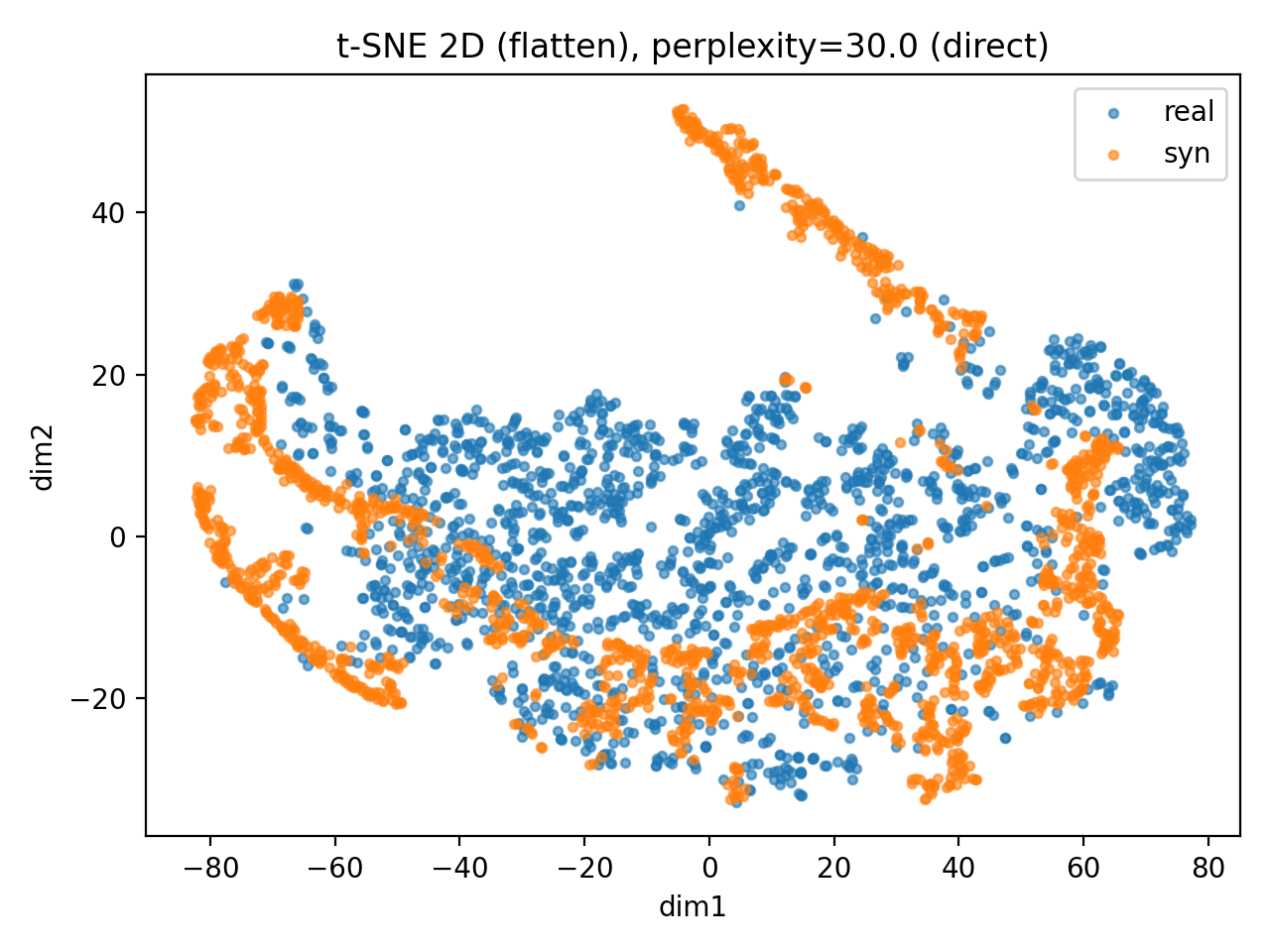}
                \label{fig:tsne_sunspot_timegan}
            \end{subfigure}
        \end{minipage}

        \vspace{0.6em}

        \begin{minipage}[c]{\rowlabelwidth}
            \centering
            \rotatebox{90}{\textbf{\large Electricity}}
        \end{minipage}%
        \hfill
        \begin{minipage}[c]{\subfigcontainer}
            \begin{subfigure}{\subfigwidth}
                \centering
                \includegraphics[width=\linewidth]{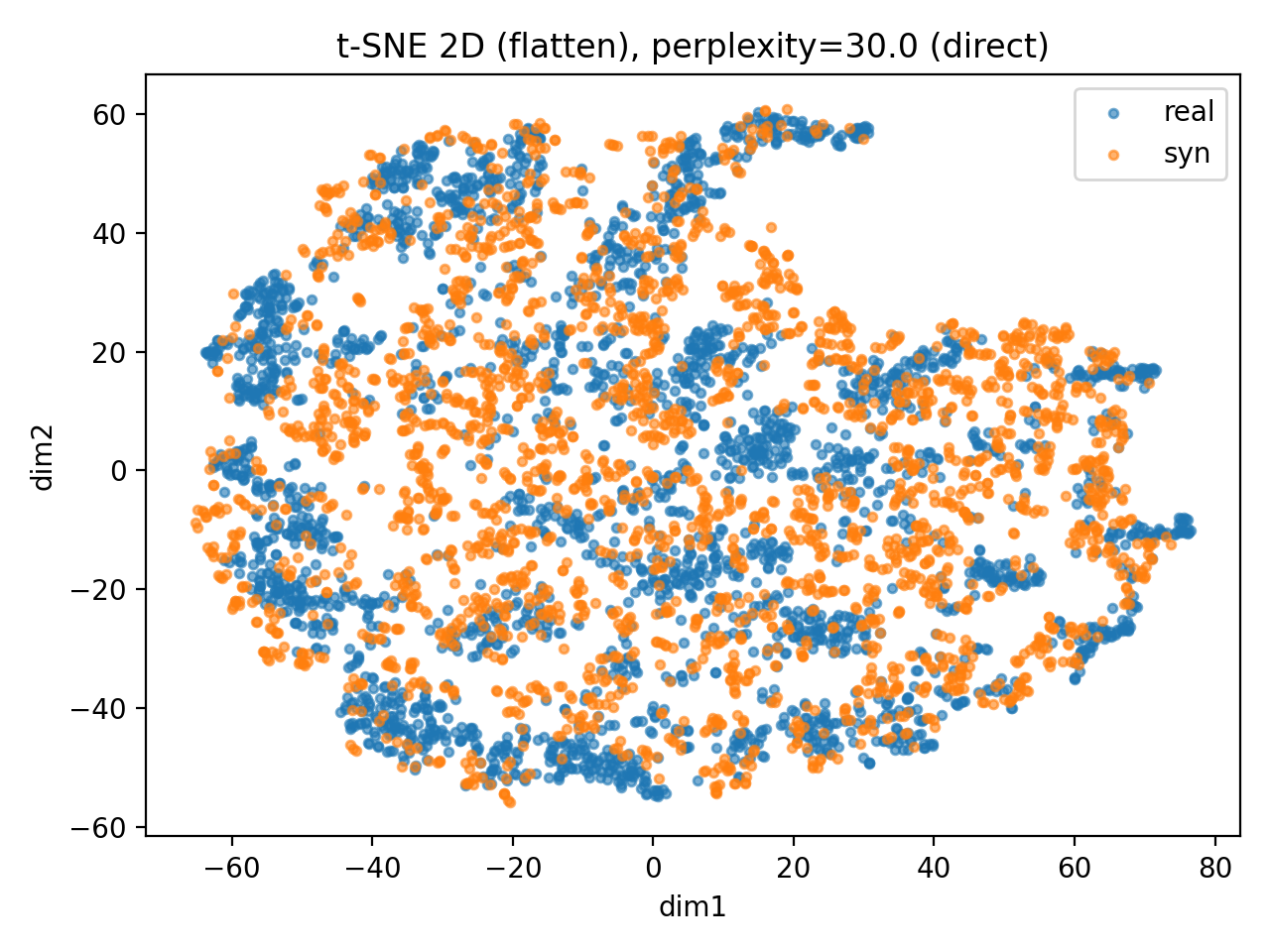}
                \label{fig:tsne_elec_cvae}
            \end{subfigure}
            \hfill
            \begin{subfigure}{\subfigwidth}
                \centering
                \includegraphics[width=\linewidth]{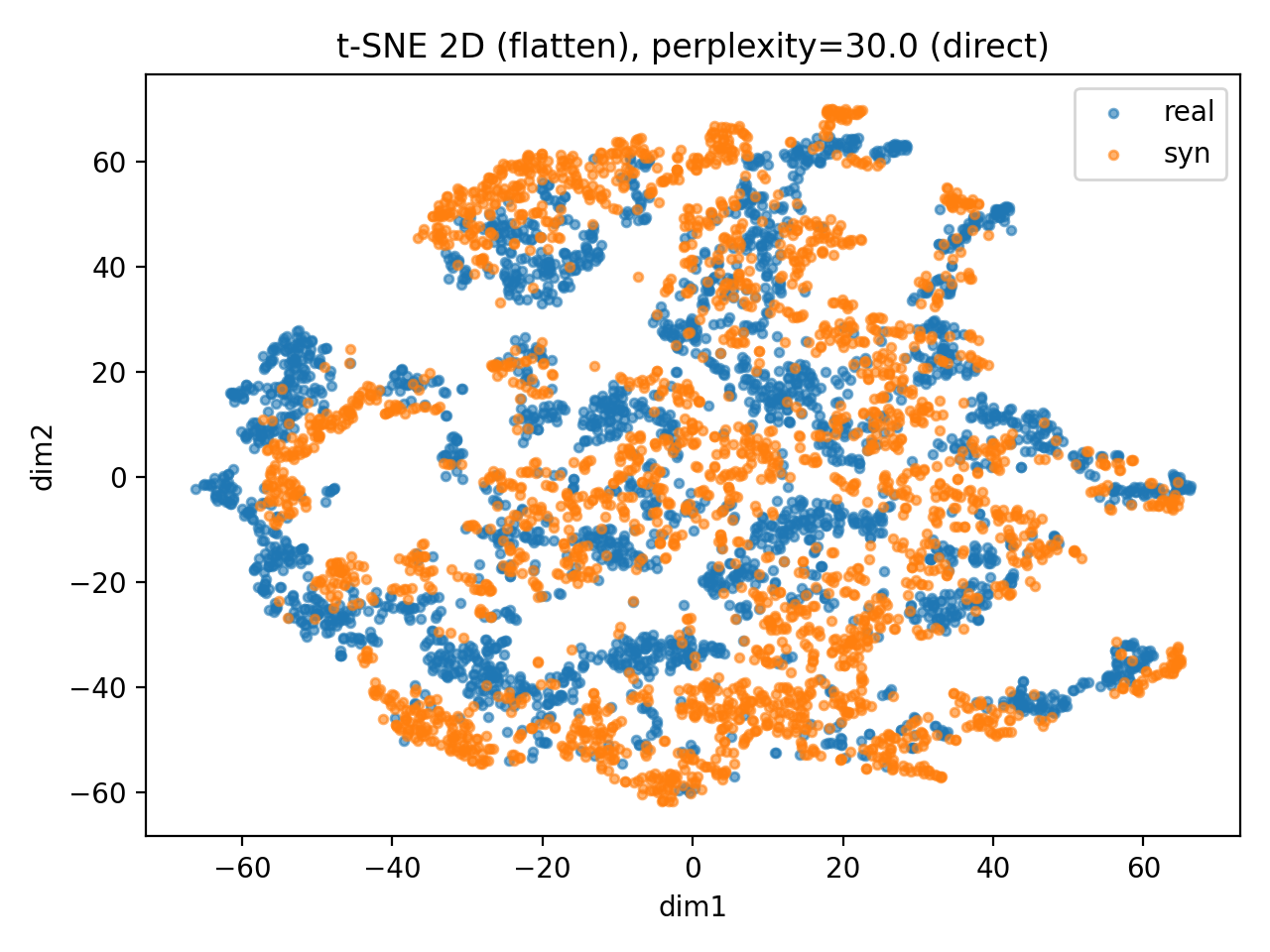}
                \label{fig:tsne_elec_diffusion}
            \end{subfigure}
            \hfill
            \begin{subfigure}{\subfigwidth}
                \centering
                \includegraphics[width=\linewidth]{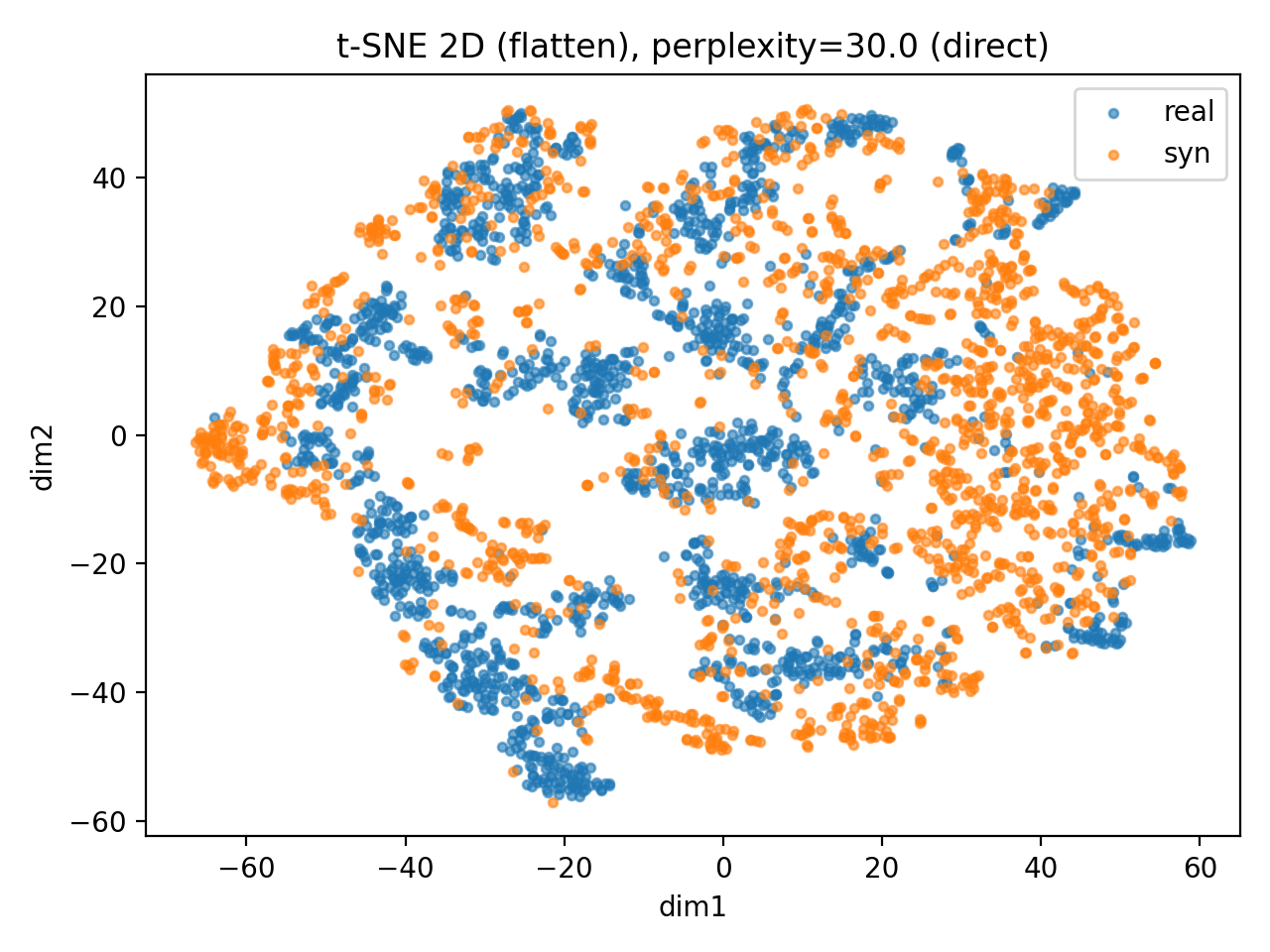}
                \label{fig:tsne_elec_timegan}
            \end{subfigure}
        \end{minipage}

        \vspace{0.6em}

        \begin{minipage}[c]{\rowlabelwidth}
            \centering
            \rotatebox{90}{\textbf{\large ECG}}
        \end{minipage}%
        \hfill
        \begin{minipage}[c]{\subfigcontainer}
            \begin{subfigure}{\subfigwidth}
                \centering
                \includegraphics[width=\linewidth]{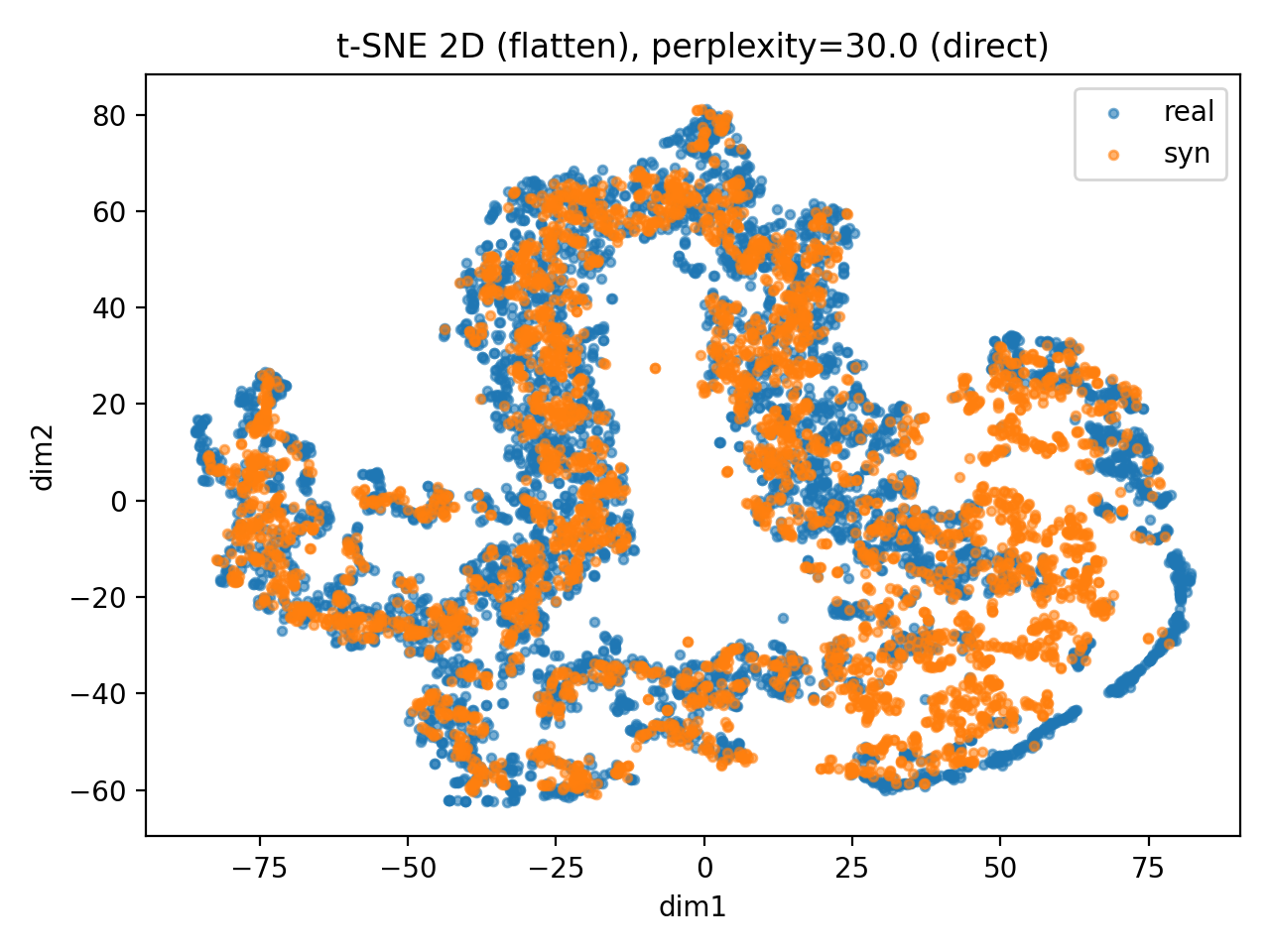}
                \label{fig:tsne_ecg_cvae}
            \end{subfigure}
            \hfill
            \begin{subfigure}{\subfigwidth}
                \centering
                \includegraphics[width=\linewidth]{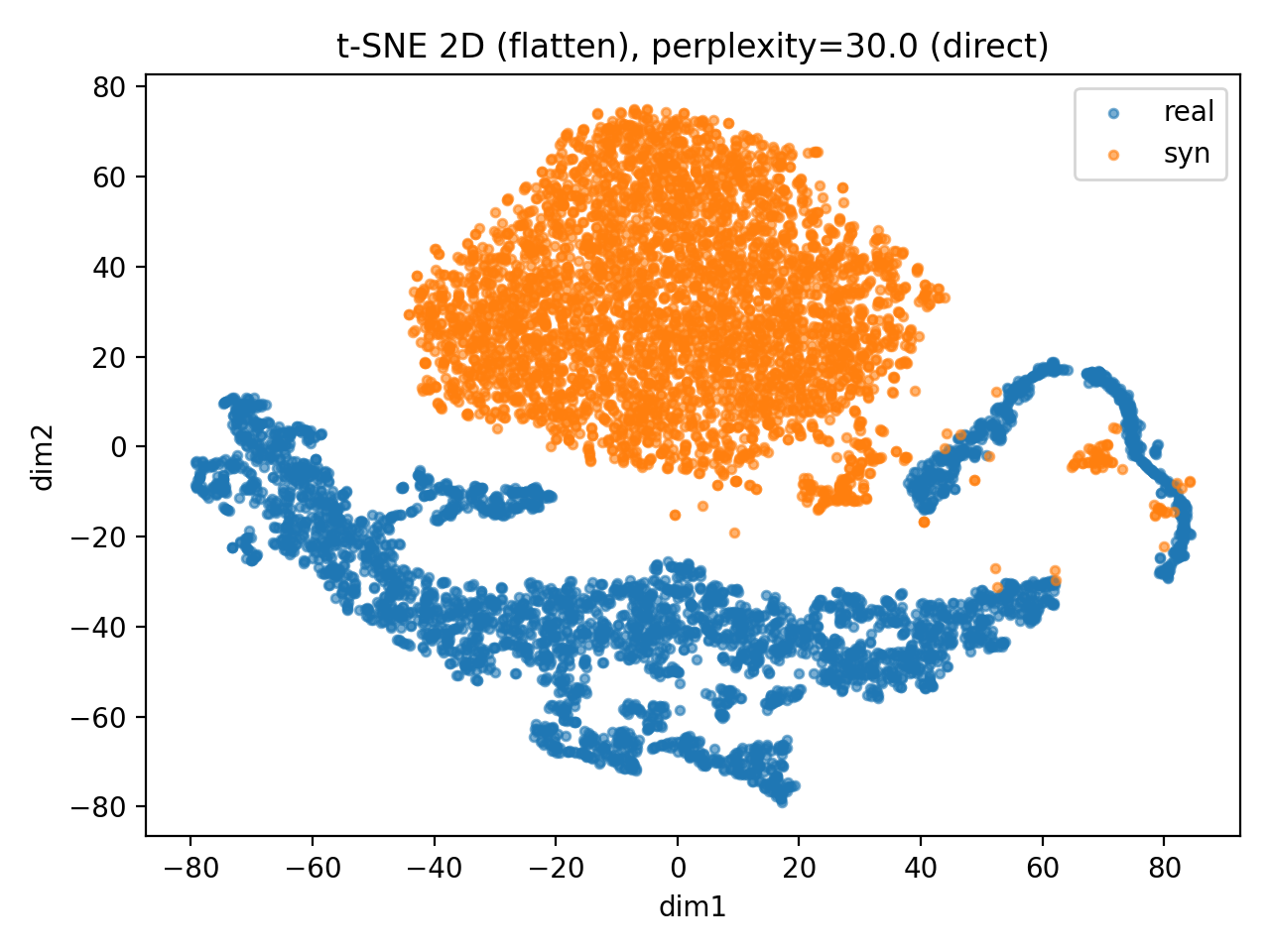}
                \label{fig:tsne_ecg_diffusion}
            \end{subfigure}
            \hfill
            \begin{subfigure}{\subfigwidth}
                \centering
                \includegraphics[width=\linewidth]{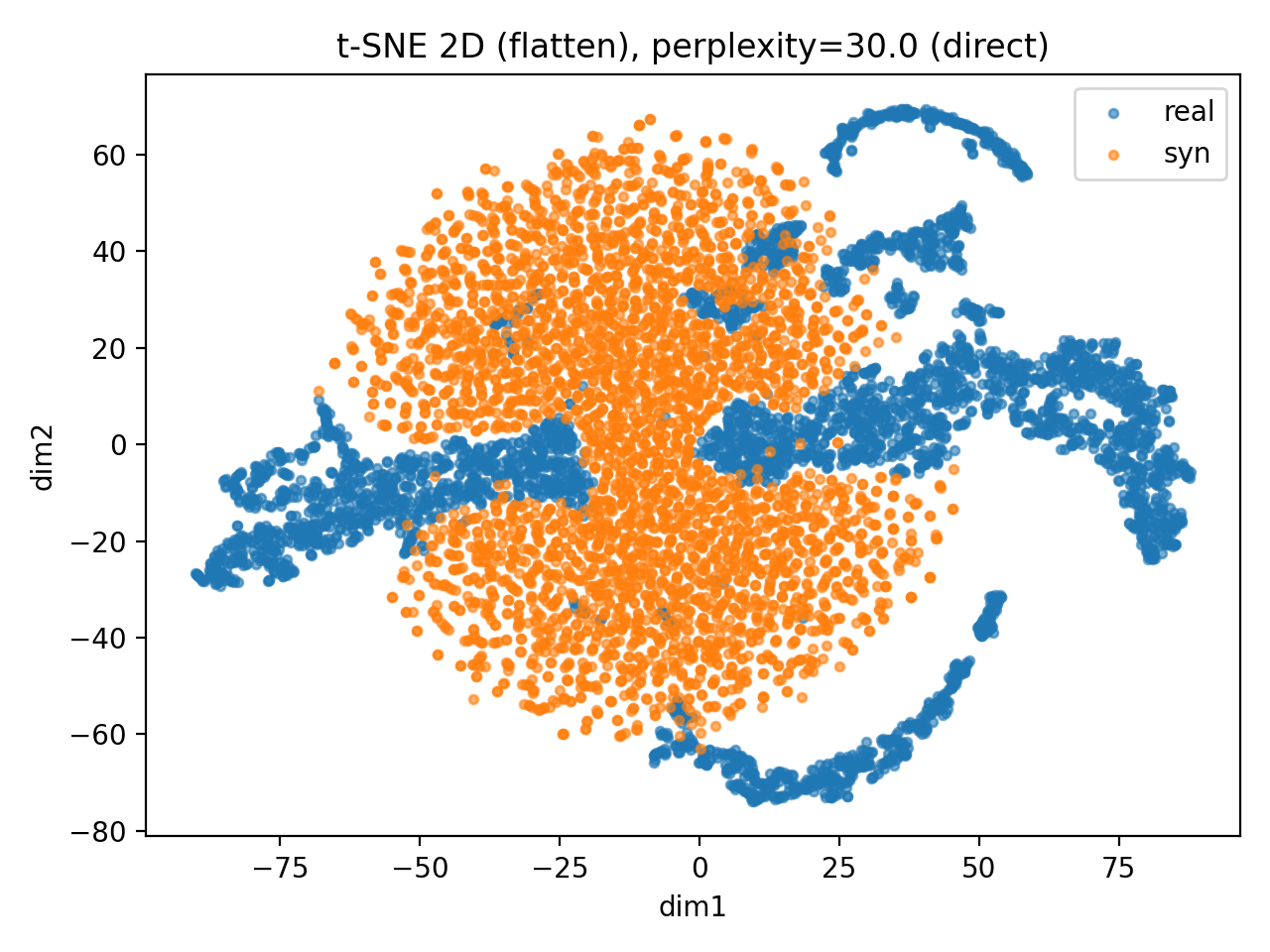}
                \label{fig:tsne_ecg_timegan}
            \end{subfigure}
        \end{minipage}

        \vspace{0.5em}

        \begin{minipage}[c]{\rowlabelwidth}
            \quad
        \end{minipage}%
        \hfill
        \begin{minipage}[c]{\subfigcontainer}
            \begin{minipage}{\subfigwidth}
                \centering
                \textbf{(a) Graph2TS (ours)}
            \end{minipage}%
            \hfill
            \begin{minipage}{\subfigwidth}
                \centering
                \textbf{(b) DiffusionTS}
            \end{minipage}%
            \hfill
            \begin{minipage}{\subfigwidth}
                \centering
                \textbf{(c) TimeGAN}
            \end{minipage}
        \end{minipage}

    \end{minipage}%
    } 

    \caption{
    t-SNE visualization of generated time series on four datasets.
    \textbf{Rows (top to bottom):} CHB-MIT, Sunspot, Electricity, and ECG.
    Red points represent real data and blue points represent generated data.
    Our \textbf{Graph2TS} model demonstrates superior overlap with the real data distribution,
    indicating it captures the global structural properties more effectively than baselines.
    }
    \label{fig:tsne_multi}
\end{figure}

Figures~\ref{fig:acf_multi}, \ref{fig:psd_multi}, and \ref{fig:tsne_multi} provides a qualitative comparison of temporal and distributional properties across models, complementing the quantitative results in Table~\ref{tab:overall_metrics}. We visualize mean autocorrelation functions (ACF), power spectral densities (PSD), and t-SNE embeddings to examine temporal dynamics, frequency content, and manifold geometry, respectively.

Across all datasets, the proposed \textbf{Graph2TS} consistently preserves the global shape of the real ACF, indicating accurate modeling of conditional mean temporal dependencies.
In contrast, DiffusionTS often exhibits inflated short-lag correlations and attenuated mid-range dynamics, while TimeGAN shows less stable and dataset-dependent correlation patterns. These behaviors reflect the tendency of diffusion models to over-smooth
and GAN-based models to distort temporal structure.

The PSD visualizations further reveal that \textbf{Graph2TS} closely matches the dominant low-frequency components of the real signals across datasets. DiffusionTS exhibits pronounced low-frequency energy suppression on high-variability signals such as ECG, whereas TimeGAN produces distorted spectral profiles with misplaced energy.
This highlights the importance of explicitly modeling stochastic residuals to avoid spectral collapse or spurious high-frequency artifacts.

Finally, t-SNE embeddings show that \textbf{Graph2TS} achieves strong overlap between real and synthetic samples, indicating faithful coverage of the data manifold. DiffusionTS tends to contract the manifold under strict structural constraints, while TimeGAN exhibits mode drifting and over-dispersion. Together, these qualitative results illustrate that combining structural conditioning with stochastic residual modeling yields stable temporal dynamics, balanced frequency characteristics, and well-aligned manifold geometry.

\clearpage

\section{Appendix: Theoretical and Implementation Details}
\label{sec-appendix}

\subsection{Formula}
\label{sec:formalua_var}
\paragraph{Derivation of the law of total variance.}
Let $\mu(G) \triangleq \mathbb{E}[X \mid G]$.
By definition,
\begin{equation}
\mathrm{Var}(X)=\mathbb{E}\!\left[\|X-\mathbb{E}[X]\|^2\right].
\end{equation}
Insert and subtract $\mu(G)$:
\begin{equation}
X-\mathbb{E}[X] = \big(X-\mu(G)\big) + \big(\mu(G)-\mathbb{E}[X]\big).
\end{equation}
Therefore,
\begin{align}
\mathrm{Var}(X)
&= \mathbb{E}\!\left[\big\| (X-\mu(G)) + (\mu(G)-\mathbb{E}[X]) \big\|^2\right] \\
&= \mathbb{E}\!\left[\|X-\mu(G)\|^2\right]
 + \mathbb{E}\!\left[\|\mu(G)-\mathbb{E}[X]\|^2\right] \nonumber\\
&\quad\; + 2\,\mathbb{E}\!\left[(X-\mu(G))^\top(\mu(G)-\mathbb{E}[X])\right].
\end{align}
The cross term is zero:
\begin{align}
\mathbb{E}\!\left[(X-\mu(G))^\top(\mu(G)-\mathbb{E}[X])\right]
&= \mathbb{E}\!\left[\mathbb{E}\!\left[(X-\mu(G))^\top(\mu(G)-\mathbb{E}[X]) \mid G\right]\right] \\
&= \mathbb{E}\!\left[(\mu(G)-\mathbb{E}[X])^\top \mathbb{E}[X-\mu(G)\mid G]\right] \\
&= \mathbb{E}\!\left[(\mu(G)-\mathbb{E}[X])^\top \cdot 0 \right] = 0,
\end{align}
because $\mathbb{E}[X-\mu(G)\mid G]=\mathbb{E}[X\mid G]-\mu(G)=0$.
Hence,
\begin{equation}
\mathrm{Var}(X)
=
\mathbb{E}\!\left[\|X-\mu(G)\|^2\right]
+
\mathbb{E}\!\left[\|\mu(G)-\mathbb{E}[X]\|^2\right].
\end{equation}
Finally, observe that
\begin{equation}
\mathbb{E}\!\left[\|X-\mu(G)\|^2\right] = \mathbb{E}\!\left[\mathrm{Var}(X\mid G)\right],
\qquad
\mathbb{E}\!\left[\|\mu(G)-\mathbb{E}[X]\|^2\right] = \mathrm{Var}(\mathbb{E}[X\mid G]),
\end{equation}
which yields the law of total variance:
\begin{equation}
\mathrm{Var}(X)
=
\mathrm{Var}\!\left(\mathbb{E}[X \mid G]\right)
+
\mathbb{E}\!\left[\mathrm{Var}(X \mid G)\right].
\end{equation}

\subsection{Why not graph neural networks?}
Although the quantile graph admits a graph interpretation, it is fundamentally a fixed-size transition matrix
with globally aligned semantics across samples. Each node corresponds to a shared quantile state, and each edge weight represents a normalized transition probability.

In this setting, the primary modeling objective is to encode global transition statistics rather than local neighborhood structure.
We therefore adopt a simple MLP encoder to embed the quantile graph, which is sufficient to capture these statistics while avoiding the additional complexity and inductive assumptions introduced by graph neural networks.

\subsection{Deterministic Graph to Time Series Model}
\label{sec:model_without_stochasticity}

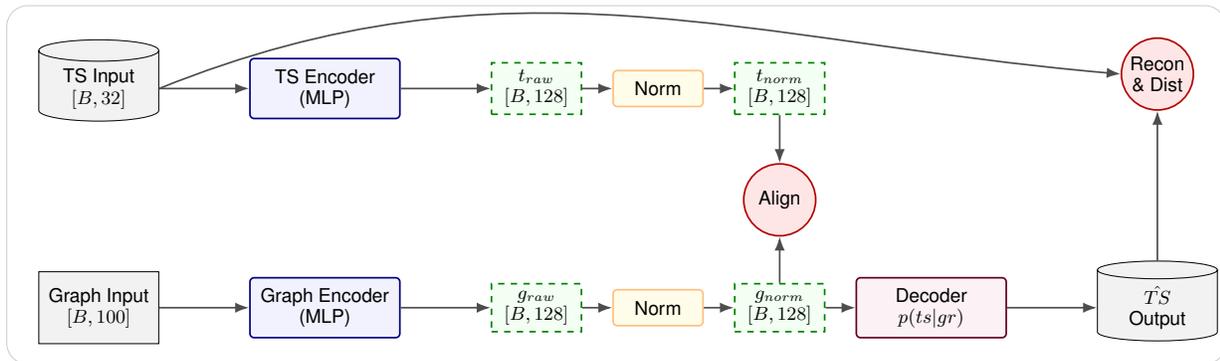
\begin{figure}[h!]
    \centering
    \resizebox{0.95\linewidth}{!}{%
        \begin{tikzpicture}[
    node distance=1.5cm and 1.5cm,
    font=\sffamily\small,
    process/.style={rectangle, draw=blue!50!black, fill=blue!5, thick, minimum width=2.5cm, minimum height=1cm, rounded corners=2pt, align=center},
    tensor/.style={rectangle, draw=green!50!black, fill=green!5, thick, minimum width=1.5cm, minimum height=0.8cm, align=center, dashed},
    input/.style={cylinder, shape border rotate=90, draw=black, fill=gray!10, aspect=0.25, minimum height=1.2cm, minimum width=2cm, align=center},
    g_input/.style={rectangle, shape border rotate=90, draw=black, fill=gray!10, aspect=0.25, minimum height=1.2cm, minimum width=2cm, align=center},
    loss/.style={circle, draw=red!70!black, fill=red!10, thick, minimum size=1.2cm, align=center, inner sep=0pt},
    connection/.style={-Latex, thick, draw=black!70},
    skip/.style={-Latex, thick, draw=blue!70, dashed}
]

    \node[input] (ts_in) {TS Input\\$[B, 32]$};
    \node[g_input, below=2.5cm of ts_in] (gr_in) {Graph Input\\$[B, 100]$};

    \node[process, right=of ts_in] (ts_enc) {TS Encoder\\(MLP)};
    \node[process, right=of gr_in] (gr_enc) {Graph Encoder\\(MLP)};

    \draw[connection] (ts_in) -- (ts_enc);
    \draw[connection] (gr_in) -- (gr_enc);

    \node[tensor, right=of ts_enc] (t_raw) {$t_{raw}$\\$[B, 128]$};
    \node[tensor, right=of gr_enc] (g_raw) {$g_{raw}$\\$[B, 128]$};

    \draw[connection] (ts_enc) -- (t_raw);
    \draw[connection] (gr_enc) -- (g_raw);

    \node[process, right=0.5cm of t_raw, minimum width=1.5cm, minimum height=0.6cm, fill=yellow!10, draw=orange!50] (norm_t) {Norm};
    \node[process, right=0.5cm of g_raw, minimum width=1.5cm, minimum height=0.6cm, fill=yellow!10, draw=orange!50] (norm_g) {Norm}; 

    \node[tensor, right=0.5cm of norm_t] (t_norm) {$t_{norm}$\\$[B, 128]$};
    \node[tensor, right=0.5cm of norm_g] (g_norm) {$g_{norm}$\\$[B, 128]$};
 
    \draw[connection] (norm_t) -- (t_norm);
    \draw[connection] (norm_g) -- (g_norm);
    
    \node[loss, below=0.8cm of t_norm] (align_loss) {Align};

    \draw[connection] (t_raw) -- (norm_t);
    \draw[connection] (g_raw) -- (norm_g); 

    \draw[connection] (t_norm.south) -- (align_loss.north);
    \draw[connection] (g_norm.north) -- (align_loss.south);

    \node[process, right=0.5cm of g_norm, fill=purple!5, draw=purple!50!black] (decoder) {Decoder\\$p(ts|gr)$};

    \draw[connection] (g_norm) -- (decoder.west);


    \node[input, right=of decoder] (ts_out) {$\hat{TS}$\\Output};
    
    \node[loss, above=2.5cm of ts_out, scale=1] (recon_dist_loss) {Recon \\ \& Dist};

    \draw[connection] (decoder) -- (ts_out);
    \draw[connection] (ts_out) -- (recon_dist_loss);
    \draw[connection] (ts_in.east) to[out=22,in=175] (recon_dist_loss.west);

    \begin{scope}[on background layer]
        \node[fit=(ts_in)(recon_dist_loss)(g_raw), fill=white, draw=black!20, rounded corners=10pt, inner sep=15pt] (bg) {};
    \end{scope}

\end{tikzpicture}
    }
    \caption{Deterministic graph to time series model (Graph2TS without stochasticity).}
    \label{fig:model_without_stochasticity}
\end{figure}

Fig.~\ref{sec:model_without_stochasticity} shows the architecture of the deterministic version of the \textbf{Graph2TS}.

\subsection{Detailed Setup}

\paragraph{Hardware and platform}
We run the all the experiments on DAS6~\cite{DBLP:journals/computer/BalELNRSSW16/das6}. All experiments were conducted on a single NVIDIA A40 GPU.

\paragraph{Baselines.}
We compare against representative time-series generators:
\href{https://github.com/jsyoon0823/TimeGAN}{\textbf{TimeGAN}} (GAN-based) and \href{https://github.com/Y-debug-sys/Diffusion-TS}{\textbf{DiffusionTS}}(diffusion-based). All baselines are trained on the same training split and generate sequences of the same length $T$. We use official implementations.

\paragraph{Training protocol and fairness.}
When computing metrics, we generate the same number of synthetic samples as the number of real samples in the evaluation split (4000 sequences) and apply balanced subsampling if needed.
Note that while the sorting permutation itself is non-differentiable, standard auto-differentiation frameworks propagate gradients through the sorted values. This allows the order-statistics loss to effectively guide the generator towards matching the marginal amplitude distribution

\paragraph{Sampling protocol.}
For our model, we sample $n$ sequences per graph by drawing independent $z$ values.
Unless otherwise stated, we set $n=1$ for one-to-one comparison with unconditional generators.

\paragraph{Other experimental settings.}
We use fixed-length univariate windows of length $T=32$ and condition generation on quantile-transition graphs with $Q=10$ global bins, represented as a flattened first-order transition matrix in $\mathbb{R}^{Q^2}$.
The model (\textsc{Graph2TS}) employs two-layer MLP encoders for the time series and the graph, each producing a 128-dimensional \emph{raw} embedding; $\ell_2$-normalized embeddings are used only for contrastive alignment.
The posterior $q_\phi(z\mid x,g)$ is parameterized by an MLP that maps the concatenated raw embeddings to $(\mu,\log\sigma^2)\in\mathbb{R}^{2d_z}$ with $d_z=32$, and the decoder $p_\theta(x\mid g,z)$ is an MLP conditioned on the concatenation of the raw graph embedding and $z$.
Training minimizes a weighted sum of (i) a symmetric InfoNCE alignment loss between time-series and graph embeddings with a learnable temperature initialized to $0.07$, (ii) an $\ell_2$ reconstruction loss $\|x-\hat{x}\|_2^2$, (iii) an order-statistics matching loss $\|\mathrm{sort}(\hat{x})-\mathrm{sort}(x)\|_2^2$, (iv) a KL regularizer $\mathrm{KL}(q_\phi(z\mid x,g)\,\|\,\mathcal{N}(0,I))$ with linear KL annealing.
Unless otherwise stated, we set $w_{\text{align}}=1$, $w_{\text{recon}}=5$, $w_{\text{dist}}=1$, and $\beta_{\max}=0.05$ with a warmup of 50 epochs.
Optimization is performed using Adam with learning rate $3\times10^{-4}$ and batch size 4096, training for 300 epochs; we select the checkpoint with the lowest average training objective and report the corresponding parameters. 

\paragraph{Source code.} We will make our source code publicly available upon acceptance.

\subsection{Datasets}
\label{sec-appendix-datasets}

All datasets are segmented into fixed-length windows of $T=32$ and normalized using z-score normalization within each dataset split.

\paragraph{CHB--MIT Scalp EEG (CHB-MIT).}
The CHB--MIT Scalp EEG Database~\cite{PhysioNet-chbmit-1.0.0} consists of long-term scalp EEG recordings from 22 pediatric subjects with intractable epilepsy.
The recordings span multiple days per subject and include expert annotations of seizure onset and offset.
Following standard practice, we segment the continuous EEG signals into fixed-length windows of $T=32$ and apply z-score normalization within each dataset split.
This dataset is used to evaluate the ability of generative models to capture noisy biomedical signals and to support downstream seizure-related tasks under class imbalance.

\paragraph{Daily Sunspot Number (Sunspot).}
The Daily Sunspot Number dataset~\cite{sunspot} records historical measurements of solar activity from 1818 to 2019.
The data exhibit strong long-term periodicity and non-stationary temporal dynamics.
We extract overlapping windows of length $T=32$ from the normalized daily series to evaluate the model’s ability to preserve global oscillatory structure while maintaining realistic variability.

\paragraph{Electricity Load (Elec).}
The Electricity Load dataset~\cite{electricityloaddiagrams20112014_321} contains electricity consumption measurements from multiple clients over time.
The data are characterized by seasonal patterns, abrupt changes, and heterogeneous dynamics across users.
We follow common preprocessing procedures and segment each univariate series into windows of length $T=32$, which are normalized using z-score normalization per split.

\paragraph{MIT-BIH Arrhythmia ECG (ECG).}
The MIT-BIH Arrhythmia Database~\cite{932724_mitbih} provides ECG recordings annotated with cardiac arrhythmia events.
The signals contain structured low-frequency morphology combined with high-frequency noise and beat-level variability.
We extract fixed-length ECG segments of $T=32$ and apply z-score normalization, using this dataset to assess structural fidelity and temporal alignment in physiological signal generation.

\subsection{Metrics Definition}
\label{sec-appendix-metrics}

Let $\mathcal{X} = \{x_i\}_{i=1}^N$ denote a set of real time series samples and
$\hat{\mathcal{X}} = \{\hat{x}_j\}_{j=1}^N$ denote an equally sized set of
synthetic samples, where each $x \in \mathbb{R}^T$.

\paragraph{Distribution fidelity.}
The \textbf{Wasserstein distance} measures the discrepancy between the marginal
distributions of real and synthetic samples:
\begin{equation}
W(\mathcal{X}, \hat{\mathcal{X}})
=
\inf_{\gamma \in \Pi(\mathcal{X}, \hat{\mathcal{X}})}
\mathbb{E}_{(x,\hat{x}) \sim \gamma}
\big[ \lVert x - \hat{x} \rVert_2 \big],
\end{equation}
where $\Pi(\mathcal{X}, \hat{\mathcal{X}})$ denotes the set of all couplings
between the empirical distributions of $\mathcal{X}$ and $\hat{\mathcal{X}}$.
Lower values indicate closer distributional alignment.

The \textbf{Kolmogorov--Smirnov (KS) statistic} is defined as
\begin{equation}
\mathrm{KS}
=
\sup_{v \in \mathbb{R}}
\big|
F_{\mathcal{X}}(v) - F_{\hat{\mathcal{X}}}(v)
\big|,
\end{equation}
where $F_{\mathcal{X}}$ and $F_{\hat{\mathcal{X}}}$ are the empirical cumulative
distribution functions of real and synthetic samples, respectively.

\paragraph{Temporal structure alignment.}
Let $\mathrm{ACF}(x, \ell)$ denote the autocorrelation of time series $x$ at lag
$\ell$. The \textbf{ACF MAE} is defined as
\begin{equation}
\mathrm{ACF\text{-}MAE}
=
\frac{1}{L}
\sum_{\ell=1}^{L}
\big|
\mathbb{E}_{x \sim \mathcal{X}}[\mathrm{ACF}(x,\ell)]
-
\mathbb{E}_{\hat{x} \sim \hat{\mathcal{X}}}[\mathrm{ACF}(\hat{x},\ell)]
\big|,
\end{equation}
where $L$ is the maximum lag considered.

Let $\mathrm{PSD}(x)$ denote the power spectral density of $x$ estimated using
Welch’s method. The \textbf{PSD $\ell_2$ distance} is defined as
\begin{equation}
\mathrm{PSD\text{-}\ell_2}
=
\left\|
\mathbb{E}_{x \sim \mathcal{X}}[\mathrm{PSD}(x)]
-
\mathbb{E}_{\hat{x} \sim \hat{\mathcal{X}}}[\mathrm{PSD}(\hat{x})]
\right\|_2.
\end{equation}

\paragraph{Representativeness and prototype quality.}
For each real sample $x_i \in \mathcal{X}$, define its nearest synthetic neighbor
as
\begin{equation}
d_i
=
\min_{\hat{x}_j \in \hat{\mathcal{X}}}
\lVert x_i - \hat{x}_j \rVert_2.
\end{equation}
The \textbf{Prototype Error (ProtoErr)} is reported as both the mean and median
of $\{d_i\}_{i=1}^N$.

Let $m_{\mathcal{X}}$ and $m_{\hat{\mathcal{X}}}$ denote the medoids of the real
and synthetic sets, respectively. The \textbf{Medoid Distance Ratio} is defined
as
\begin{equation}
\mathrm{MDR}
=
\frac{
\lVert m_{\mathcal{X}} - m_{\hat{\mathcal{X}}} \rVert_2
}{
\mathbb{E}_{x \sim \mathcal{X}} \big[ \lVert x - m_{\mathcal{X}} \rVert_2 \big]
}.
\end{equation}

\paragraph{Coverage of the real data manifold.}
Let $\tau_q$ denote the $q$-quantile of the real-to-real nearest-neighbor
distance distribution.
The \textbf{Coverage@$\tau_q$} metric is defined as
\begin{equation}
\mathrm{Coverage@}\tau_q
=
\frac{1}{N}
\sum_{i=1}^{N}
\mathbf{1}
\Big(
\min_{\hat{x}_j \in \hat{\mathcal{X}}}
\lVert x_i - \hat{x}_j \rVert_2
\le \tau_q
\Big),
\end{equation}
where $\mathbf{1}(\cdot)$ is the indicator function.
Higher coverage indicates better support overlap between synthetic and real
data manifolds.

\subsection{Acknowledgement of Use of Generative AI}
Generative AI tools were used to assist in the writing and preparation of this manuscript. The authors have reviewed and verified the content and take full responsibility for all materials presented.



\end{document}